\documentclass[final]{cvpr}

\usepackage{times}
\usepackage{epsfig}
\usepackage{graphicx}
\usepackage{amsmath}
\usepackage{amssymb}
\usepackage{tabularx}
\usepackage{xcolor}
\usepackage{tabularx}
\usepackage{amssymb} 
\usepackage[linesnumbered,ruled,vlined]{algorithm2e}
\usepackage{caption}
\usepackage{subcaption}
\usepackage{adjustbox}
\usepackage{caption}
\usepackage{listings}

\usepackage{hyperref}
\hypersetup{
colorlinks=true,
linkcolor=red
}

\SetCommentSty{mycommfont}
\SetKwInput{KwInput}{Input}                % Set the Input
\SetKwInput{KwOutput}{Output}              % set the Output

\usepackage{hyperref}
\hypersetup{
    colorlinks=true,
    linkcolor=blue,
    filecolor=magenta,      
    urlcolor=cyan,
}

\newcommand{\AL}[1]{{\color{blue}{[Andrew: #1]}}}
\newcommand{\AK}[1]{{\color{cyan}{[Amlan: #1]}}}

\def\cvprPaperID{3586} % *** Enter the CVPR Paper ID here
\def\confYear{CVPR 2021}
\begin{document}

%%%%%%%%% TITLE
\title{Towards Good Practices for Efficiently Annotating Large-Scale Image Classification Datasets}

\author{
Yuan-Hong Liao$^{1,2}$, \hspace{2pt} Amlan Kar$^{1,2,3}$, \hspace{2pt} Sanja Fidler$^{1,2,3}$ \vspace{2mm}\\ 
$^{1}$ University of Toronto, $^{2}$ Vector Institute, $^{3}$ NVIDIA\\
\tt\small{\{andrew, amlan, fidler\}@cs.toronto.edu}
}

\maketitle

%%%%%%%%% ABSTRACT
\begin{abstract}

Data is the engine of modern computer vision, which necessitates collecting large-scale datasets. %, both in terms of the volume of data as well as the number of classes. 
This is expensive, and guaranteeing the quality of the labels is a major challenge. 
In this paper, we investigate efficient annotation strategies for collecting multi-class classification labels for a large collection of images. While methods that exploit learnt models for labeling exist, a surprisingly prevalent approach is to query humans for a fixed number of labels per datum and aggregate them, which is expensive.
% Prior work has proposed joint probabilistic modeling of human annotations and machine generated beliefs as a way to minimize cost and maintain robustness to noisy annotators, but has been limited either by scale or was evaluated in simulation that lacked realistic worker noise that occurs in practice.
Building on prior work on online joint probabilistic modeling of human annotations and machine-generated beliefs, we propose modifications and best practices aimed at minimizing human labeling effort. Specifically, we make use of advances in self-supervised learning, view annotation as a semi-supervised learning problem, identify and mitigate pitfalls and ablate several key design choices to propose effective guidelines for labeling. 
%a set of good practices and guidelines for labeling. 
%We propose modifications and best practices aimed at minimizing human labeling effort ``in the wild". 
Our analysis is done in a more realistic simulation that involves querying human labelers, which uncovers issues with evaluation using existing worker simulation methods. Simulated experiments on a 125k image subset of the ImageNet100 show that it can be annotated to 80\% top-1 accuracy with 0.35 annotations per image on average, a 2.7x and 6.7x improvement over prior work and manual annotation, respectively.
%\AK{Maybe code repo could be a bit more informative? good-practices seems too broad to me}
\footnote{Project: \href{https://fidler-lab.github.io/efficient-annotation-cookbook}{\color{blue}{https://fidler-lab.github.io/efficient-annotation-cookbook}}}

\end{abstract}

%%%%%%%%% BODY TEXT
%-------------------------------------------------------------------------
\vspace{-5.5mm}
\section{Introduction}
\vspace{-0.5mm}

%Success stories of applying deep learning algorithms to real world computer vision problems are now ubiquitous.
 Data, the basic unit of machine learning, has tremendous impact on the success of learning-based applications. Much of the recent A.I. revolution can be attributed to the creation of the ImageNet dataset~\cite{deng2009imagenet}, which showed that image classification with deep learning at scale~\cite{krizhevsky2012imagenet} can result in learning strong feature extractors that transfer to domains and tasks beyond the original dataset. Using citations as a proxy, ImageNet has supported at least 40,000 research projects to date. It has been unmatched as a pre-training dataset to downstream tasks, due to its size, diversity and the quality of labels. Since its conception, interest in creating large datasets serving diverse tasks and domains has skyrocketed. Examples include object detection~\cite{wu2019detectron2}, action-recognition-~\cite{EPICKITCHENS20}, %semantic segmentation~\cite{chen2017deeplab}, 
 and 3D reconstruction~\cite{nerf,DIBR19}, in domains such as self-driving~\cite{kitti,nuscenes}, %aerial imaging~\cite{demir2018deepglobe}, 
 and medical imaging~\cite{wang2017chestx}.

\begin{figure}[t!]
\vspace{-1mm}
  \centering
  \includegraphics[width=0.95\linewidth]{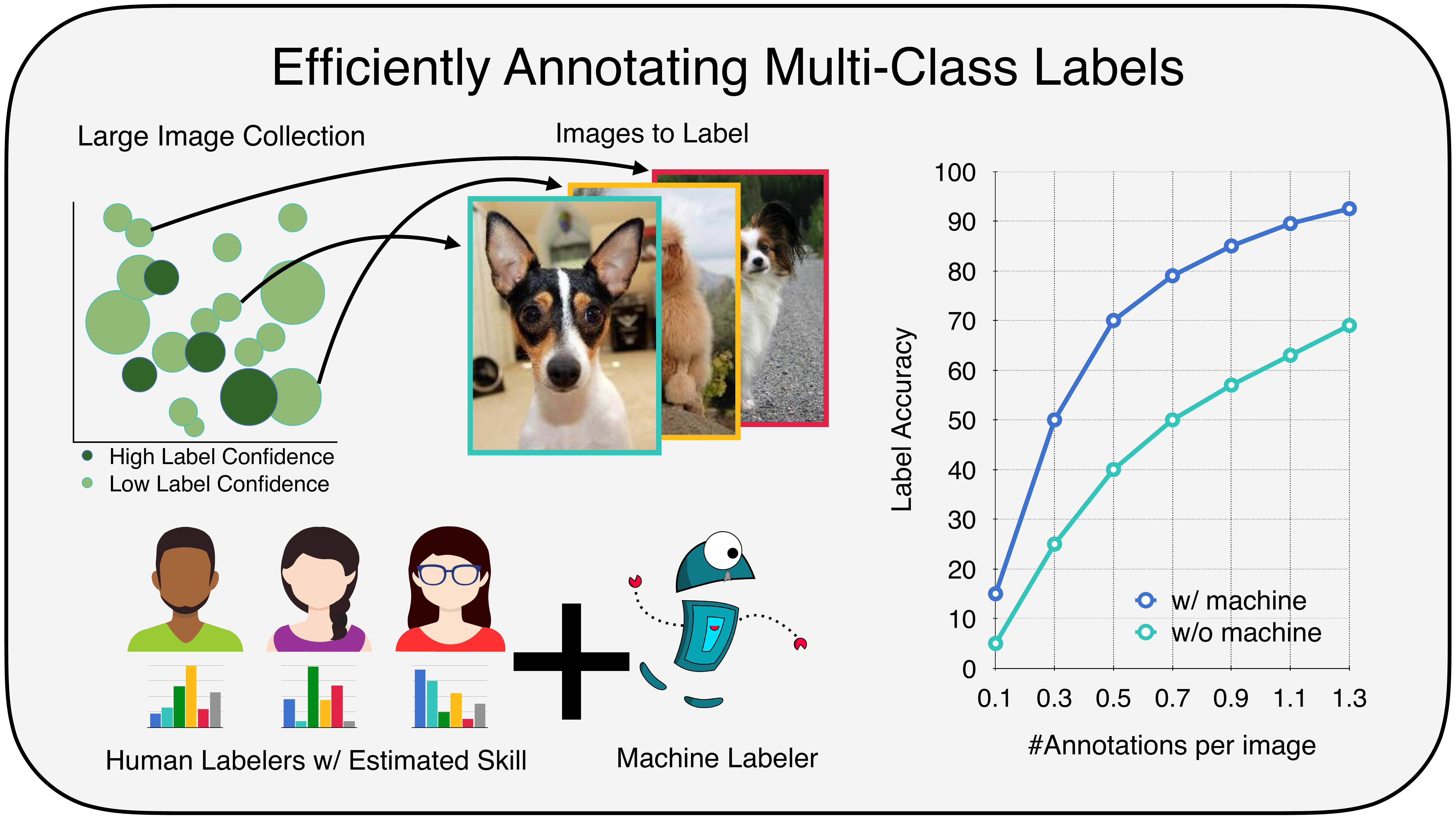}
  \label{fig:intro}
   \vspace{-3.0mm}
  \caption{\small We %present methods for 
 tackle
 efficient model-assisted annotation of multi-class labels at scale. We propose improvements to prior work by incorporating self- and semi-supervised learning and address associated challenges. Extensive ablation of common design choices in realistically simulated experiments leads us to provide best practice recommendations to minimize human annotation effort.} %\AL{Large Image Dataset? Blue and green are not distinct enough for the circles on the left.}}
 \vspace{-3mm}
\end{figure}

ImageNet and its successors such as OpenImages~\cite{OpenImages} collected their data using search engines on the web, followed by human verification of either the search query term or automatically generated labels. Thus, their labeling is formulated as a verification task, \ie, does this image really belong to the class, allowing efficient annotation at scale.

%\AL{The motivation here is misleading. Avoid using the word deployed sensor.} \textcolor{red}{In many practical use cases, the data and labels of interest are often known apriori, \eg when data is from proprietary/deployed sensors or a platform. T}
%, with the process of finding the images providing meta-data for annotation. 
In contrast to ImageNet labeling, in many practical use cases, the data and labels of interest are often known apriori. This departs from the case above where arbitrary images could be used by querying keywords online.
A common approach used in practice is to query humans to get a fixed number of labels per datum and aggregate them~\cite{lin2014microsoft, kaur2019foodx}, presumably because of its simplicity and reliability. This can be prohibitively expensive and inefficient in human resource utilization for large datasets, as it assumes equal effort needed per datum. We build on prior work and investigate integration of modern learning methods to improve annotation efficiency for multi-class classification at \emph{scale}.

%\SF{Do we need this here? it's a long paragraph and the main message seems to be that we use a fixed worker pool -- why not just say this as your setting, and provide a short argument for it?} 
% \AK{We're also introducing different avenues of related work, the DS model and lean crowdsourcing here too? I also think it is important for people to appreciate the difference between crowdsourcing and this, which is why there are 2-3 sentences at the end. I separated these into two paragraphs, does that work better?}
Recent work~\cite{Branson_2017_CVPR} explored integrating a learnt classifier into the DS model~\cite{dawid1979maximum} in an online setting. Their method allows principled online estimation of worker skills and label uncertainty. This is used to decide whether another human should be queried for a datum. We follow this framework, while noting that directions such as design of user-interfaces~\cite{deng2014scalable}, computing optimal task assignment~\cite{ho2013adaptive} etc. can provide complementary benefits.

Having a pool of workers that can be repeatably queried improves both skill estimation over time and reduces annotation noise typically found in \emph{crowdsourcing}, where workers perform micro-tasks and their presence is fleeting. Thus, in this work we choose to focus on a \emph{fixed worker pool}.\footnote{This is also a growing trend in the industry, with large companies using trained labelers over offering micro-tasks.}

%\SF{we first look at/investigate? Can we be more affirmative? eg say, We investigate the 1) use of the latest machine learning models in the loop, specifically self-supervised learners as they do not require labels. Using machine learning models opens the door to continuously improving during data annotation, which we investigate this and this way. Using machine learning models is also dangerous because of this and this, and we provide simple but effective ways to incorporate ML models as additional workers with carefully estimated skills...} 
We first investigate integrating advances in self-supervised learning in our setting. Next, we view online labeling as a semi-supervised problem and show consequent efficiency gains. These additions can sometimes lead to negative feedback cycles, which we identify and remedy. Finally, to encourage adoption into a practitioner's toolchain, we ablate several key design choices and provide a set of good practices and guidelines. We avoid the expense of running large experiments with human workers by proposing a more realistic annotator simulation that involves collecting statistics from human annotators. Prior work~\cite{Branson_2017_CVPR, van2018lean} collected a large number of human labels for all experiments, leading to 1) smaller individual experiment scale and 2) a barrier for further research since these labels are not available and expensive to collect. We note that ~\cite{van2018lean} also look into efficient multi-class annotation for large label sets, with a focus on efficient factorization and learning of worker abilities. This is important and orthogonal to our exploration into integration of learning methods. In summary, we make the following contributions:
\begin{itemize}
  \item Explore the usage of advances in self-supervised learning to efficient annotation for multi-class classification
  \item Propose to view the annotation process as a semi-supervised learning problem, identify resulting instabilities and provide remedies
  \item Ablate several key design choices for the annotation process providing a set of best practices and guidelines to facilitate adoption into a practitioner's toolchain
  \item Provide a realistic annotator simulation to conduct such experiments at scale while avoiding the high cost of involving human annotators for every experiment
  \item Release a modular codebase to facilitate adoption and further research into efficient human-in-the-loop multi-class labeling
\end{itemize}

We experiment on subsets of varying difficulty from ImageNet~\cite{deng2009imagenet}. We show 87\% top-1 label accuracy on a 100 class subset of ImageNet, with only 0.98 annotations per image. 80\% top-1 label accuracy needed 0.35 annotations per image, a 2.7x reduction with respect to prior work and a 6.7x reduction over %fully 
manual annotation.
On the small-scale experiment using human annotations, we achieve 91\% label accuracy with 2x fewer annotations.

\vspace{-3mm}
\section{Related Work}

\subsection{Image Annotation in Computer Vision}
Large datasets~\cite{krizhevsky2012imagenet, lin2014microsoft} have had a pivotal role in recent advances in computer vision.~\cite{kovashka2016crowdsourcing} make a comprehensive survey of crowdsourcing techniques. As an example,~\cite{Russakovsky_2015_CVPR} integrates machines and human-labelers as an MDP and optimize questions asked to labelers to collect an object detection dataset. LSUN \cite{yu2015lsun} interleaves worker annotation and training machine models, making a large high-quality dataset at a low cost. The machine models here are used to perform confirmation (ExistOrNot question). Instead, we adopt a probabilistic framework~\cite{dawid1979maximum} that incorporates machine beliefs with human annotated labels in a principled manner.
We follow the method proposed in~\cite{Branson_2017_CVPR,van2018lean}, who extend~\cite{dawid1979maximum} to an online crowdsourcing framework that includes the learner as a prior. 
We however, consider the learning problem as a semi-supervised task. It is also akin to active learning, where the task to optimize for is the quality of the collected dataset itself.

\subsection{Semi-supervised Learning}
We treat the learning task in online image annotation as a semi-supervised problem enabling us to incorporate various algorithms. Graph-based semi-supervised learning \cite{zhou2004learning, wang2007label} leverages the structure in both labeled and unlabeled data for \emph{transductive} learning. In neural networks, several methods aim at smoothing the decision boundaries of the learnt model by enforcing consistency between image augmentations \cite{xie2019unsupervised, miyato2018virtual}, leveraging pseudo labels on unlabelled data \cite{tarvainen2017mean, xie2020self, sohn2020fixmatch}, interpolating between data points \cite{verma2019interpolation}. 

Recently, \textbf{self-supervised learning} to learn strong representations from unlabelled image collections has been shown to be highly performant, allowing learning tasks with limited labels. Multiple pre-text tasks are proposed to learn representation encoders from unlabelled images, such as predicting patch position \cite{doersch2015unsupervised}, predicting image rotation \cite{gidaris2018unsupervised}, solving an image jigsaw puzzle \cite{noroozi2016unsupervised}, etc.
Contrastive learning methods have gained popularity recently \cite{he2019moco, chen2020mocov2, grill2020bootstrap, chen2020simple, chen2020big, chuang2020debiased, robinson2020contrastive}, where consistency across augmentations of an image and inconsistency across multiple images is used as a learning signal.

\subsection{Truth Inference}
Crowdsourcing provides low-cost noisy annotations. To infer a true label from noisy observations, one needs to infer the importance of each annotation. This problem was discussed 40 years ago in the medical domain with the Dawid-Skene model \cite{dawid1979maximum}, optimized with EM. Many variants have since 
%been proposed to 
extended the DS model. GLAD \cite{whitehill2009whose} assumes a scenario with heterogeneous image difficulty and worker reliability. \cite{welinder2010multidimensional} consider each worker as a multidimensional entity with variables representing competence, expertise and bias.
\cite{van2018lean} parametrize worker skills factorized by a taxonomy of %the 
target concepts.
Instead of using EM, BCC~\cite{kim2012bayesian} infers the unobserved variables with bayesian inference.
EBCC~\cite{li2019exploiting} extends BCC by considering underlying worker correlations and use rank-1 approximations to scale to large worker cohorts. However, there is still no dominant truth inference strategy~\cite{zheng2017truth}. Work in~\cite{Branson_2017_CVPR,van2018lean} is the closest to ours.
\cite{Branson_2017_CVPR} extend the DS model to an online setting and incorporate a learning model as a prior. We follow this direction for multiclass classification, view the problem as a semi-supervised learning problem and rigorously ablate various design parameters as a means to provide best practice guidelines. \cite{van2018lean} also work on multiclass classification, but focus on worker parametrization, whereas we focus on improving %strong 
machine learning models in the procedure.

\newcommand\ww{0.31}
\newcommand\hh{1.1cm}
\begin{figure*}[t!]
\begin{minipage}{0.66\linewidth}
\begin{scriptsize}
  \begin{subfigure}[b]{0.32\textwidth}
      \centering
      \includegraphics[width=\ww\linewidth,height=\hh]{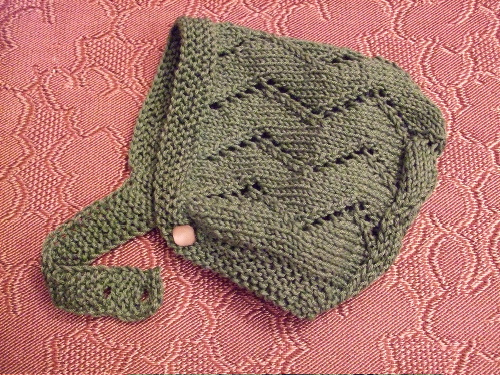}
      \includegraphics[width=\ww\linewidth,height=\hh]{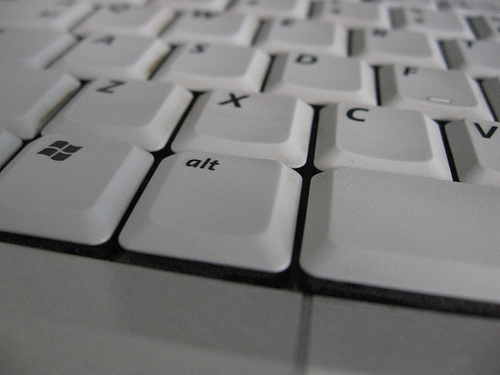}
      \includegraphics[width=\ww\linewidth,height=\hh,trim=0 160 0 60, clip]{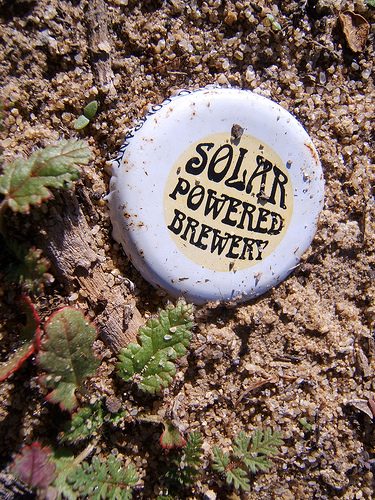}
      \includegraphics[width=\ww\linewidth,height=\hh]{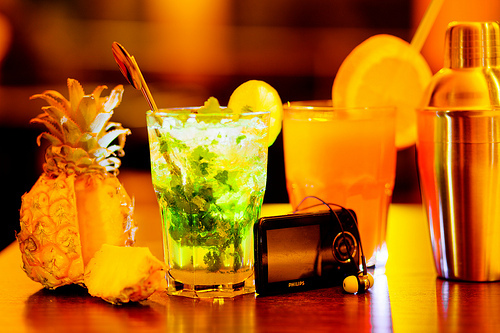}
      \includegraphics[width=\ww\linewidth,height=\hh]{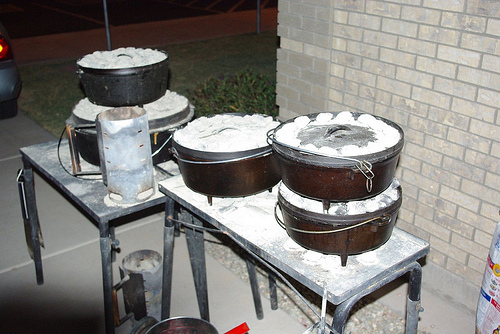}
      \includegraphics[width=\ww\linewidth,height=\hh, trim=0 0 0 35, clip]{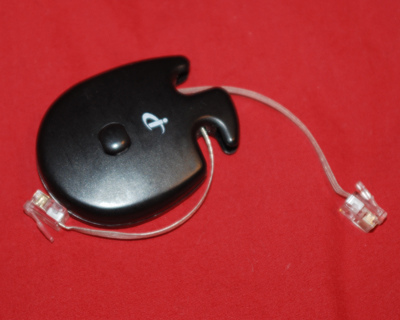}\\[-1.2mm]
      \caption{Commodity}
  \end{subfigure}
  \hfill
  \begin{subfigure}[b]{\ww\textwidth}
      \centering
      \includegraphics[width=\ww\linewidth,height=\hh]{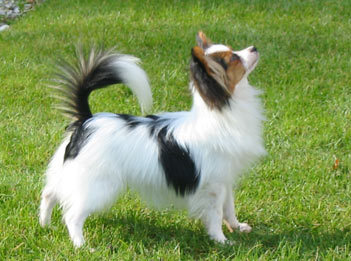}
      \includegraphics[width=\ww\linewidth,height=\hh]{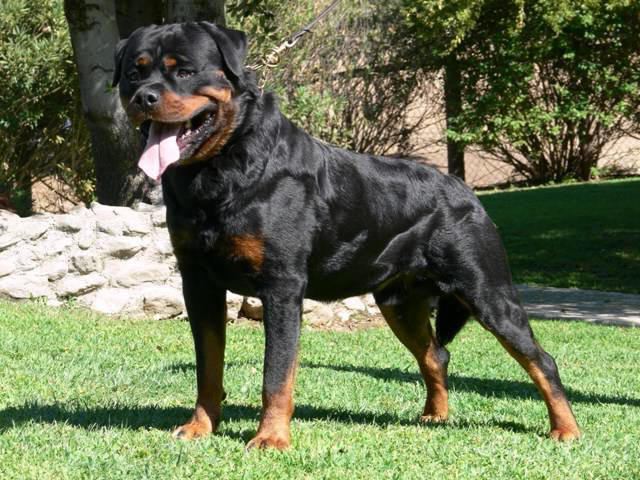}
      \includegraphics[width=\ww\linewidth,height=\hh]{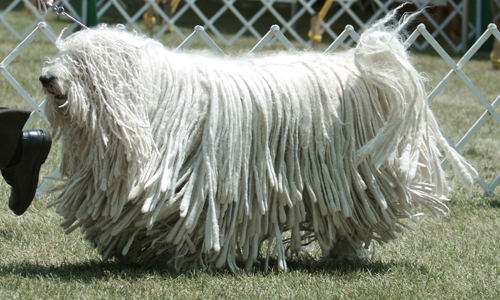}
      \includegraphics[width=\ww\linewidth,height=\hh]{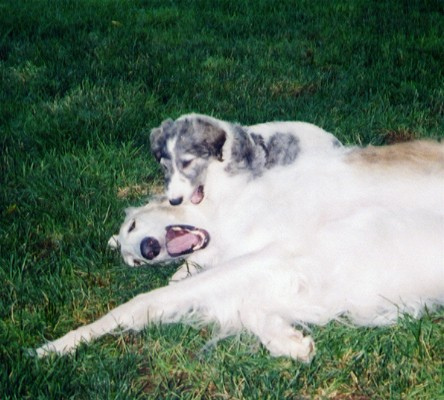}
      \includegraphics[width=\ww\linewidth,height=\hh, trim=0 110 0 50, clip]{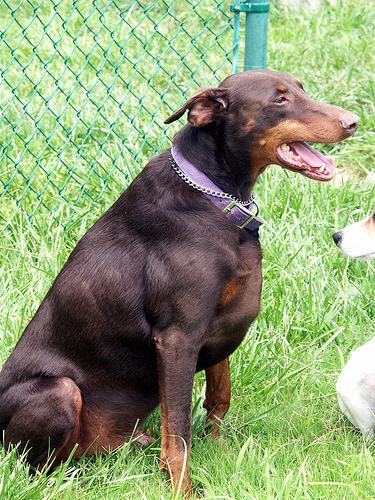}
      \includegraphics[width=\ww\linewidth,height=\hh, trim=0 30 0 0, clip]{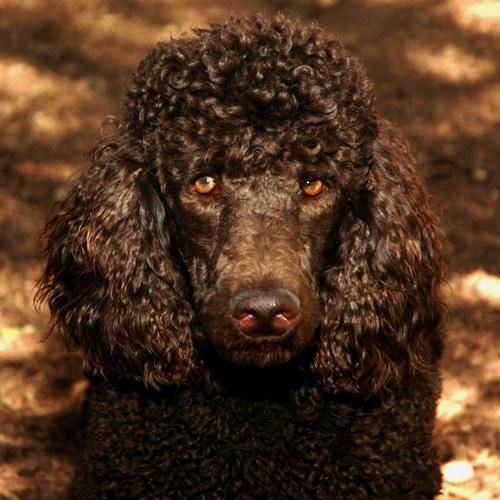}\\[-1.2mm]
      \caption{Dog}
  \end{subfigure}
  \hfill
  \begin{subfigure}[b]{\ww\textwidth}
      \centering
      \includegraphics[width=\ww\linewidth,height=\hh]{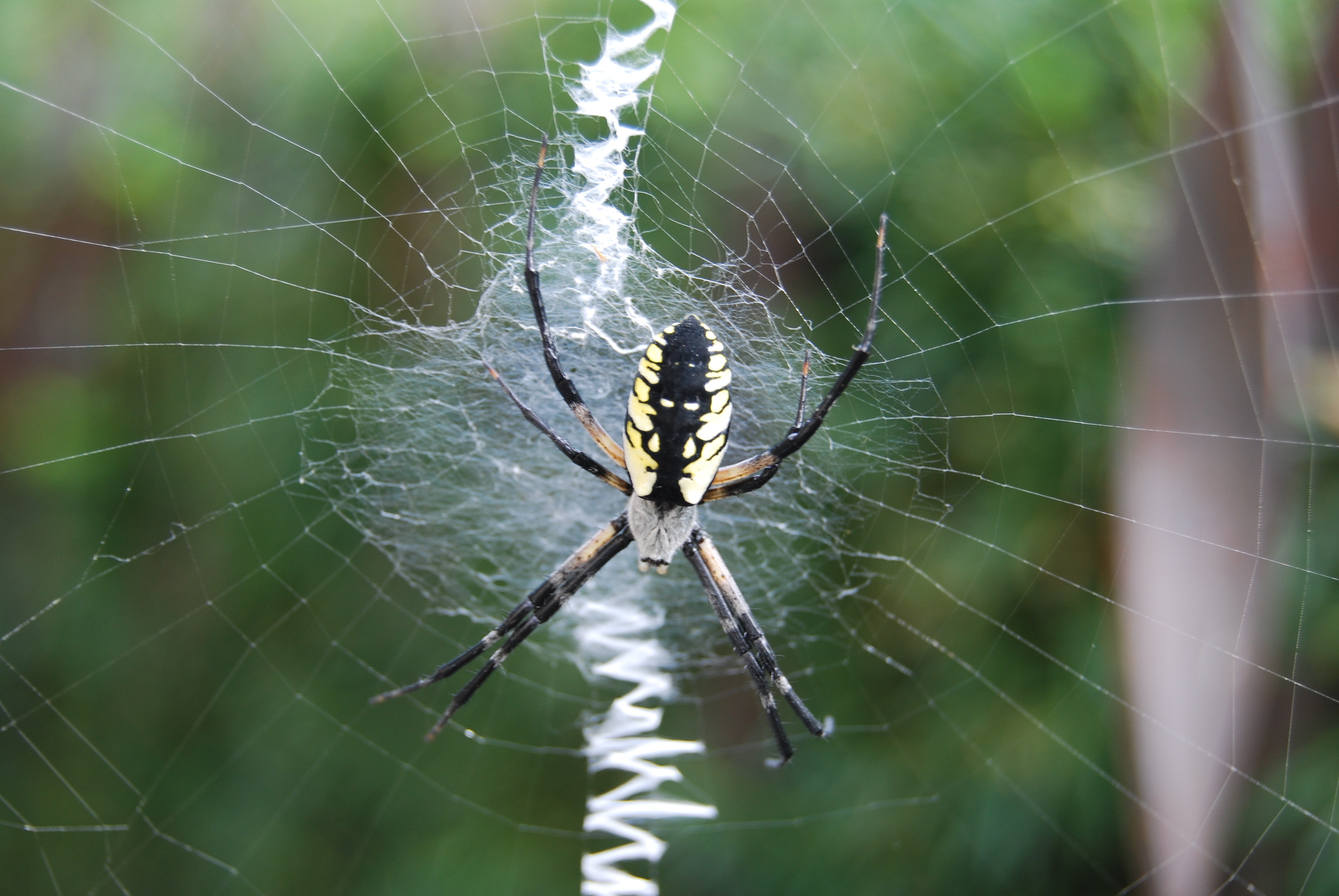}
      \includegraphics[width=\ww\linewidth,height=\hh]{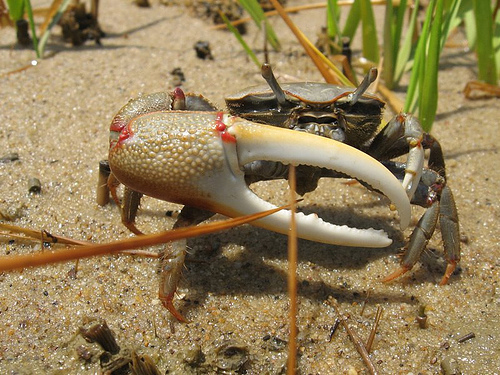}
      \includegraphics[width=\ww\linewidth,height=\hh]{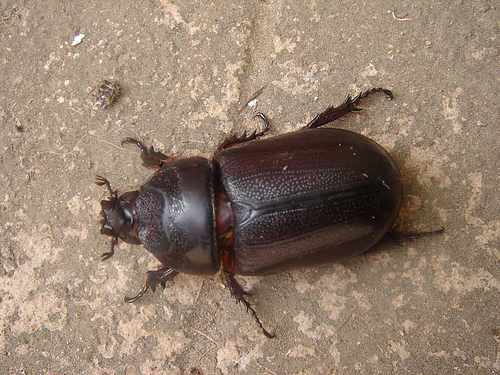}
      \includegraphics[width=\ww\linewidth,height=\hh]{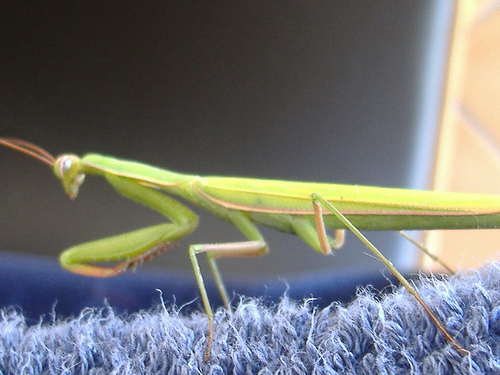}
      \includegraphics[width=\ww\linewidth,height=\hh]{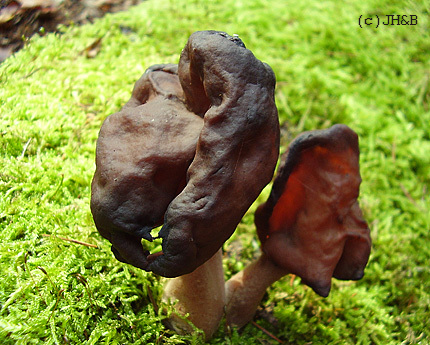}
      \includegraphics[width=\ww\linewidth,height=\hh]{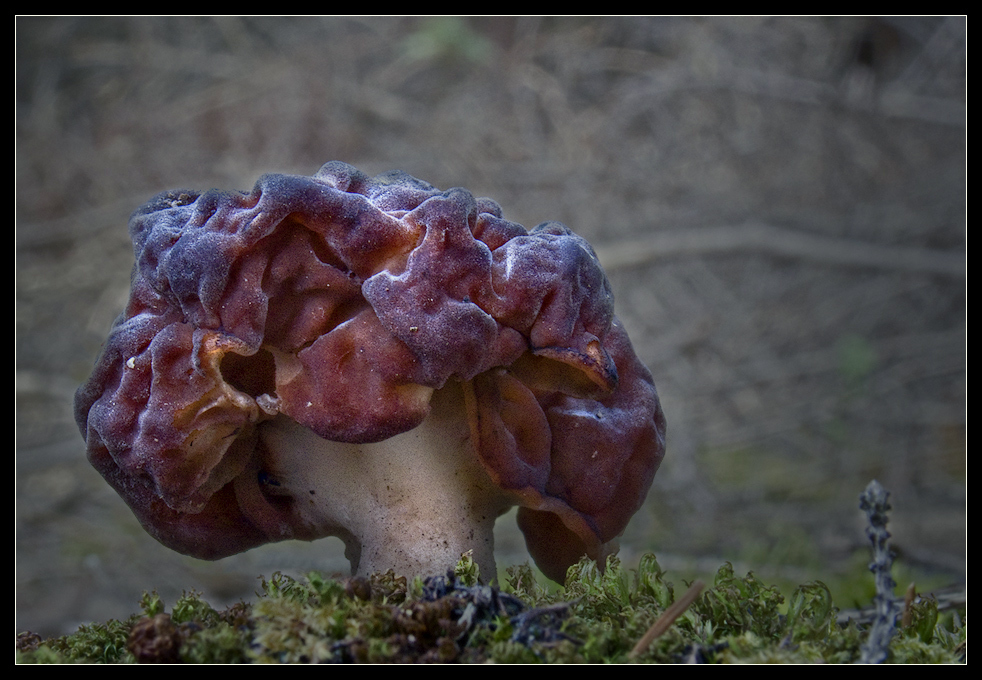}\\[-1.2mm]
      \caption{Insect + Fungus}
  \end{subfigure}
  \end{scriptsize}
  \vspace{-2.5mm}
  \caption{\small \textbf{Example images from datasets} (Tab.~\ref{tab:imagenet100_sub_tasks}). \textit{Commodity} dataset consists of data with coarse-labels, while the \textit{Dog} and \textit{Insect+Fungus} datasets are more fine-grained and difficult to annotate, which is reflected in the lower avg. worker accuracy.}
  \label{fig:example_images}
  \vspace{-2.5mm}
  \end{minipage}
  \hfill
  \begin{minipage}{0.32\linewidth}
%  \begin{table}[t!]
%\vspace{-2mm}
\centering
\resizebox{1.0\linewidth}{!}{%
\addtolength{\tabcolsep}{-4.5pt}
\begin{tabular}{l|c|c|c|c}
Dataset               & \# Images & \# Classes & Worker Acc. & Fine-Grain. \\ \hline\hline
Commodity             & 20140        & 16            & 0.76            &              \\ \hline
Vertebrate            & 23220        & 18            & 0.72            &              \\ \hline
Insect + Fungus       & 16770        & 13            & 0.65            & \checkmark   \\ \hline
Dog                   & 22704        & 19            & 0.43            & \checkmark   \\ \hline
Dog + Vertebrate      & 45924        & 37            & 0.59            & \checkmark   \\ \hline
ImageNet100           & 125689       & 100           & 0.70            & \checkmark   \\ \hline
\end{tabular}%
}
\vspace{-2mm}
\caption{\small \textbf{ImageNet100 sub-tasks.} We separate ImageNet100 into different difficulty levels.}
\label{tab:imagenet100_sub_tasks}
\vspace{-4mm}
%\end{table}
\end{minipage}
\end{figure*}

\vspace{-2mm}
\section{Background}

In this section, we first formulate our problem and introduce the notation used (Sec.~\ref{ss:problem_form}). Next, we describe the DS model~\cite{dawid1979maximum} for probabilistic label aggregation (Sec.~\ref{ss:ds_model}) and its extension to an online data collection setting with a learning algorithm in the loop~\cite{Branson_2017_CVPR} (Sec.~\ref{ss:lean}).

\iffalse
We follow the previous work \cite{Branson_2017_CVPR} that adopt a probabilistic framework to sequentially collect worker annotations and jointly infer the worker skills and truth label.
Additionally, a computer vision is trained online and considered as prior in the framework.
Despite its good performance, we wonder ``Does this framework work in real world problems?'' or ``Can we use this to annotate an ImageNet-like dataset?''.
We will approach the answer to this question from several perspectives and provide in-depth analysis in dataset with various levels of difficulty.
Before presenting the proposed framework and the in-depth analysis, we will introduce multiclass online crowdsourcing in this section.
\fi

\subsection{Problem Formulation}
\label{ss:problem_form}
Given a dataset with $N$ images $\mathcal{X}=\{x_i\}_{i=1:N}$ and a set of $K$ target labels, the goal is to infer the corresponding ``true" label $\mathcal{Y}=\{y_i | y_i \in [K]\}_{i=1:N}$ from worker annotated labels $\mathcal{Z}$. Labels are sampled from $M$ workers $\mathcal{W}=\{w_j\}_{j=1:M}$.
Worker annotations are noisy, hence each image is labeled by few workers $\mathcal{W}_i = \{ j | z_{ij} \in \mathcal{Z}\}$, where $z_{ij}$ is the label assigned by worker $w_j$ on image $x_i$. In an online setting (Sec.~\ref{ss:lean}), at each time step $t$, a requester constructs a batch (of size $B$) of Human Intelligence Tasks (HITs) and assigns them to $B$ workers.
Online estimates of the true label $\mathcal{Y}^t$ at each step can be inferred from all previous annotations $\mathcal{Z}^{1:t}$. The process ends until the requester is satisfied with the current $\mathcal{Y}^t$ or if a time horizon (related to having a budget) $T$ is reached. We omit $t$ %unless necessary 
for simplicity.
 
 \subsection{Dawid-Skene Model}
 \label{ss:ds_model}
 The Dawid-Skene model~\cite{dawid1979maximum} views the annotation process as jointly inferring true labels and worker reliabilities. The joint probability of labels $\mathcal{Y}$, worker annotations $\mathcal{Z}$, and worker reliability $\mathcal{W}$ (overloaded notation for simplicity) is defined as $P(\mathcal{Y}, \mathcal{Z}, \mathcal{W}) = \prod_{i \in [N]} p(y_i) \prod_{j \in [M]}p(w_j) \prod_{i,j \in \mathcal{W}_i} p(z_{ij} | y_i, w_j)$, where $p(y_i)$ is the prior over $K$ possible labels, $p(w_j)$ is the prior reliability of worker $j$, and $p(z_{ij} | y_i, w_j)$ models the likelihood of worker annotations. The worker reliability is usually represented as a confusion matrix over the label set. In practice, inference is performed using expectation maximization, where parameters for one image or worker are optimized at a time,
\begin{align} 
\bar{y_i} &= \arg\max_{y_i} p(y_i) \prod_{j \in \mathcal{W}_i} p(z_{ij} | y_i, \bar{w_j}) 
\label{eq:label_aggregation} \\
\bar{w_j} &= \arg\max_{w_j} p(w_j) \prod_{i \in \mathcal{I}_j} p(z_{ij} | \bar{y_i}, w_j) 
\label{eq:worker_skill_inference}
\end{align}
where $\mathcal{I}_j = \{i | z_{ij} \in Z\}$ is the set of images annotated by worker $j$. We refer readers to \cite{zheng2017truth} for a comprehensive explanation and comparison of other work on truth inference.

\subsection{Extension to Online Labeling}
\label{ss:lean}
Lean Crowdsourcing~\cite{Branson_2017_CVPR} extends the DS model with a learning model in the loop and implements it in an online setting. The authors replace the label prior $p(y_i)$ with predicted probabilities from a learnt model $p(y_i | x_i, \theta)$. At each time step, after running Eq.~\ref{eq:label_aggregation} and Eq.~\ref{eq:worker_skill_inference}, they additionally optimize the %learning 
model parameters $\theta$ from $\mathcal{D} = \{x_i, \bar{y_i} | \mathopen|\mathcal{W}_i\mathclose| > 0\}$, \ie using the current label estimate for images with at least one human annotation. Their learning model involves a fixed feature extractor $\phi$ with a classifier head. In this work, we use a 2 layer MLP, optimized with gradient descent.\footnote{Previous work~\cite{Branson_2017_CVPR} used a linear SVM as the classifier head, but we found that using 2 layer MLP sufficient to achieve comparable performance with far less time.} Its parameters are learnt from scratch at every step by minimizing a loss function $H$.
\begin{align} 
\begin{split}
\bar{\theta} = \arg\min_{\theta} \mathbb{E}_{(x_i, y_i) \sim \mathcal{D}}  H(\bar{y_i}, p(y_i | \phi(x_i), \theta))
\label{eq:learning_in_the_loop} 
\end{split}
\end{align} 
 
To construct $B$ HITs for the next step, they compute the bayesian risk of $\bar{y_i}$ as the expected cost of mis-labeling,
\begin{align} 
 \begin{split}
 \mathcal{R}(\bar{y_i}) 
 &= \sum_{y_i=1}^{K} H(\bar{y_i}, y_i) p(y_i | \mathcal{Z}_i, \theta) \\
 &= \sum_{y_i=1}^{K} H(\bar{y_i}, y_i) \frac{p(y_i | x_i, \bar{\theta}) \prod_{j \in \mathcal{W}_i} p(z_{ij} | y_i, \bar{w_j})}{\sum_{y=1}^K p(y | x_i, \bar{\theta}) \prod_{j \in \mathcal{W}_{i}} p(z_{ij} | y, \bar{w_j})} 
 \end{split}
 \label{eq:risk_estimation}
 \end{align}
At every step, they construct $B$ HITs by randomly sampling from a set of unfinished examples, $\mathcal{U} = \{ x_i | \mathcal{R}(\bar{y_i}) \geq C \}$ \ie images with risk greater than a threshold.
To compare with online labeling without any learning model, we adopt this sampling scheme and remove the model learning, referred to as ``online DS'' in the following sections.

In the following sections, we propose improvements to this online labeling framework,
both in how learnt models are used and in practical design choices. Our proposed improvements are validated on multiple subsets of varying difficulty from the ImageNet dataset, using realistically simulated labelers, both of which we also introduce next.
% \iffalse
% \section{ImageNet100-sandbox}

% Evaluating and ablating multi-class label annotation efficiency at scale requires using large datasets with diverse and relatively clean labels. We construct multiple subsets of the ImageNet dataset~\cite{deng2009imagenet} for our experiments. We use 100 classes sampled from the ImageNet label set~\cite{tian2019contrastive} and construct smaller sub-tasks of varying difficulty using the label hierarchy. The constructed tasks range from 20k to 125k images in size. We use the average accuracy of human workers on these datasets (see Sec.~\ref{ss:realistic_workers}) as a proxy for their difficulty, with the average human accuracy ranging from 0.45 to 0.79. Tab.~\ref{tab:imagenet100_sub_tasks} details the different sub-tasks, the number of images, classes and average human accuracy on them. Fig.~\ref{fig:example_images} shows example images from these tasks. For each class, we use 10 images as prototype images, to be provided by the task creator. These help ground concepts for human annotators, while also helping with learning and model selection (Sec.~\ref{sec:exps}).
% \fi

\begin{figure}[t!]
\vspace{-3mm}
    \centering    
    \includegraphics[width=0.73\linewidth, height=4.5cm]{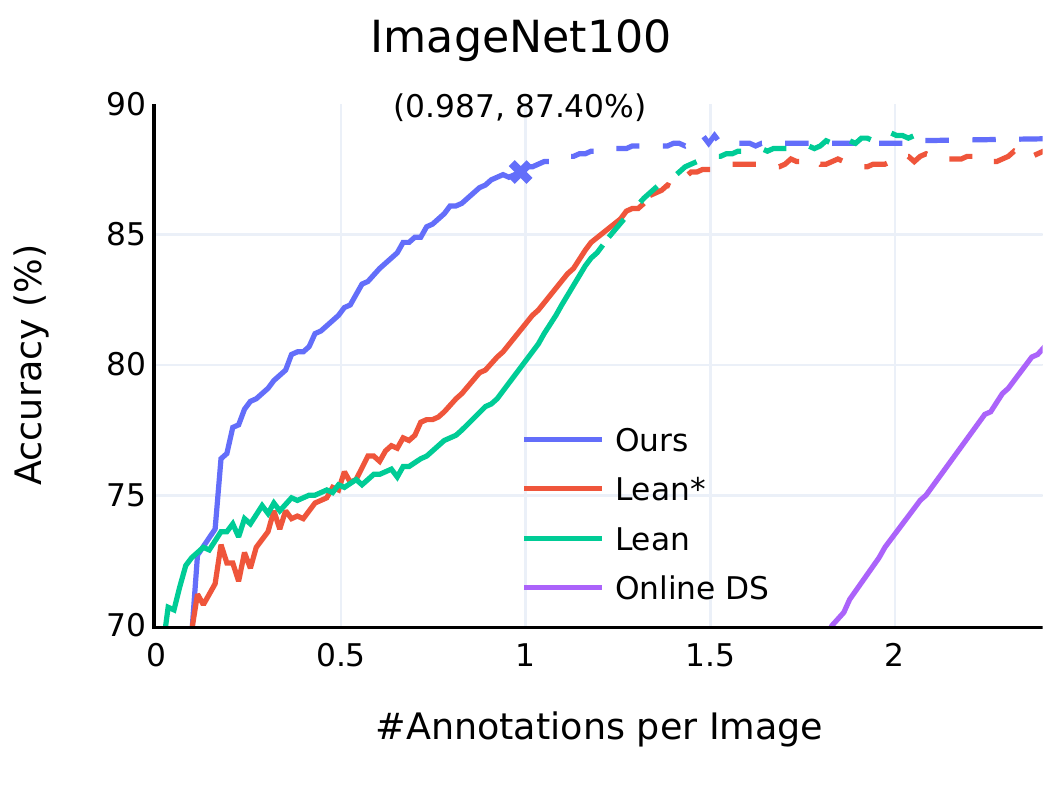}
    \vspace{-3mm}
    \caption{\small \textbf{Results on ImageNet100.} We compare our full framework with ~\cite{Branson_2017_CVPR} and the online DS model on our ImageNet100 dataset (125k images). Our framework achieves 80\% top-1 label accuracy with 0.35 annotation per image, a 2.7x reduction from~\cite{Branson_2017_CVPR}, and 6.7x compared to the online DS model.}
    \label{fig:exps_imagenet100}
     \vspace{-1mm}
\end{figure}

\begin{figure}[t!]
\vspace{-3mm}
    \centering
    \includegraphics[width=0.99\linewidth,trim=0 5 0 0,clip]{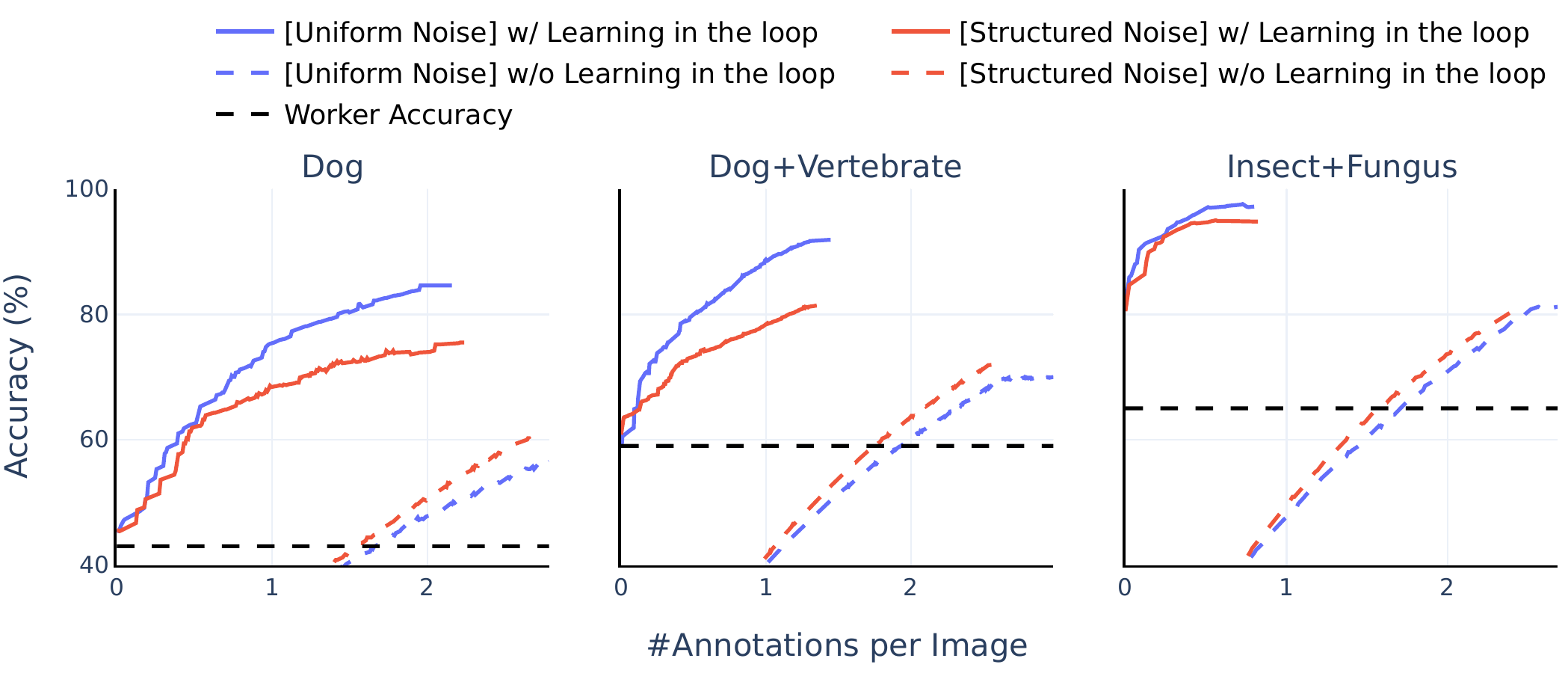}
    \vspace{-3mm}
    \caption{\small \textbf{Over-optimistic results from workers with uniform noise.} Human workers tend to make ``structured'' mistakes. Simulated workers with uniform label noise (blue) can result in over-optimistic annotation performance. Experiments under workers with structured noise reflect real-life performance better.}
    \label{fig:exps_unfiorm_structured_noise_pseudolabel}
    \vspace{-6mm}
\end{figure}

 \begin{figure*}[t]
 \vspace{-1mm}
 \begin{minipage}{0.49\linewidth}
    \centering
    \includegraphics[width=0.99\linewidth]{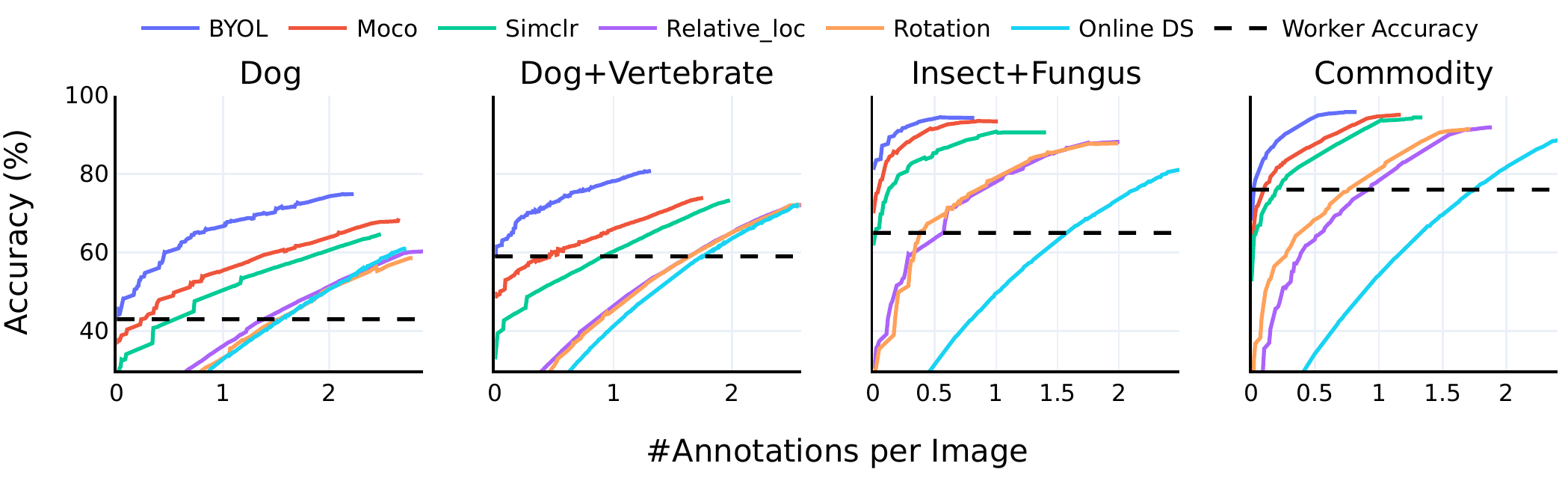}
    \vspace{-3.8mm}
    \caption{\footnotesize \textbf{Self-supervised features advance online labeling.} We compare different self-supervised features, showing that improvements in self-supervised learning translate into improvements in online labeling. For the \textit{Dog}, using BYOL~\cite{grill2020bootstrap} helps reach the same accuracy achieved without learning in the loop (Online DS) with 5x fewer annotations. (Sec.~\ref{ss:self_sup})}%\AL{We are not using semi-supervised learning here.}}
    \label{fig:exps_self_supervised}
    \end{minipage}
    \hspace{2mm}
    \begin{minipage}{0.50\linewidth}
        \centering
    \includegraphics[width=0.95\linewidth]{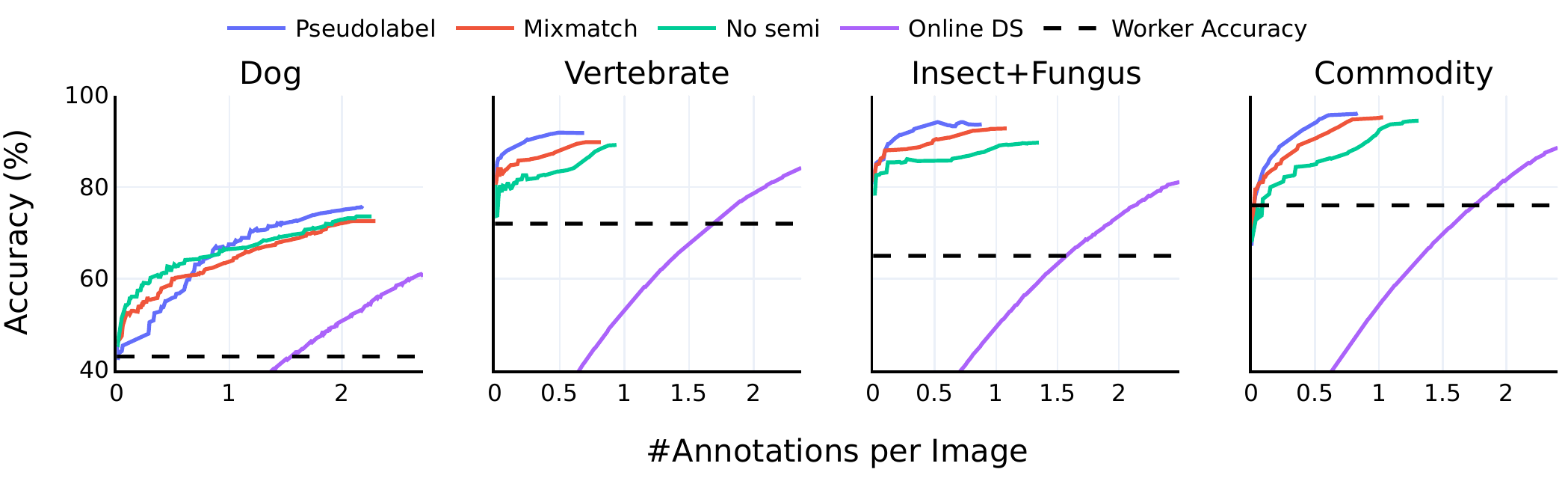}
    \vspace{-3.5mm}
    \caption{\footnotesize \textbf{Semi-supervised learning advances online labeling.} The learning problem in online-labeling can be seen as a semi-supervised problem (Sec.~\ref{ss:semi_sup}). We compare two semi-supervised techniques: Pseudo-Labeling~\cite{lee2013pseudo} and MixMatch~\cite{berthelot2019mixmatch}. Both improve annotation efficiency on all subsets, particularly for fine-grained datasets. Surprisingly, Pseudolabels performs slightly better than a modified version of MixMatch (Sec.~\ref{ss:semi_sup}).}
    \label{fig:exps_semi_supervised}
        \end{minipage}
     \vspace{-6mm}
\end{figure*}

% \iffalse
% \begin{figure*}
%     \centering
%     \includegraphics[width=0.8\linewidth]{assets/images/plot_semi.pdf}
%     \vspace{-3mm}
%     \caption{\small Learning during online-labeling can be seen as a semi-supervised problem (Sec.~\ref{ss:semi_sup}). We compare two semi-supervised learning techniques: Pseudo-Labeling~\cite{lee2013pseudo} and MixMatch~\cite{berthelot2019mixmatch}. Semi-supervised techniques improve annotation efficiency across all datasets, particularly for fine-grained datasets. Surprisingly, Pseudolabels slightly outperform a modified version of MixMatch (see Sec.~\ref{ss:semi_sup} for a discussion).}
%     \label{fig:exps_semi_supervised}
%      \vspace{-4mm}
% \end{figure*}
% \fi

\vspace{-2mm}
\section{Improving Annotation Efficiency}
\label{sec:exps}

In this section, we explore modifications to the online-labeling algorithm proposed in Lean Crowdsourcing~\cite{Branson_2017_CVPR}, focused on improving efficiency of learning in the loop and improving practical implementation choices. We present these as proposals followed by their resulting impact on annotation efficiency and label accuracy.

We first introduce datasets in Sec.~\ref{ss:imagenet100-sandbox} constructed by taking subsets of ImageNet~\cite{deng2009imagenet} which we use for experiments. Experiments and ablations at scale with human labelers are prohibitively expensive. Hence, we also propose a more realistic worker simulation in Sec.~\ref{ss:realistic_workers}.

In Sec.~\ref{ss:self_sup}, we investigate whether self-supervised learning can be used to replace the feature extractor $\phi$ in the algorithm. Next, in Sec.~\ref{ss:semi_sup} we cast the learning problem in online-labeling as a semi-supervised problem. We identify that semi-supervised learning during online-labeling can cause a negative feedback loop in the algorithm, which we mitigate. In Sec.~\ref{ss:practical}, we ablate several key design choices and provide good practices and guidelines to encourage future adoption. 
%We provide implementation details that apply to all models in Appendix. 
Finally, in Sec.~\ref{ss:imagenet100}, we apply all of our proposed methods and best practices on our ImageNet100 dataset. These results are shown in Fig.~\ref{fig:exps_imagenet100}, where we observe that we can achieve 80\% top-1 label accuracy with 0.35 annotations per image, which is a 2.7x improvement over ~\cite{Branson_2017_CVPR} and 6.7x improvement over the online DS model. We achieve 87.4\% top-1 label accuracy with 0.98 annotations per image at convergence, compared to 79.5\% and 46.9\% in label accuracy for the competing methods. 
To validate our proposed approach in experiments with human annotations, we conduct a small-scale experiment in Sec.~\ref{ss.human_workers}. We show that our approach achieves 91\% top-1 accuracy with 2x fewer annotations compared to the previous work~\cite{Branson_2017_CVPR}.

\subsection{ImageNet100 Sandbox}
\label{ss:imagenet100-sandbox}
Evaluating and ablating multi-class label annotation efficiency at scale requires large datasets with diverse and relatively clean labels. We construct multiple subsets of the ImageNet dataset~\cite{deng2009imagenet} for our experiments. We use 100 classes sampled from the ImageNet label set~\cite{tian2019contrastive} and construct smaller sub-tasks of varying difficulty using the label hierarchy. These tasks range from 20k to 125k images in size. We use the average accuracy of human workers on these datasets (Sec.~\ref{ss:realistic_workers}) as a proxy for their difficulty, with the average human accuracy ranging from 43\% to 76\%. Tab.~\ref{tab:imagenet100_sub_tasks} details the different sub-tasks, the number of images, classes, and average human accuracy. Fig.~\ref{fig:example_images} shows example images from these tasks. For each class, we use 10 images as prototype images to be provided by the requestor. They ground concepts for human annotators and also help with learning and model selection.

\subsection{Simulating Realistic Workers}
\label{ss:realistic_workers}

Prior work~\cite{long2015multi, hua2013collaborative} simulated workers as confusion matrices. Class confusion was modeled with symmetric uniform noise, which can result in over-optimistic performance estimates. Human annotators exhibit \emph{asymmetric} and \emph{structured} confusion \ie, classes get confused with each other differently. Fig.~\ref{fig:exps_unfiorm_structured_noise_pseudolabel} compares the number of annotations per image in simulation using uniform label noise vs. structured label noise that we crowdsource. We see large gaps between the two. This arises particularly when using learnt models in the loop, due to sensitivity to noisy labels coming from structured confusion in the workers.

To efficiently crowdsource a diverse set of worker confusion matrices for ImageNet100, we split the dataset into 6 disjoint subsets using the ImageNet label hierarchy and crowdsource annotations (using HITs on Amazon Mechanical Turk) for each subset. The assumption is that these sets contain most of the confusion internally and are rarely confused with each other, which we also verify. We collect 40 annotations per worker, which gives us a noisy estimated confusion matrix. To simulate a worker, we sample a confusion matrix per subset and smooth it using an affine combination with the average confusion matrix. 

Through additional HITs, we verify that workers confuse classes across our subsets with a very low probability (0.03). Therefore, we uniformly spread a probability mass of 0.03 for class confusion across our subsets. Overall, we crowdsourced a total of 2680 annotations from 67 workers for these statistics. In comparison, a single experiment with 20k images with 0.5 annotation each would need 10k crowdsourced labels. More details are in the Appendix.

%In the following sub-sections, we introduce our proposed improvements that led to Fig.~\ref{fig:exps_imagenet100} with accompanying results.

\subsection{Self-supervised Learnt Features}
\label{ss:self_sup}
%\noindent\textit{Can we use self-supervised learning to improve the classifier during online-labeling?}

With recent advances in self-supervised learning, it is feasible to learn strong image feature extractors that rival supervised learning, using pretext tasks without any labels. This allows learning in-domain feature extractors for annotation tasks, as opposed to using features pre-trained on ImageNet~\cite{Branson_2017_CVPR}. We compare the efficacy of using BYOL~\cite{grill2020bootstrap}, SimCLR~\cite{chen2020simple}, MoCo~\cite{he2019moco}, relative location prediction~\cite{doersch2015unsupervised} and rotation prediction~\cite{gidaris2018unsupervised} learnt on full ImageNet raw images as the feature extractor $\phi$ in Eq.~\ref{eq:learning_in_the_loop}. We compare the performances in online labeling in Fig.~\ref{fig:exps_self_supervised}. Improvements in self-supervised learning consistently improve efficiency for datasets with both fine and coarse-grained labels, with up to 5x improvement at similar accuracy compared to not using a model in the loop. For the rest of the experiments, we use BYOL~\cite{grill2020bootstrap} as our fixed feature extractor.

\subsection{Learning in Online-Labeling is a Semi-Supervised Problem}
\label{ss:semi_sup}
During online-labeling, the goal is to infer true labels for \emph{all} images in the dataset, making learning $\theta$ akin to transductive learning, where the test set is observed and can be used for learning. Thus, it is reasonable to expect efficiency gains if the dataset's underlying structure is exploited by putting the unlabeled data to work, using semi-supervised learning. This has also been demonstrated recently in the related field of active learning~\cite{mittal2019parting}. Eq.~\ref{eq:learning_in_the_loop} accordingly can be modified to,
$$
\bar{\theta} = \arg\min_{\theta} \mathbb{E}_{(x_i, y_i) \sim \textcolor{red}{\{x_i, \bar{y_i}\}_{1:N}}}  H (\bar{y_i}, p(y_i | \phi(x_i), \theta)) 
$$

Various methods have been proposed in the literature to improve semi-supervised learning. Since the feature extractor $\phi$ is fixed in our setting, we adopt the off-the-self semi-supervised learning algorithms that work directly on a feature space. Specifically, we investigate using Pseudolabels \cite{lee2013pseudo} and MixMatch \cite{berthelot2019mixmatch}.

\textbf{Pseudolabeling} refers to using the current model belief on unlabeled data as labels for learning. We use hard pseudo-labels (argmax of belief), and only retain those labels whose highest class probability falls above a threshold $\tau$. We use model predictions from the previous time step to generate pseudo labels and set the threshold $\tau$ to be 0.1.
\begin{align*} 
\bar{\theta} = \arg\min_{\theta} \mathbb{E}_{(x_i, y_i) \sim \{x_i, \bar{y_i} | p(\bar{y_i} | \mathcal{Z}_i) >  1-\tau \} } H (\bar{y_i}, p(y_i | \phi(x_i), \theta))
\label{eq:pseudolabel} 
\end{align*}

\textbf{MixMatch} constructs virtual training examples by mixing the labeled and unlabeled data using a modified version of MixUp \cite{zhang2017mixup}. We modify MixMatch for our use-case and provide details in the Appendix.
% The labeled set is defined by the data points with at least one worker annotation, $\mathopen|\mathcal{W}_i\mathclose| > 0$, and the unlabeled set is defined by the data points whose largest probability is larger than a predefined threshold and is not in the labeled set. 
% We use the same threshold as the one used in pseudo labels.
% The mixmatch loss consists of cross entropy of the labeled set and the l2 minimization of the mixed set.
% The mixed set is constructed by sampling $(x_1, p_1)$ from the labeled set and $(x_2, p_2)$ from the unlabeled set and interpolate both input and output.
% \begin{align} 
%  \begin{split}
%  \lambda & \sim Beta(\alpha, \alpha) \\
%  \lambda' &= \max(\lambda, 1- \lambda) \\
%  x' &= \lambda' x_1 + (1 - \lambda') x_2 \\
%  p' &= \lambda' p_1 + (1 - \lambda') p_2 \\
%  \mathcal{S}_{\text{mixed}} & \leftarrow (x', p') \\ 
%  L &= \mathbb{E}_{(x, y) \sim \{x_i, \bar{y_i} | \mathopen|\mathcal{W}_i\mathclose| > 0\}} H( \bar{y_i} , p(y | \phi(x_i), \theta)) \\
%  &+  \mu \mathbb{E}_{(x_i, p_i) \sim \mathcal{S}_{\text{mixed}}} \left\lVert p_i - p(y | \phi(x_i), \theta) \right\rVert^2_2
%  \end{split}
%  \end{align}
 
%  \noindent where $\mu$ is the hyperparameters. 
 
 Fig.~\ref{fig:exps_semi_supervised} shows results using PseudoLabels and MixMatch during learning. Semi-supervised learning consistently improves annotation efficiency across datasets, with more pronounced improvements shown in coarse-grained datasets. 
 For \textit{Dog}, while all methods fail to reach 80\% accuracy due to the poor quality of worker annotations, using PseudoLabels reaches the performance of its counterpart at convergence with 30\% fewer annotations.
 %For coarse-grained datasets such as Commodity, we find that using self-supervised features alone already reaches 80\% accuracy with only 0.15 annotations per image, leading to smaller observed gains with  semi-supervised learning.

 Despite its simplicity and other published results, we surprisingly find Pseudolabeling performs better than MixMatch in our case. We remind the reader that for online-labeling, the inputs to the semi-supervised learnt models are the feature vectors $\phi(x)$ instead of raw images. Therefore, various data augmentation strategies applied at an image level cannot be applied directly, which we hypothesize to be the reason for the observed trend. In the rest of the experiments, we use Pseudolabels by default.
 
\begin{figure}[t!]
    \centering
    \includegraphics[height=4cm]{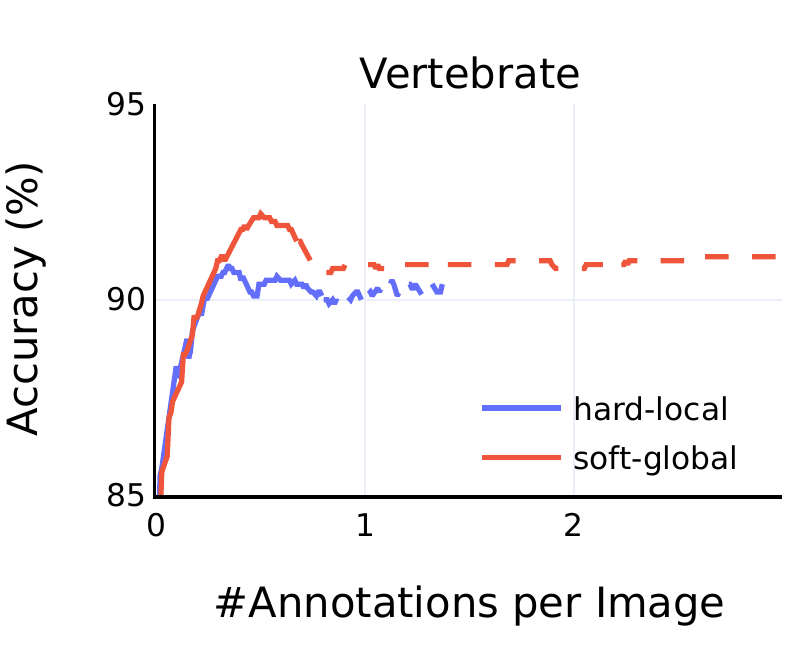}
     \vspace{-4mm}
    \caption{\footnotesize \textbf{Instability in model parameter learning (gradient descent) and the label aggregation (EM)} during semi-supervised learning can lead to a negative feedback loop. Using two simple heuristics, \textbf{global} model selection and \textbf{soft} convergence in EM, avoids this issue. The dashed lines show results beyond our early stopping criterion (Sec.~\ref{subsubsec:when_to_stop})}
    \label{fig:exps_semi_pitfall}
     \vspace{-4mm}
\end{figure}

\textbf{Semi-supervised learning during online-labeling can be unstable:}
Learning in the loop helps provide a better prior for label-aggregation (Eq.~\ref{eq:label_aggregation}) while also improving worker skill inference (Eq.~\ref{eq:worker_skill_inference}). We find that alternate maximation of these two objectives can lead to a negative feedback cycle once either of them converges to a bad solution, often resulting in divergence. It is known that EM optimization does not always converge to a fixed point, with the likelihood function of the iterates likely to grow unbounded~\cite{mclachlan_krishnan_2008}. Using semi-supervised learning exacerbates this issue since wrong confident intermediate labels can become more confident through subsequent time-steps. Fig.~\ref{fig:exps_semi_pitfall} shows this divergence behavior. We mitigate these with simple heuristics related to EM convergence and model selection through time. During EM, there are two popular ways to determine convergence: 1) Hard Constraint: stopping if the E-step does not change across steps, 
%\ie, $p(y_i^t | z_i^t) = p(y_i^{t-1} | z_i^{t-1}), \forall i \in [N]$. 
and 2) Soft Constraint: stopping if the average likelihood of the observation changes minimally across steps. %\ie, $\sum_{i \in [N]} \sum_{j \in \mathcal{W}_i} p(z_i^t | y_i^t, w_j^t) - \sum_{i \in [N]} \sum_{j \in \mathcal{W}_i} p(z_i^{t-1} | y_i^t, w_j^{t-1})$. 
%  \AL{I want to mention that usually the EM diverge since the surface of the likelihood function looks like a ridge so it goes steady for awhile and suddenly diverges.}
% Move to supplementary
\begin{align} 
\text{Hard: } & \arg\max p(y_i^t | z_i^t) = \arg\max p(y_i^{t-1} | z_i^{t-1}), \forall i  \\
 \begin{split}
\text{Soft: } & \mathopen| \frac{\sum_{i \in [N]} \sum_{j \in \mathcal{W}_i} p(z_i^t | y_i^t, w_j^t)}{\mathopen| \mathcal{Z}^{t} \mathclose|} - \\
& \frac{\sum_{i \in [N]} \sum_{j \in \mathcal{W}_i} p(z_i^{t-1} | y_i^{t-1}, w_j^{t-1})}{\mathopen| \mathcal{Z}^{t-1} \mathclose|} \mathclose|\leq \epsilon
\label{eq:em_soft_constraint}
\end{split}
\end{align}
We find that adopting the latter mitigates divergence during EM with $\epsilon = 0.01$. Moreover, we perform model selection across time-steps \ie, do not replace the model if its validation performance does not improve. We use a fixed validation set comprised of the prototype images across steps, ablated in Sec.~\ref{ss:practical}. We show the efficacy of applying the proposed heuristics in Fig.~\ref{fig:exps_semi_pitfall}. 
 
\begin{figure*}[t!]
\scriptsize
 \begin{adjustbox}{width=0.62\linewidth}
  \begin{subfigure}[b]{0.5\textwidth}
      \centering
      \parbox{0.21\linewidth}{\centering Coyote}
      \parbox{0.21\linewidth}{\centering Honeycomb}
      \parbox{0.21\linewidth}{\centering Bottlecap}
      \parbox{0.21\linewidth}{\centering Blue Heron}
      \includegraphics[width=0.21\linewidth,height=1.6cm]{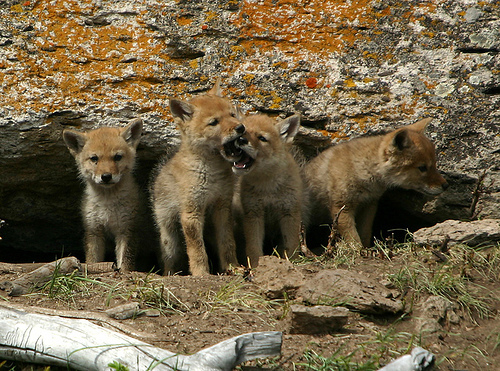}
      \includegraphics[width=0.21\linewidth,height=1.6cm]{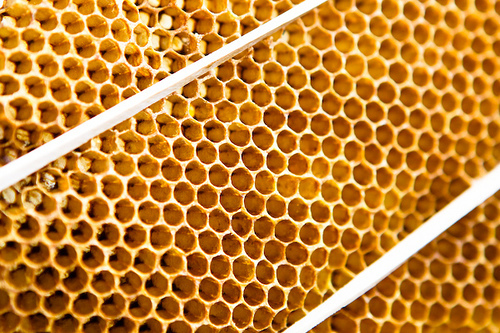}
      \includegraphics[width=0.21\linewidth,height=1.6cm]{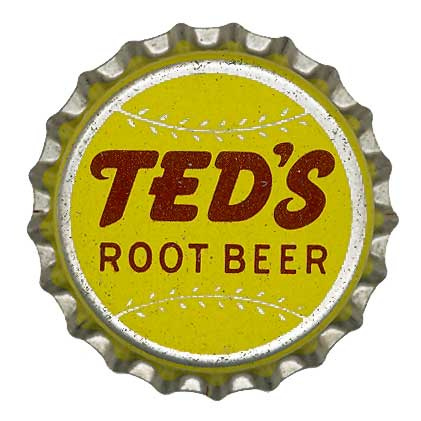}
      \includegraphics[width=0.21\linewidth,height=1.6cm]{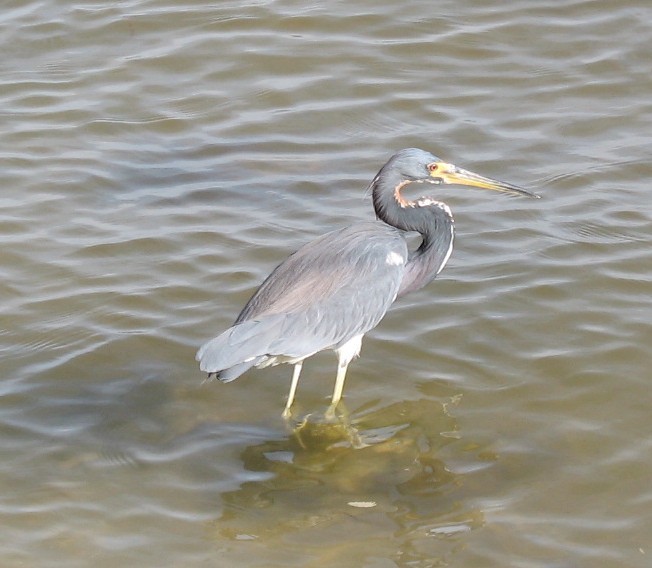} 
      \parbox{0.21\linewidth}{\centering Cauliflower}
      \parbox{0.21\linewidth}{\centering Harmonica}
      \parbox{0.21\linewidth}{\centering Window Screen}
      \parbox{0.21\linewidth}{\centering Pirate}
      \includegraphics[width=0.21\linewidth,height=1.6cm]{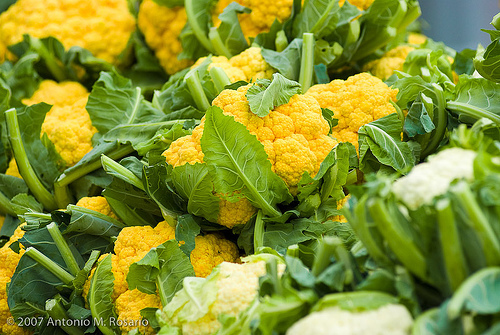}
      \includegraphics[width=0.21\linewidth,height=1.6cm]{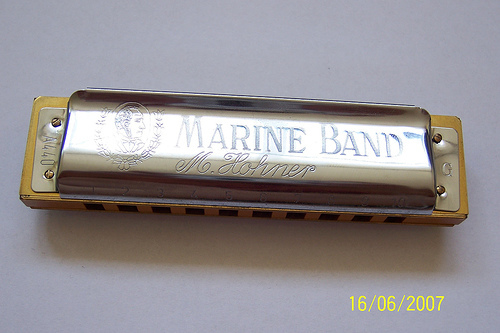}
      \includegraphics[width=0.21\linewidth,height=1.6cm]{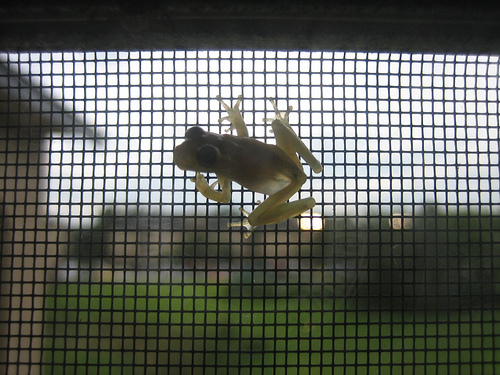}
      \includegraphics[width=0.21\linewidth,height=1.6cm]{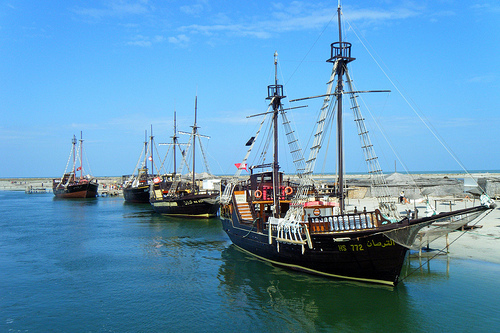} 
   %   \parbox{0.21\linewidth}{\centering Walker Hound}
    %  \parbox{0.21\linewidth}{\centering Carbonara}
    %  \parbox{0.21\linewidth}{\centering Leafhopper}
    %  \parbox{0.21\linewidth}{\centering Pirate}
    %  \includegraphics[width=0.21\linewidth,height=1.6cm]{assets/imagenet100_qualitative/n02089867/n02089867_802.JPEG}
    %  \includegraphics[width=0.21\linewidth,height=1.6cm]{assets/imagenet100_qualitative/n07831146/n07831146_209.JPEG}
    %  \includegraphics[width=0.21\linewidth,height=1.6cm]{assets/imagenet100_qualitative/n02259212/n02259212_5429.JPEG}
    %  \includegraphics[width=0.21\linewidth,height=1.6cm]{assets/imagenet100_qualitative/n03947888/n03947888_45822.JPEG}\\[-1.2mm]
    
     \vspace{-1mm}
      \caption{\large Images with no annotations}
      \label{fig:imagenet100_no_annotation}
  \end{subfigure}
  \hspace{-10mm}
  \begin{subfigure}[b]{0.5\textwidth}
      \centering
      \parbox{0.21\linewidth}{\centering Boxer}
      \parbox{0.21\linewidth}{\centering Cocktail Shaker}
      \parbox{0.21\linewidth}{\centering Kuvasz}
      \parbox{0.21\linewidth}{\centering Milk Can}
      \includegraphics[width=0.21\linewidth,height=1.6cm]{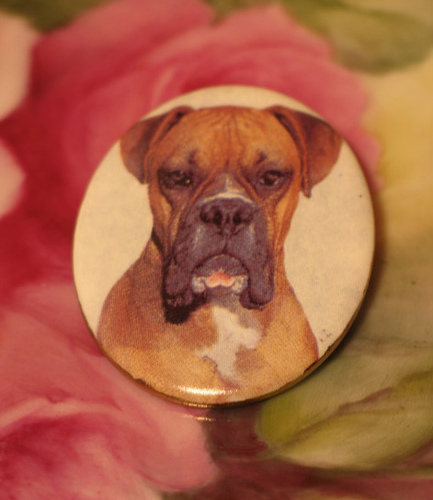}
      \includegraphics[width=0.21\linewidth,height=1.6cm]{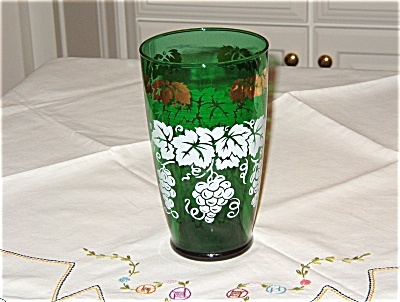}
      \includegraphics[width=0.21\linewidth,height=1.6cm]{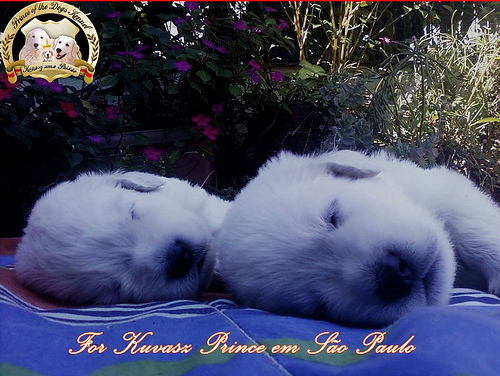}
      \includegraphics[width=0.21\linewidth,height=1.6cm]{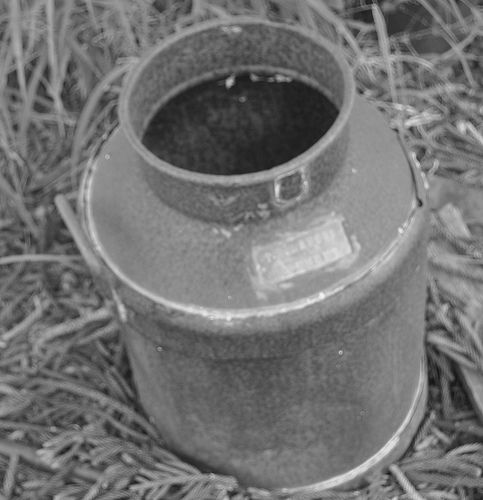} 
      \parbox{0.21\linewidth}{\centering Bottlecap}
      \parbox{0.21\linewidth}{\centering American Lobster}
      \parbox{0.21\linewidth}{\centering Mexican Hairless}
      \parbox{0.21\linewidth}{\centering Walking Stick}
      \includegraphics[width=0.21\linewidth,height=1.6cm]{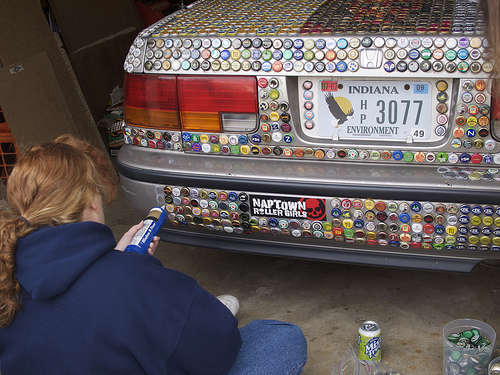}
      \includegraphics[width=0.21\linewidth,height=1.6cm]{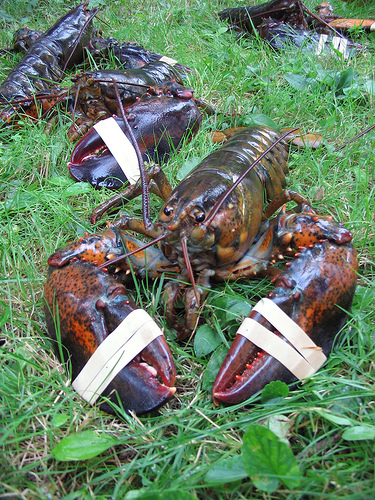}
      \includegraphics[width=0.21\linewidth,height=1.6cm]{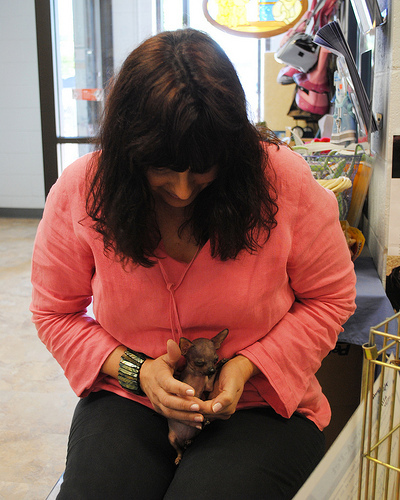}
      \includegraphics[width=0.21\linewidth,height=1.6cm]{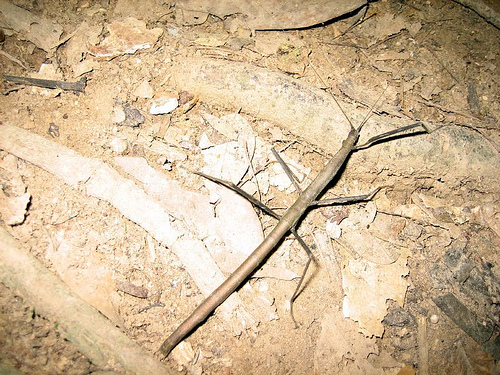} 
    % \parbox{0.24\linewidth}{\centering Computer Keyboard}

     % \parbox{0.24\linewidth}{\centering Cocktail Shaker}
      %\parbox{0.24\linewidth}{\centering Kuvasz}
     % \parbox{0.24\linewidth}{\centering Modem}
      %\includegraphics[width=0.24\linewidth,height=1.5cm]{assets/imagenet100_qualitative/n03085013/n03085013_3632.JPEG}
      %\includegraphics[width=0.24\linewidth,height=1.5cm]{assets/imagenet100_qualitative/n03062245/n03062245_2416.JPEG}
      %\includegraphics[width=0.24\linewidth,height=1.4cm]{assets/imagenet100_qualitative/n02104029/n02104029_2037.JPEG}
      %\includegraphics[width=0.24\linewidth,height=1.4cm]{assets/imagenet100_qualitative/n03777754/n03777754_8874.JPEG}\\[-1.2mm]
      \vspace{-1mm}
      \caption{\large Images with 3 annotations}
      \label{fig:imagenet100_3annotation}
  \end{subfigure}
  \end{adjustbox}
 %   \end{minipage}
  %\vspace{-2mm}
  \hfill
  \begin{minipage}{0.38\linewidth}
  \vspace{-24mm}
  \caption{\footnotesize \textbf{Example images with different numbers of annotations under our full framework on ImageNet100.} Images without annotations (left) usually contain only one centered object. Images with 3 annotations (right) either have multiple objects (Bottlecap), a small object of interest (Mexican Hairless, Walking Stick), ambiguous objects (Boxer, Cocktail Shaker), and some not in their typical state (Kuvasz puppies). We also find simple examples (Milk Can) with 3 annotations, pointing to room for improvement.}
  \label{fig:imagenet100_qualitative_results}
  \end{minipage}
  \vspace{-6mm}
\end{figure*}

\subsection{Practical Considerations}
\label{ss:practical}
We ablate several design choices involved in implementing such an online-labeling system, leading us to identify guidelines for future practitioners. We find these design choices can significantly affect efficiency results, discussed sequentially in the following sections.

\vspace{-3mm}
\subsubsection{Generating calibrated model likelihoods}
\begin{figure*}[t!]
\begin{minipage}{0.32\linewidth}
    \centering
    \includegraphics[height=3cm]{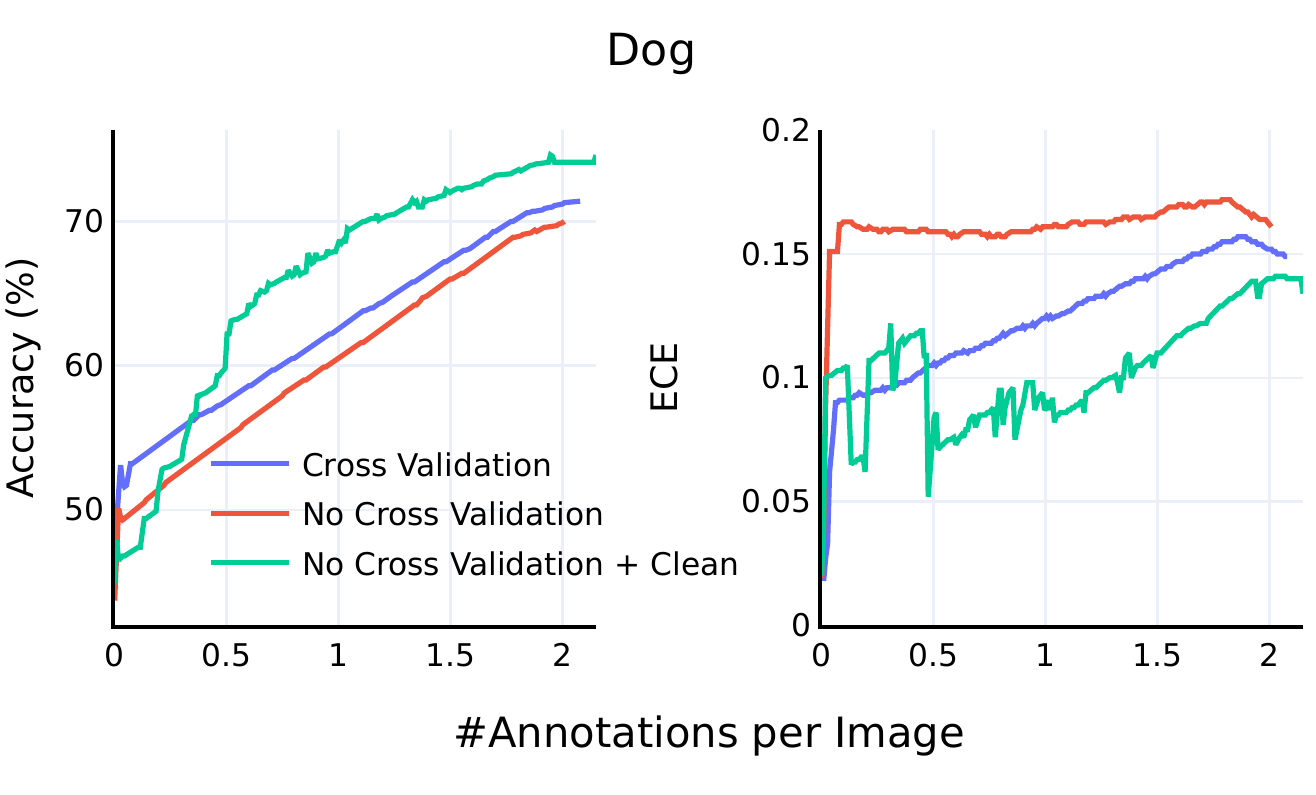}
     \vspace{-4mm}
    \caption{\footnotesize \textbf{Validation set selection and calibration.} We find using a clean validation set is more important than performing cross-validation.} 
    \label{fig:exps_calibration}
    \end{minipage}
    \hspace{2mm}
    \begin{minipage}{0.32\linewidth}
        \centering
    \includegraphics[height=3cm]{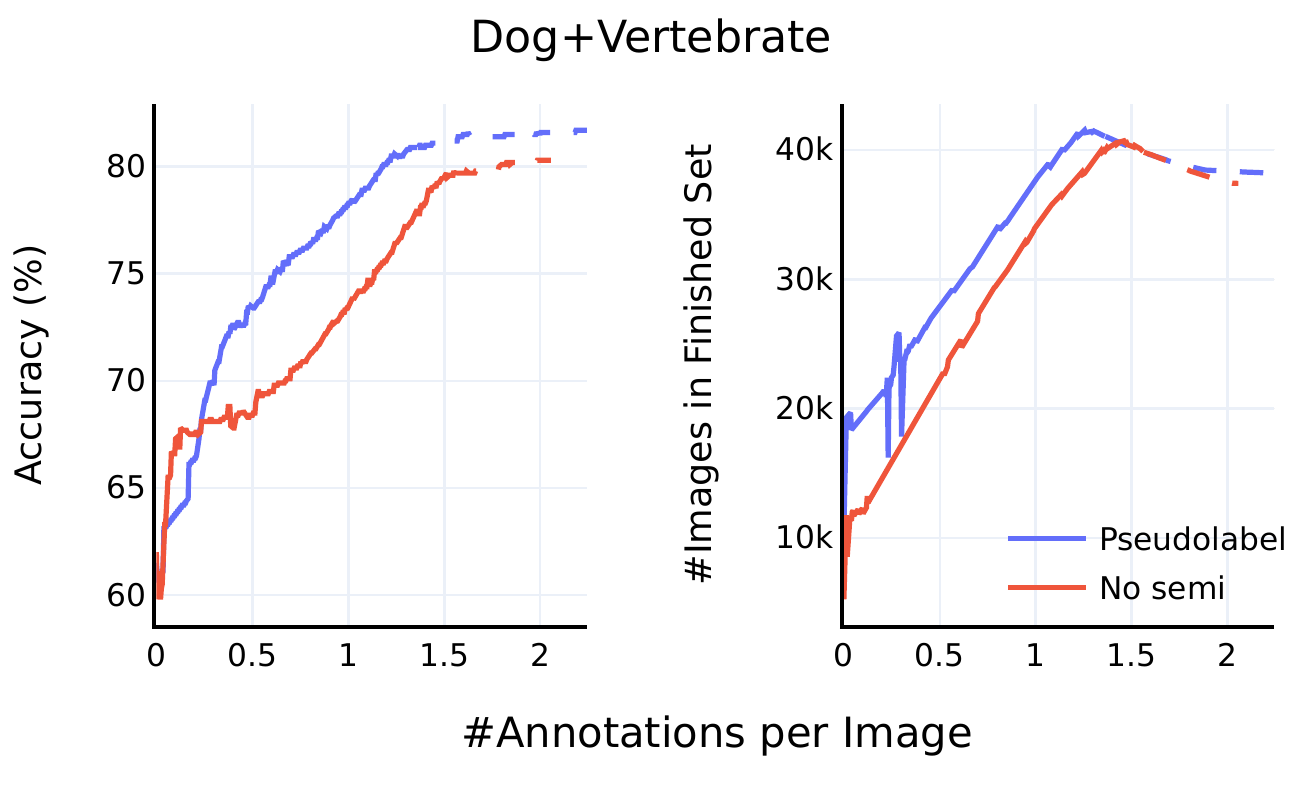}
     \vspace{-4mm}
    \caption{\footnotesize \textbf{Early-stop} by monitoring the size of the finished set. This avoids over-sampling for confusing images. Dashed lines represent trajectories using stopping criterion from ~\cite{Branson_2017_CVPR}.}
    \label{fig:exps_when_to_stop}
      \end{minipage}
      \hspace{2mm}
       \begin{minipage}{0.33\linewidth}
          \centering
    \vspace{-1mm}
    \includegraphics[height=3cm]{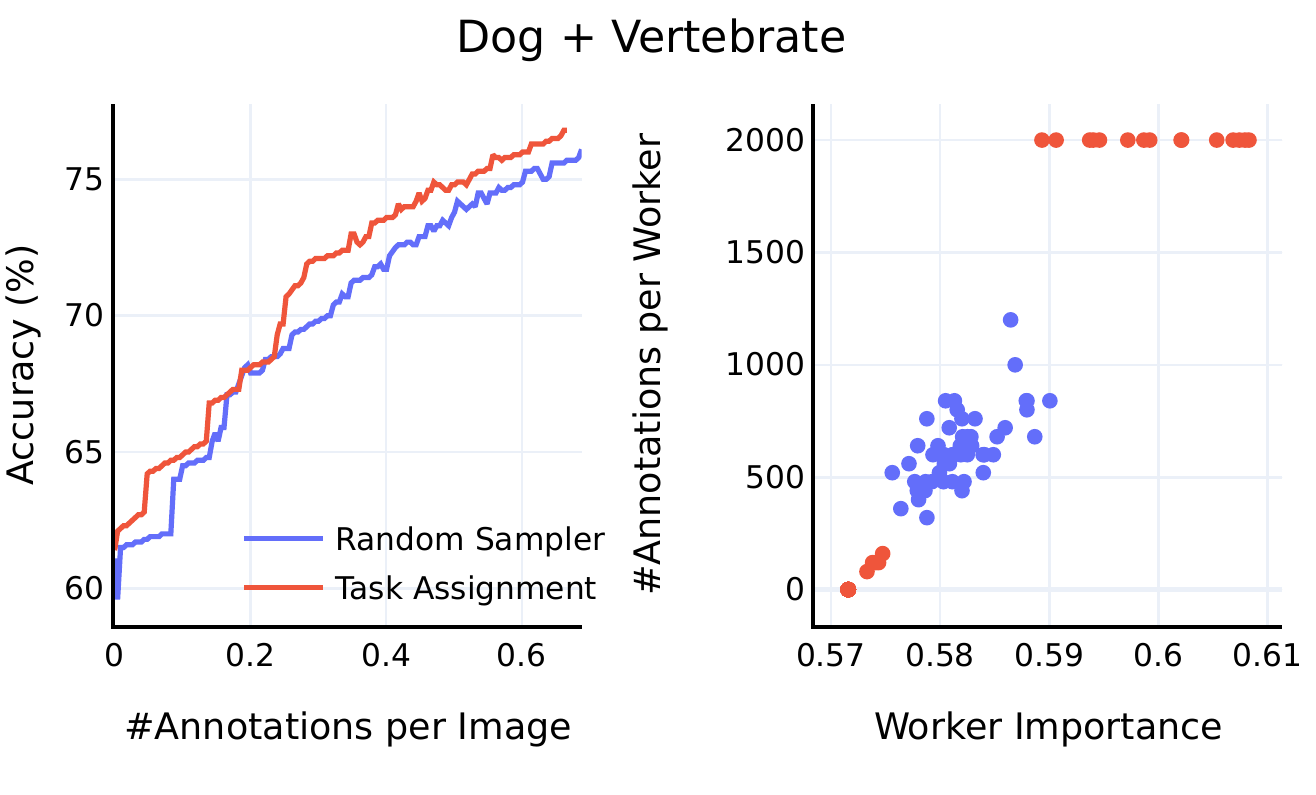}
     \vspace{-3mm}
    \caption{\footnotesize \textbf{Task Assignment.} We adopt a simple greedy task assignment scheme using learnt worker skills. We show that the learnt skills help assign more tasks to ``important'' workers.}
    \label{fig:exps_task_assignment}
      \end{minipage}
     \vspace{-4mm}
\end{figure*}

\vspace{-1mm}
\label{subsubsec:calibration}
Prior work~\cite{Branson_2017_CVPR} uses a modified cross-validation approach to generate model likelihoods. They ensure that the estimated prior $p(y_i | x_i, \bar{\theta})$ is produced using a model that was not trained or validated on labels from image $i$. We find that this can underperform when estimated labels are noisy, which pollutes validation splits and makes calibration challenging. Instead, we propose to use the clean prototype images as the validation set. We ablate the importance of having clean validation and performing cross-validation in Fig.~\ref{fig:exps_calibration}. We find that having a clean validation set is more important than using cross-validation. All our experiments use temperature calibration~\cite{guo2017calibration} with a clean validation set comprised of prototype images to be provided by the task requestor.

\vspace{-3mm}
\subsubsection{Model update frequency}
\vspace{-1mm}
%How often should one update the model vs. collect human labels? The latency of model updates vs. collection dictates varies across applications. We investigate how model update frequency affects results in Fig.~\ref{fig:model_update_frequency}. For fine-grained datasets, lower update frequencies (higher number of annotations per update) tend to overshoot in number of annotations, while the method is robust to low update frequency on coarse-grained datasets.
How often should one update the model vs. collect human labels? The latency of model updates vs. collection dictates varies across applications. We find that for fine-grained datasets, lower update frequencies (higher number of annotations per update) tend to overshoot in the number of annotations, while the method is robust to low update frequency on coarse-grained datasets. See more details in the Appendix.

\vspace{-3mm}
\subsubsection{Tuning Hyperparameters}
\vspace{-1mm}
Tuning hyperparameters is difficult when the task is to label data itself. We split our prototypes (10 per class) into train and validation sets to tune hyperparameters. Though there are few prototypes, we find that this works almost as well as tuning with backdoor label access. The requestor can always add more images to the prototype set from the up-to-date annotations for improved hyperparameters search.

\vspace{-3mm}
\subsubsection{Pre-identifying Worker Skills}
\vspace{-1mm}
We explore to leverage class-dependent and worker-dependent priors. In reality, the requestor can ask gold standard questions or apply prior knowledge to design the prior.
In our experiment, we find that this is especially useful for fine-grained datasets. The best explored prior improves accuracy by 15\% in \textit{Dog}, while in \textit{Commodity}, the improvement is marginal. See more details in the Appendix.

\vspace{-3mm}
\subsubsection{Task Assignment with Inferred Skills}
\vspace{-1mm}
There are certain particularly hard classes, with only a few workers having enough expertise to annotate them correctly.  We ask whether the learnt skills can be used to assign tasks better. Prior work on (optimal) task assignment tackle crowdsourcing settings with vastly different simplifying assumptions~\cite{ho2013adaptive, chen2015statistical}, and designing a new task assignment scheme is out of the scope of this paper.
%To verify if the learnt worker skills help with task assignment, we propose a simple greedy algorithm. For an image sampled for annotation from the unfinished set $\mathcal{U}$, we use $\bar{y_i}$ to find the best possible workers from the workers pool using the currently estimated worker skill, with a cap on the maximum number of annotations $M$ allowed per worker.
To verify if the learnt worker skills help with task assignment, we propose a simple greedy algorithm with a cap on the maximum number of annotations $\alpha$ allowed per worker. 
%\vspace{-1mm}
%\begin{align} 
%    j \leftarrow \arg\max_{j \in \hat{\mathcal{W}}} \mathbb{E}_{y \sim \bar{y_i}} \bar{w_j}[y, y], \hat{\mathcal{W}}=\{j | \mathopen|\mathcal{Z}_j\mathclose| \leq \alpha\}
%    \label{eq:task_assignment}
%\end{align}
%\vspace{-2mm}

Fig.~\ref{fig:exps_task_assignment} shows results with task assignment with $\alpha=2000$. The simple task assignment allows saving 13\% of annotations to reach 75\% label accuracy. 
%On the right, we show the distribution between worker importance, measured by a weighted sum of the worker's per-class reliability and the number of annotations assigned to the worker. The weights are inversely proportional to the model's per-class accuracy. Ideally, number worker annotations would be highly correlated with worker importance. The results show that, while not perfect, the learnt skills indeed help assign more tasks to important workers.
On the right, we show the distribution between worker importance. Ideally, number worker annotations would be highly correlated with worker importance. The results show that, while not perfect, the learnt skills indeed help assign more tasks to important workers.

\vspace{-3mm}
\subsubsection{Number of Workers} 
\vspace{-1mm}
%We explore how annotation efficiency is affected with different number of simulated workers in Fig.~\ref{fig:num_workers}. We find that having lesser workers usually results in better label quality, while the convergence point with respect to total number of annotations does not change. Higher label quality is expected since individual worker skills are better identified. 
We explore how annotation efficiency is affected by the different number of simulated workers. With the fixed number of annotations, we find that having more workers hurts the performance due to the fewer observations of each worker, resulting in poor worker skill estimation. On the contrary, having fewer workers significantly helps in fine-grained datasets. Reducing the number of workers from 50 to 10 improves the accuracy by 17\% in \textit{Dog}. See more details in the Appendix.

\vspace{-3mm}
\subsubsection{Pre-Filtering Datasets}
\label{sss.pre_filtering}
\begin{figure}[t!]
\begin{minipage}{0.47\linewidth}
    \centering
    \includegraphics[height=3.2cm]{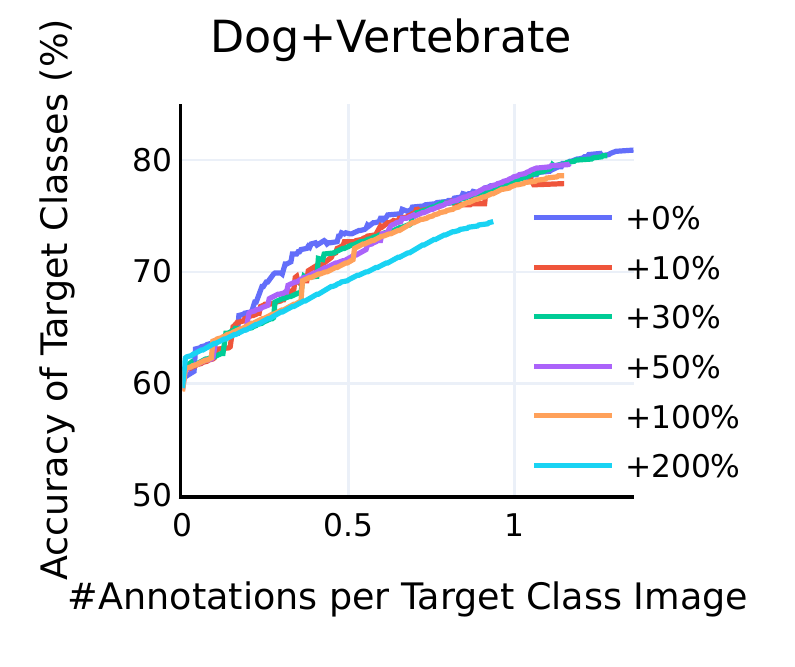}
     %\vspace{-4mm}
     \caption{\footnotesize \textbf{Dataset contains OOD images.} Our method retains efficiency till around 100\% out-of-distribuion images. (Sec.~\ref{sss.pre_filtering})}
     \label{fig:exps_distracting_images}
\end{minipage}
\hspace{2mm}
\begin{minipage}{0.47\linewidth}
    \centering
    \includegraphics[height=3.2cm]{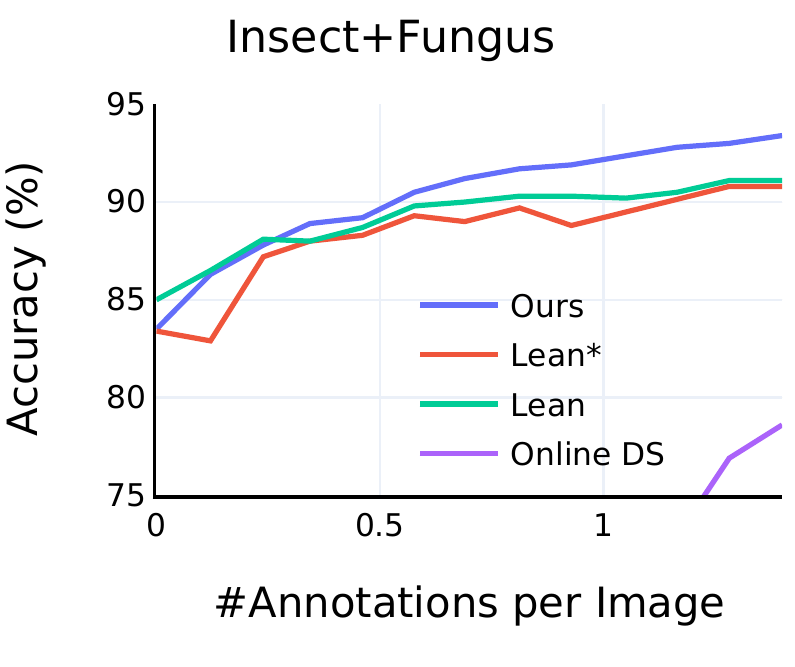}
     %\vspace{-4mm}
     \caption{\footnotesize \textbf{Transfer to human workers.} Our method increase efficiency w.r.t. previous work~\cite{Branson_2017_CVPR} even when using human annotations, achieving 91\% accuracy with 2x reduction. (Sec.~\ref{ss.human_workers})}
     \label{fig:human_workers}
\end{minipage}
\end{figure}

\vspace{-1mm}
We have assumed that the requestor performs perfect filtering before annotation, \ie, all the images to be annotated belong to the target classes, which does not always hold. We add an additional ``None of These'' class and ablate annotation efficiency in the presence of unfiltered images. We include different numbers of images from other classes and measure the mean precision with the number of annotations of the target classes. In Fig.\ref{fig:exps_distracting_images}, we see that even with 100\% more images from irrelevant classes, comparable efficiency can be retained on a fine-grained dataset.

\vspace{-3mm}
\subsubsection{Risk Threshold}
\vspace{-1mm}
The risk threshold $C$ in online labeling determines how fast the annotation process converges and the possible final label quality. As expected, having lower risk threshold results in slower convergence, but improves final label quality. We use $C=0.1$ for all other experiments in this paper. See more details in the Appendix.

% \begin{figure}[t!]
%     \centering
%     \includegraphics[width=0.85\linewidth, height=3.cm]{assets/images/plot_risk_change.pdf}
%      \vspace{-4mm}
%     \caption{}
%     \label{fig:risk_changes}
%      \vspace{-1mm}
% \end{figure}

%\begin{figure}[t!]
%    \centering
%    \includegraphics[width=0.82\linewidth,height=3cm]{assets/images/plot_step_size_change.pdf}
%     \vspace{-4mm}
%    \caption{\footnotesize Comparing efficiency change by varying the frequency of model updates. For fine-grained datasets, very low frequency of update can lead to overshooting. The frequency does not seem to matter for coarse-grained data in our experiments.}
%    \label{fig:model_update_frequency}
%     \vspace{-4mm}
%\end{figure}

%\begin{figure}[t!]
%    \centering
%    \includegraphics[width=0.82\linewidth,height=3cm]{assets/images/plot_worker_change.pdf}
%    \vspace{-4mm}
%    \caption{ \footnotesize Lower number of workers usually results in greater annotation quality but does not necessarily improve efficiency.}
%    \label{fig:num_workers}
%    \vspace{-4mm}
%\end{figure}

\vspace{-3mm}
\subsubsection{When to Stop Annotating?}\label{subsubsec:when_to_stop}
\vspace{-1mm}
A clear criterion to stop annotation is when the unfinished set of images (images with estimated risk greater than a threshold) is empty~\cite{Branson_2017_CVPR,van2018lean}. However, we observe that the annotation accuracy usually saturates and then grows slowly because of a small number of data points that are heavily confused by the pool of workers used. Therefore we suggest that the requestor 1) stop annotation at this time and separately annotate the small number of unfinished samples, possibly with expert annotators, and 2) set a maximum number of annotations per image — we use 3 in our paper. However, how do we automatically decide when to stop without access to true labels? We find that performing early stopping on the size of the finished set of images is sufficient, as shown in Fig.~\ref{fig:exps_when_to_stop}. If the finished set size does not increase from its global maximum value for $\beta$ consecutive steps, we stop annotation. We adopt this stopping criterion and set $\beta=5$ for all other experiments in this paper. 
%\textcolor{red}{Despite the effectiveness of the early stop, one might have concern that it drops the images that are crucial to the model accuracy. We train a ResNet18 from scratch using labels collected with our method with different budgets or using ImageNet labels. Tab~\ref{} shows the validation accuracy of these models on ImageNet, along with the label accuracy in brackets. ... More details can be found in the Appendix.}

\subsection{Putting it Together on ImageNet100}
\label{ss:imagenet100}
We finally compare our full framework (Algo~\ref{algo:efficient_annotation}), with ~\cite{Branson_2017_CVPR} (Lean), an adapted version with prototypes as validation set and without cross-validation (Lean*) and the online DS model in Fig.~\ref{fig:exps_imagenet100}. Dashed lines represent results of removing the stopping criterion mentioned in Sec.~\ref{subsubsec:when_to_stop}. Our framework consistently provides higher accuracy and stable improvement over time. We achieve nearly 87\% top-1 label accuracy on a 100 class subset of ImageNet with only 0.98 annotations per image and 80\% top-1 label accuracy with 0.35 annotation per image, a 2.5x reduction reduction w.r.t. ``Lean*'', 2.7 reduction w.r.t ``Lean'' and a 6.7x reduction over ``Online DS''. 

In Fig.~\ref{fig:imagenet100_qualitative_results}, we visualize images with 0 and 3 annotations received, respectively, along with their ground truth ImageNet label. We find that images that got 0 annotations usually contain only one clear centered object. Images with 3 annotations sometimes have multiple objects (Bottlecap), or have a small object (Walking Stick, Mexican Hairless) or an ambiguous object (Boxer, Cocktail Shaker). We also note that there are simple images that receive 3 annotations (Milk Can), showing room for improvement.

\subsection{Transfer to Human Workers}
\label{ss.human_workers}
To validate if these good practices apply outside our simulation and in the real world, we collect 3088 annotations of 1878 images from \textit{Insesct+Fungus}. Fig.~\ref{fig:human_workers} shows that our method works well when transferring to human workers, achieveing 91\% accuracy with 2x fewer required annotations w.r.t. previous work~\cite{Branson_2017_CVPR}.
%\AL{Number needs to be confirmed.}
More details in the Appendix.

% \subsection{Implementation Details}

% In this work, we use a one hidden layer fully connected layer as the classification head. The hidden size is 32 and the batch size is 1024. On ImageNet100, we use a batch size of 8192. Learning rate and weight decay are found via grid search on prototypes. 
% All self-supervised features~\cite{zhan2020online} use ResNet50~\cite{he2016deep} as backbone.

\setlength{\textfloatsep}{0pt}
\begin{algorithm}[t!]
\DontPrintSemicolon
  \scriptsize
  \KwInput{Unlabeled images $\mathcal{X}=\{x_i\}_{i=1}^{N}$ and workers $\mathcal{W}=\{w_j\}_{j=1}^M$}
  \KwOutput{labels $\mathcal{Y}=\{y_i\}_{i=1}^N$}
  Set unfinished set $\mathcal{U}=\{i\}_{i=1}^N$, finished set $\mathcal{F} = \emptyset$, $loss^*=\inf$, and $\bar{\theta^*}$ is randomly initialized
    
  $\bar{\theta} \leftarrow \bar{\theta^*}$
  
  $\phi \leftarrow \text{Self-supervised learning on } \mathcal{X}$
  
  \While{stopping criterion~\ref{subsubsec:when_to_stop} is not met}
   {
   		Construct $B$ HITs from $\mathcal{U}$, sample $B$ workers
   		
   		Obtain annotations $\mathcal{Z}$
   		       
   		Initialize worker skills $\bar{\mathcal{W}}$

        \While{Eq.~\ref{eq:em_soft_constraint} is not met}
        {
        	$\bar{\mathcal{Y}} \leftarrow$ Aggregate $\mathcal{Y}$ by Eq.~\ref{eq:label_aggregation}
        	
        	$\bar{\mathcal{W}} \leftarrow$ Maximize $\mathcal{W}$ by Eq.\ref{eq:worker_skill_inference}
        	
        }

        $\bar{\theta} \leftarrow \text{Model Parameter learning by Eq.~\ref{eq:pseudolabel}}$ 
        
        $\bar{\theta} \leftarrow \text{Calibrate with prototypes~\ref{subsubsec:calibration}}$
        
        $loss \leftarrow \text{Measure loss on prototypes}$ 
        
        \If{$loss \leq loss^*$}
        {
        	$\bar{\theta^*} \leftarrow \bar{\theta}$;\quad $loss^* = loss$
        }
        \Else
        {
        	$\bar{\theta} \leftarrow \bar{\theta^*}$

        }
        
        \For{ $i \in \{i\}_{1:N}$}
        {
        	if $\mathcal{R}(\bar{y_i}) < C$: $\mathcal{F} \leftarrow \mathcal{F} \cup i$, $\mathcal{U} \leftarrow \mathcal{U} \setminus i$
        }
        
   }
   \Return{$\mathcal{Y}$}
\caption{Efficient Annotation}
\label{algo:efficient_annotation}
\end{algorithm}

% \textbf{Which ML model to use:}
% We evaluate parametric (SVM (lean crowdsourcing), small MLP) and non-parametric (LP)
% methods for accuracy, calibration error, time-to-train and complexity of hyperparameter
% selection. Stress the importance of calibration error here.
\section{Discussion and Conclusion}
We presented improved online-labeling methods for large multi-class datasets. In a realistically simulated experiment with 125k images and 100 labels from ImageNet, we observe a 2.7x reduction in annotations required w.r.t. prior work to achieve 80\% top-1 label accuracy. Our framework goes on to achieve 87.4\% top-1 accuracy at 0.98 labels per image. Along with our improvements, we leave open questions for future research. 1) Our simulation is not perfect and does not consider individual image difficulty, instead only modeling class confusion. 2) How does one accelerate labeling beyond semantic classes, such as classifying viewing angle of a car? 3) ImageNet has a clear label hierarchy, which can be utilized to achieve orthogonal gains~\cite{van2018lean} in the worker skill estimation. 4) Going beyond classification is possible with the proposed model by appropriately modeling annotation likelihood as demonstrated in~\cite{Branson_2017_CVPR}. However, accelerating these with learning in the loop requires specific attention to detail per task, which is an exciting avenue for future work. 5) Finally, we discussed annotation at scale, where improvements in learning help significantly. How can these be translated to small datasets? We discuss these questions more in the Appendix, and release a codebase to facilitate further research in these directions.

\textbf{Acknowledgments:}
This work was supported by ERA, NSERC, and DARPA XAI. SF acknowledges the Canada CIFAR AI Chair award at the Vector Institute.

{\small
%\bibliographystyle{ieee_fullname}
%\bibliography{egbib}

}

\clearpage
\appendix
\onecolumn
\section{More Ablations}

\subsection{Risk Threshold}
\begin{figure}[h]
  \centering
  \includegraphics[width=0.8\textwidth]{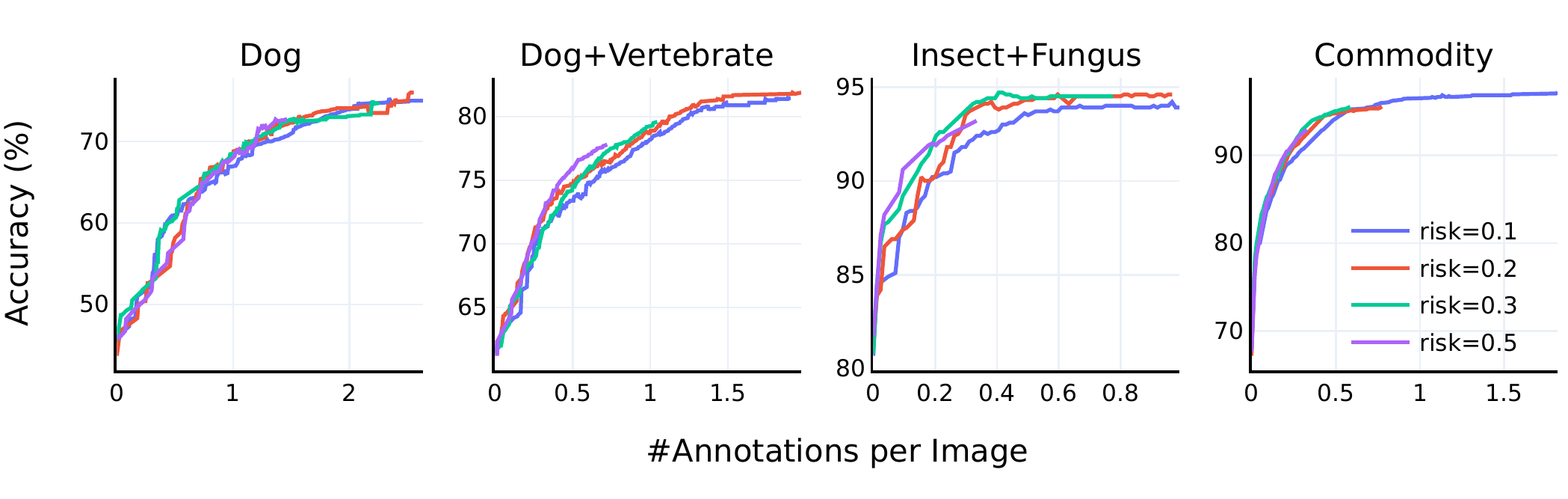}
  \caption{\textbf{Ablation of different risk thresholds.} Having lower risk threshold results in slower convergence, and improved final label quality.}
  \label{sup:fig:risk_changes}
\end{figure}

The risk threshold $C$ in online annotation determines how fast the annotation process converges and possibly affects the final label quality. In Fig.\ref{sup:fig:risk_changes}, we find that using a lower risk threshold stably results in stable label quality improvement. For larger risk threshold, some incorrectly annotated data can be considered confident in the early stage, resulting in unstable label quality improvement. It is somewhat surprising that varying the risk threshold does not affect the final label quality a lot. We set $C=0.1$ for all other experiments in this paper.

\subsection{Number of Workers} 
\begin{figure}[h]
  \centering
  \includegraphics[width=0.8\textwidth]{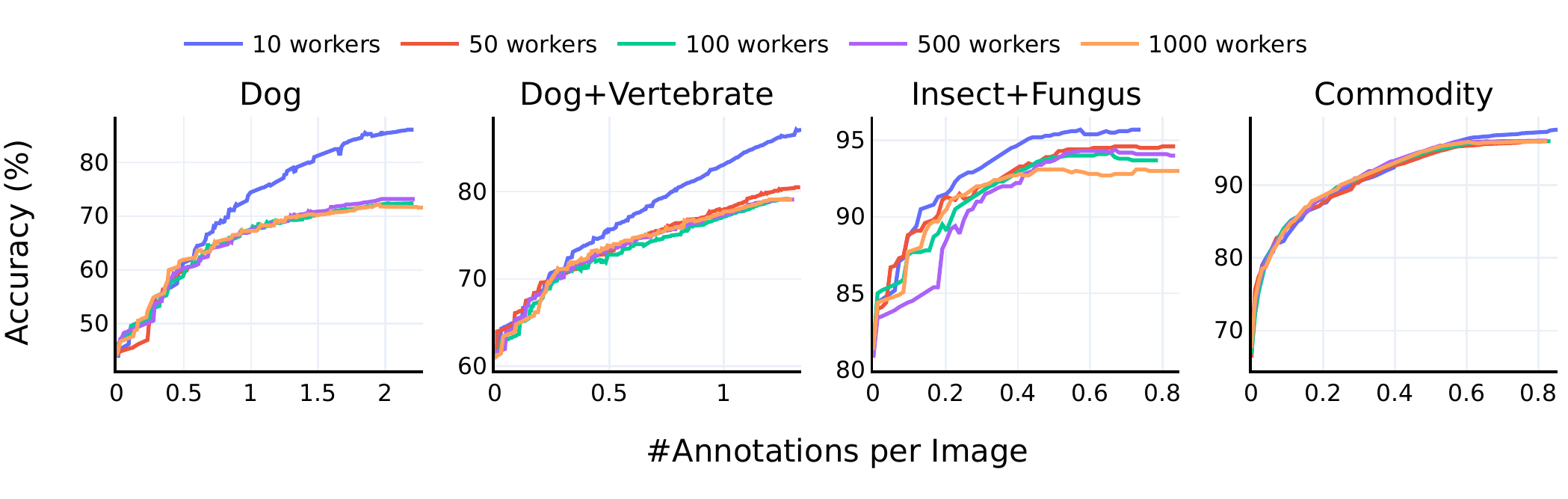}
  \caption{\textbf{Ablation of different numbers of workers.} Lower number of workers usually results in greater annotation quality but does not necessarily improve efficiency.}
  \label{sup:fig:num_workers}
\end{figure}

We explore how the number of workers affects annotation efficiency. In Fig.\ref{sup:fig:num_workers}, we find that having a fewer number of workers results in better label quality due to the better worker skills estimation, especially in the fine-grained dataset where the worker skills estimation matters more.

\subsection{Model Update Frequency}
\begin{figure}[h]
  \centering
  \includegraphics[width=0.8\textwidth]{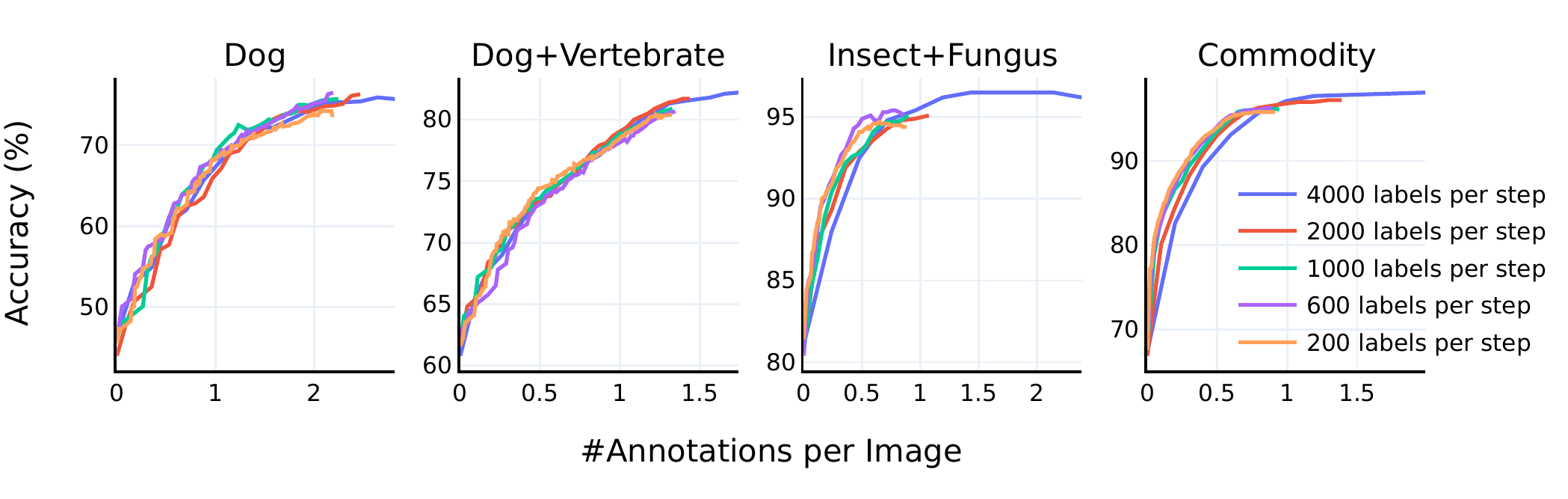}
  \caption{\textbf{Ablation of different numbers of labels per step.} Models that collect more labels per step tends to overhoot in number of annotations, while the final label quality remains similar.}
  \label{sup:fig:model_update_frequency}
\end{figure}

In reality, we need to consider the frequency of updating the model and collecting human labels, and the latency of model updates varies across applications.
In Fig.\ref{sup:fig:model_update_frequency}, we find that having lower update frequencies (higher number of annotations per update) tends to overshoot in the number of annotations, while the method is robust to low update frequency.

\subsection{Worker Prior Changes}
\label{sup:sec:prior_change}

\begin{figure}[h]
  \centering
  \includegraphics[width=0.8\textwidth]{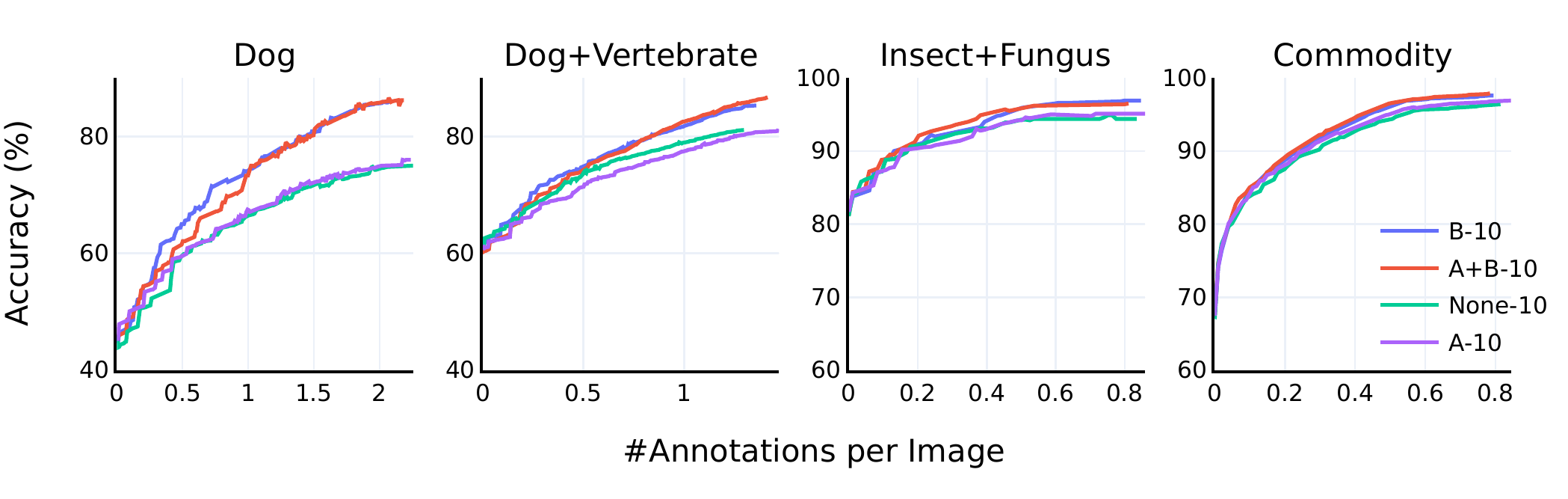}
  \caption{\textbf{Ablation of different priors.} For fine-grained dataset, considering worker identity in the prior improves a lot since the worker skills can vary a lot. However, having better prior does not necessarily lead to better performance.}
  \label{sup:fig:prior_change-10}
\end{figure}

There are multiple ways to design a better prior.
Here, we discuss two possible ways: \textbf{A)} Considering class identity and \textbf{B)} Considering worker identity.
To consider the class identity, the task designer needs to have a clear thought of which classes are harder.
To consider the worker identity, the task designer needs to ask several gold standard questions to each worker.
In Fig~\ref{sup:fig:prior_change-10}, we ablate the choice of having \textbf{None}, \textbf{A}, \textbf{B}, and \textbf{A+B} with prior strength 10.
For the fine-grained datasets, considering worker identity in the prior improves a lot since the worker skills can vary significantly.
For the coarse-grained datasets, the improvement is marginal.
For all other experiments in this paper, we adopt \textbf{None}.% and set $n_\beta=10$.

In reality, we need to consider the frequency of updating the model and collecting human labels, and the latency of model updates varies across applications.
In Fig.\ref{sup:fig:model_update_frequency}, we find that having lower update frequencies (higher number of annotations per update) tends to overshoot in the number of annotations, while the method is robust to low update frequency.

\subsection{Task Assignment}

\begin{figure}[h]
  \centering
  \includegraphics[width=0.8\textwidth]{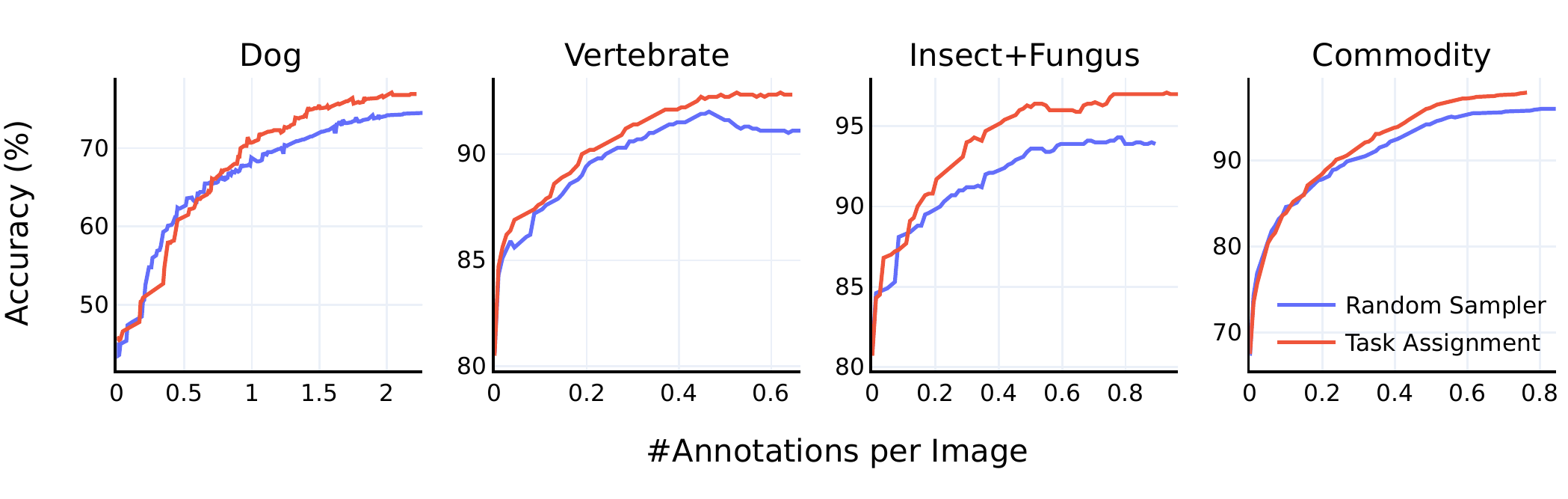}
  \caption{\textbf{Task Assignment.} The learnt skills consistently increase the annotation efficiency.}
  \label{sup:fig:task_assignment}
\end{figure}

To verify if the learnt worker skills help with task assignment, we propose a simple greedy algorithm. For an image sampled for annotation from the unfinished set $\mathcal{U}$, we use $\bar{y_i}$ to find the best possible workers from the workers pool using the currently estimated worker skill, with a cap on the maximum number of annotations $\alpha$ allowed per worker.

\begin{align} 
    j \leftarrow \arg\max_{j \in \hat{\mathcal{W}}} \mathbb{E}_{y \sim \bar{y_i}} \bar{w_j}[y, y], \hat{\mathcal{W}}=\{j | \mathopen|\mathcal{Z}_j\mathclose| \leq \alpha\}
\end{align}
The worker importance is measured by a weighted sum of the worker's per-class reliability and the number of annotations assigned to the worker. The weights are inversely proportional to the model's per-class accuracy. Ideally, number worker annotations would be highly correlated with worker importance. 
In Fig.~\ref{sup:fig:task_assignment}, we show the results on different splits .

\section{Transfer to Human Workers}
To validate if the proposed approach can gracefully transfer to human workers, we further collect 2519 annotations of 1657 images from \textit{Insesct+Fungus}. We perform data annotation from 11 workers on Toronto Annotation Suite. The average annotation accuracy is 0.908.
For the experiments, we include 13000 additional images as unlabeled data for semi-supervised learning.
Other details remain the same.

\section{Data in the Unfinished Set}

\begin{figure}[h]
    \centering
    \includegraphics[width=0.8\linewidth, height=5cm]{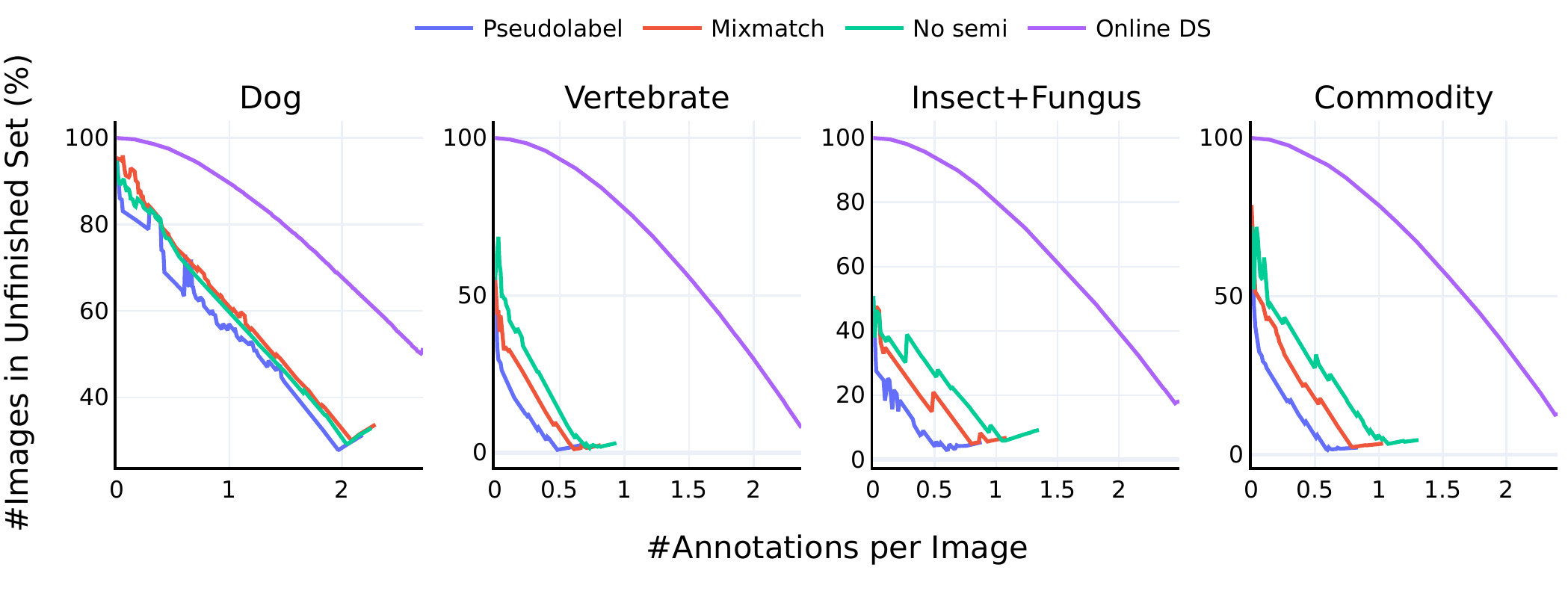}
    \caption{\textbf{Normalized number of images in the unfinished set.}}
    \label{sup:fig:semi_unfinished_set}
\end{figure}

For all the experiments, we suggest performing early stopping and leave the rest of the unfinished set $\{x_i | \mathcal{R}(\bar{y_i} > C)\}$ to a separate process, possibly expert annotators.
We show the normalized number of images in the unfinished set in Fig.~\ref{sup:fig:semi_unfinished_set}. 
For coarse-grained datasets, there are almost no images left in the unfinished set, while in fine-grained datasets, \eg \textit{Dog}, there can be as many as 31\% of images left in the unfinished set.

\subsection{Precision of the Finished Set}
\begin{figure}[h]
  \centering
  \includegraphics[width=0.8\linewidth, height=5cm]{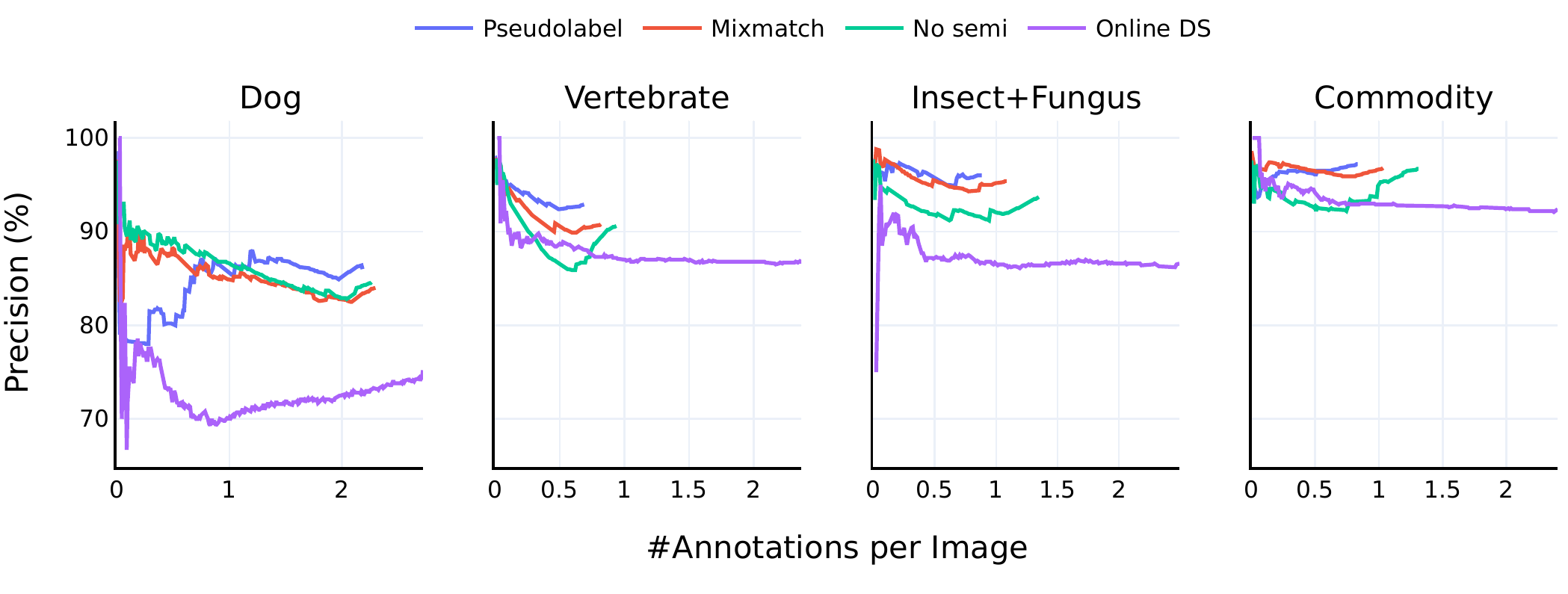}
  \caption{The precision of the finished set of images.}
  \label{sup:fig:precision}
\end{figure}

In the main paper, we show the overall label accuracy. Here, we are interested in the precision of the finished set. 
In Fig~\ref{sup:fig:precision}, we measure the precision of the finished set of different methods and our proposed framework.
In the \textit{Dog}, the overall accuracy is 75.9\%, while the precision of the finished set is 86.4\%. 
The finished set size is 15598, \ie we have 13368 correctly labeled images out of 22704 images in the finished set. 
%We also find that the top-5 label precision is usually 
%\AK{We find that the top-5 label precision can be very high...}

\section{Top 5 in ImageNet100}
\begin{figure}[h]
  \centering
  \begin{subfigure}[b]{0.4\textwidth}
    \centering
  \includegraphics[width=\textwidth, height=6cm]{assets/camera_ready/images/plot_imagenet100.pdf}
  \end{subfigure}
  \begin{subfigure}[b]{0.4\textwidth}
    \centering
  \includegraphics[width=\textwidth, height=6cm]{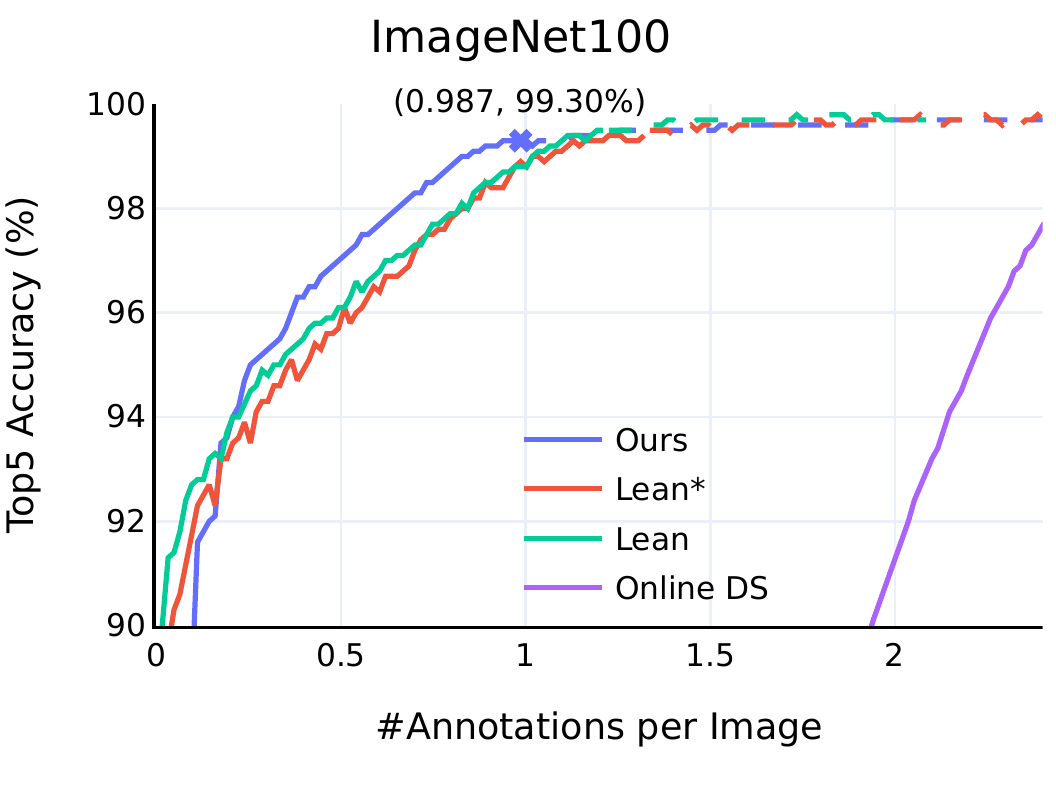}
  \end{subfigure}
  \caption{Top1 and top5 label accuracy on ImageNet100.}
  \label{sup:fig:imagenet100_all}
\end{figure}

The images in ImageNet can have multiple objects, making top1 accuracy unreliable.
In Fig.~\ref{sup:fig:imagenet100_all}, we show the comparison in terms of top5 accuracy.
Our proposed approach still performs slightly better than its counterparts. 
Note that the top5 accuracy is more error-tolerant, making the improvement gap smaller.

\section{Implementation Details}
For each experiment, the learning rate is $\lambda * \text{BatchSize}$, where $\lambda$ is the learning rate ratio.
Regarding the worker prior, we follow previous work~\cite{Branson_2017_CVPR} to use a tiered prior system.
Ignoring the worker identity, we use a homogeneous prior \ie we ignore both worker identity and class identity. 
If gold standard questions are used, one can use a more sophisticated prior.
We show the ablation of using different prior in Sec.~\ref{sup:sec:prior_change}.
Each worker reliability is represented as a confusion matrix over the label set.
We assume a worker annotation $z$ given $y=k$ is a Dirichlet distribution $\text{Dir}(n_\beta \alpha_k)$, where $n_\beta$ is the strength of the prior and $\alpha_k$ is estimated by pooling across all workers and classes.
We set $\alpha_k = 0.7$ and $n_\beta=10$ for all experiments. 

We perform hyperparameters search on learning rate ratio, weight decay ratio. For mixmatch, we perform an additional search over $\mu$, and $\gamma$.
In Tab.~\ref{sup:tab:search_range}, we show the search range of these hyperparameters.

\begin{table}
\begin{tabularx}{0.95\linewidth}{l|l}
  \centering
  \textbf{Parameter} & \textbf{Search Values} \\
  \hline \hline
  $\lambda$ &  $0.001, 0.0005, 0.0001, 0.00005$ \\ \hline
  weight decay & $0.001, 0.005, 0.0005, 0.0001$ \\ \hline
  $\mu$ & $3, 5, 10$ \\  \hline
  $\gamma$ & $50, 75, 100, 150$ \\  \hline  
\end{tabularx}
\caption{The search range of the hyperparameters.}
\label{sup:tab:search_range}
\end{table}

\section{Crowdsourcing on AMT}

\begin{figure}[h]
  \begin{adjustbox}{width=1.\linewidth}
  \begin{subfigure}[b]{0.48\textwidth}
      \centering
      \includegraphics[height=5.5cm]{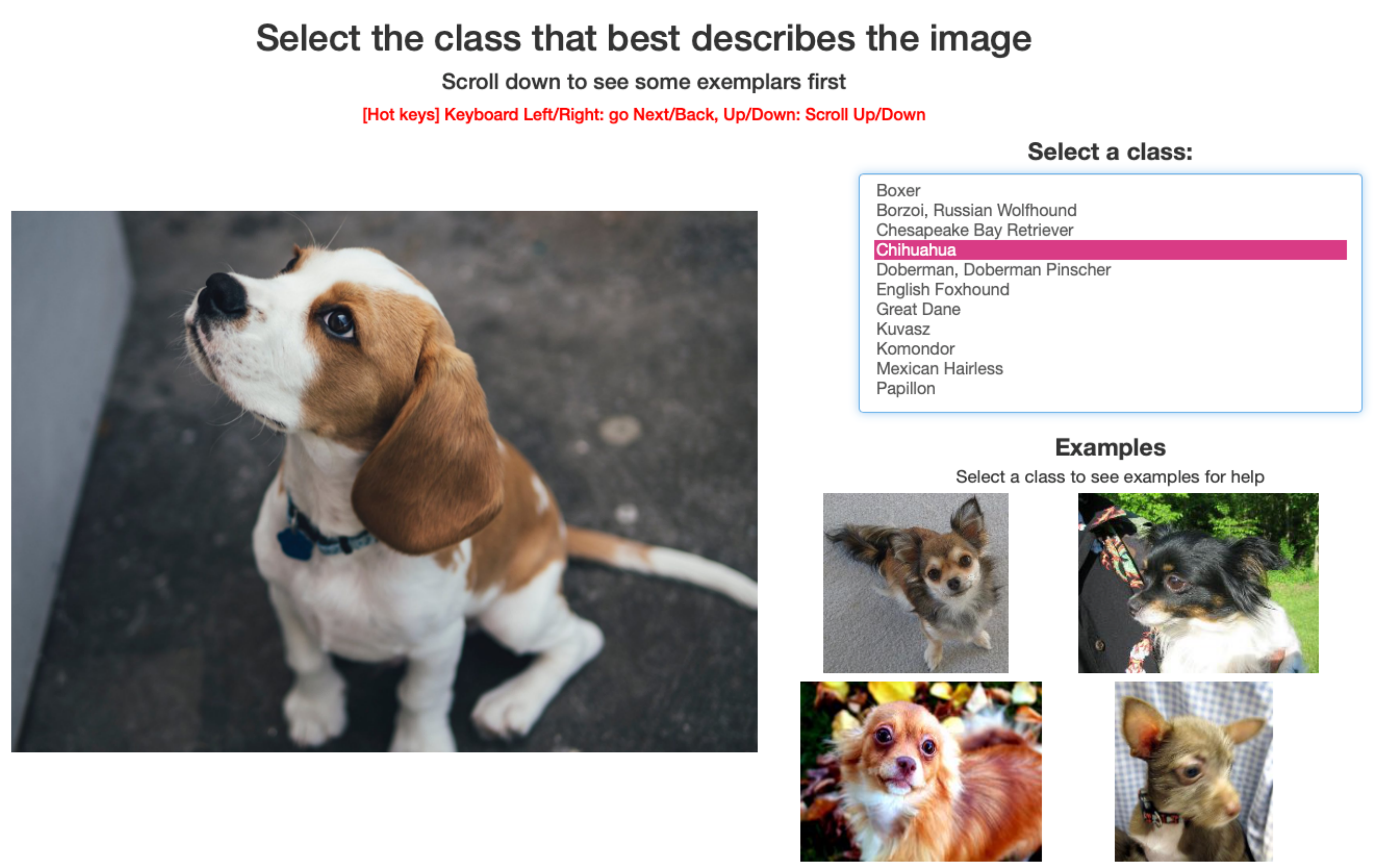}
      \caption{The user interface on AMT for crowdsourcing.}
      \label{sup:fig:user_interfaces}
  \end{subfigure}
  \hspace{3mm}
  \begin{subfigure}[b]{0.48\textwidth}
      \centering
      \includegraphics[height=5.5cm]{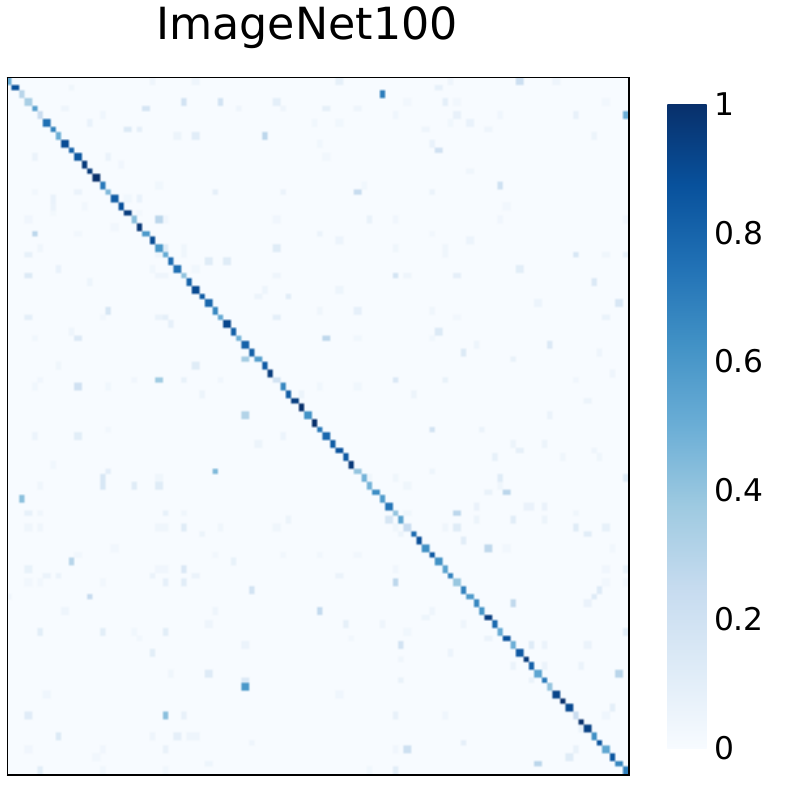}
      \caption{Average confusion matrix of all workers on the ImageNet-100 dataset}
      \label{sup:fig:global_worker_cm}
  \end{subfigure}
  \end{adjustbox}
\end{figure}

Prior work~\cite{hua2013collaborative, long2015multi} simulate workers as confusion matrices, and the class confusions are modeled with symmetric uniform noise, which can result in over-optimistic performance estimates. Human annotators usually exhibit \textit{asymmetric} and \textit{structured} confusion. We thus crowdsource the confusion matrices from human workers for the simulation. Fig.~\ref{sup:fig:user_interfaces}, shows our user interface on Amazon Mechanical Turk (AMT) for crowdsourcing.

\subsection{Workers Exhibit Structured Noise}
\begin{figure}[h]
    \centering
    \includegraphics[width=1\linewidth, height=4.3cm]{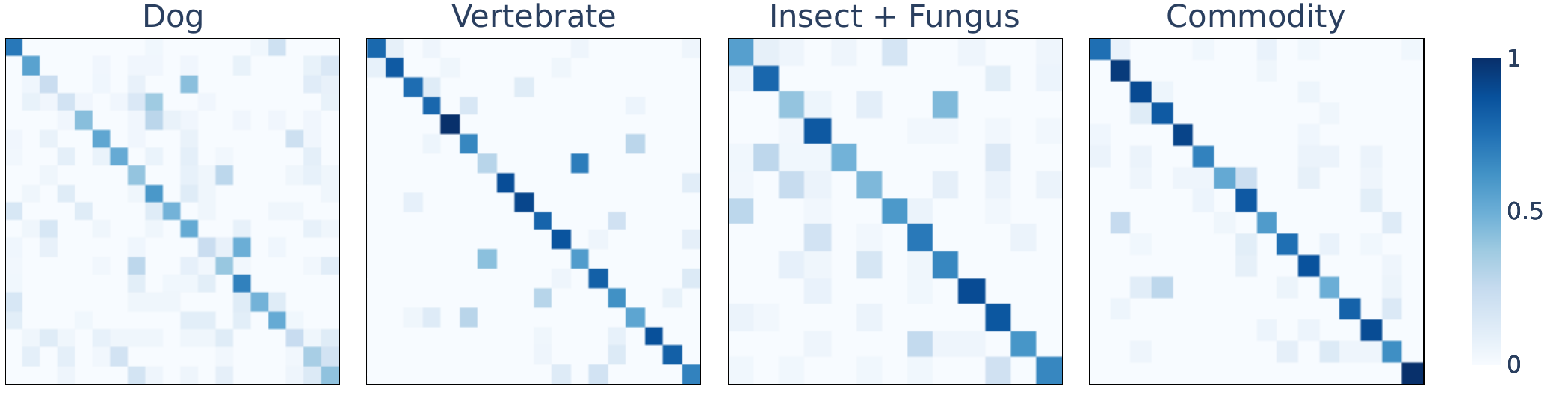}
    \caption{\textbf{The confusion matrix ImageNet100 splits.} We can see that the label confusion is neither unifom nor symmetric.}
    \label{sup:fig:global_worker_cm_split}
\end{figure}

We show the crowdsourced confusion matrices for different splits in Fig.~\ref{sup:fig:global_worker_cm_split}. For coarse-grained datasets, \eg \textit{Commodity}, there is low confusion, while in fine-grained datasets (Rest), the confusions are strong and correlated to class identities.
In Fig.~\ref{sup:fig:global_worker_cm}, we also show the average confusion matrix of all the workers on ImageNet100.

\section{Simulating Workers}
Here, we show the Python implementation to sample the simulated workers used in our experiments in Listing~\ref{sup:algo:worker_simulation}.

\begin{lstlisting}[language=Python, label=sup:algo:worker_simulation, caption=Worker Simulation Code]
import numpy as np

def sample_confusion_matrix(smooth_ratio: float, 
                            noise_level: float, 
                            target_classes: list, 
                            imagenet100: list, 
                            group: list, 
                            group_workers: dict):
      
  cm = []
  for k, v in group_workers.items():
      v = np.array(v)
      global_v = v.sum(0)
      idx = self.npr.choice(range(len(v)), 1) 
      cm.append(smooth_ratio * global_v + v[idx]) 
  cm = sum(cm)  # (100, 100) np.ndarray

  idx_to_keep = np.array([imagenet100.index(c.lower()) for c in target_classes])
  cm = cm[idx_to_keep, :][:, idx_to_keep]
  cm = cm / (cm.sum(1, keepdims=True) + 1e-8)

  def __which_group(i):
      for g_idx, g in enumerate(groups):
          if i in g:
              return g_idx


  # Add uniform noise in off-diagonal terms                                                    
  for i, c_i in enumerate(target_classes):
      c_i_group = __which_group(c_i)
      same_group_mask = np.zeros(cm.shape[0]).astype(np.bool)
      same_group_mask[i] = True
      for j, c_j in enumerate(target_classes):
          if i != j and c_i_group == __which_group(c_j):
              same_group_mask[j] = True
          
      if same_group_mask.sum() > 0:
          density_to_spread = cm[i, same_group_mask].sum()
          cm[i, same_group_mask] = cm[i, same_group_mask] * (1 - noise_level)
          cm[i, ~same_group_mask] += density_to_spread * (noise_level) \
                                      / max(sum(~same_group_mask), 1e-8)

  return cm                                                       

\end{lstlisting}

\section{Semi-Supervised Learning: MixMatch}

\textbf{MixMatch} constructs virtual training examples by mixing the labeled, and unlabeled data using a modified version of MixUp \cite{zhang2017mixup}. We modify MixMatch for online annotation, where the input to the model being learnt is the feature vector $\phi(x)$.
The labeled set is defined by the data points with at least one worker annotation $\{x_i | \mathopen|\mathcal{W}_i\mathclose| > 0\}$, 
and the unlabeled set is defined by the data points whose largest probability is larger than a predefined threshold and is not in the labeled set, $\{ x_i | p(\bar{y_i} | \mathcal{Z}_i) > 1-\tau \& \mathopen|\mathcal{W}_i\mathclose| = 0\}$.
We use the same threshold as the one used in pseudo labels.
The mixmatch loss consists of cross entropy of the labeled set and the l2 minimization of the mixed set.
The mixed set is constructed by sampling $(x_1, p_1)$ from the labeled set and $(x_2, p_2)$ from the unlabeled set and interpolate both input and output.

\begin{align} 
\begin{split}
  \lambda & \sim \text{Beta}(\alpha, \alpha) \\
  \lambda' &= \max(\lambda, 1- \lambda) \\
  x' &= \lambda' x_1 + (1 - \lambda') x_2 \\
  p' &= \lambda' p_1 + (1 - \lambda') p_2 \\
  \mathcal{S}_{\text{mixed}} & \leftarrow (x', p') \\ 
  L &= \mathbb{E}_{(x, y) \sim \{x_i, \bar{y_i} | \mathopen|\mathcal{W}_i\mathclose| > 0\}} H( \bar{y_i} , p(y | \phi(x_i), \theta)) \\
  &+  \mu \mathbb{E}_{(x_i, p_i) \sim \mathcal{S}_{\text{mixed}}} \left\lVert p_i - p(y | \phi(x_i), \theta) \right\rVert^2_2
\end{split}
\end{align}
 
\noindent where $\mu$ is the hyperparameters. 
When the labeled set is small, we usually sample $\gamma$ times more data from the unlabeled set.
In our experiments, $\mu$ and $\gamma$ are set by performing hyperparameters search mentioned in the main paper.

\section{Unexplored Questions}
We discuss shortcomings and additional directions for progress in this section.

\textbf{Using multiple ML models as workers:}
Different self-supervised pretext tasks can provide orthogonal benefits for downstream tasks ~\cite{sun2019unsupervised}. Downstream labeling tasks could go beyond semantic classes, such as annotating the viewing angle of a car in an image as a classification problem. One could imagine a scenario where classifiers trained on multiple self-supervised features are treated as machine ``workers", whose skills for the task at hand are simultaneously estimated, similar to those for human workers.

\textbf{Annotating at a small scale / beyond ImageNet:}
We do not discuss annotating small-scale data and restrict ourselves to ImageNet in this work. When the target dataset is small, we expect to be able to finetune self-supervised features on it to initialize the feature extractor $\phi$.

\textbf{Leveraging label hierarchies:}
~\cite{van2018lean} propose a method to utilize label hierarchies to efficiently factorize large confusion matrices used to represent worker skills. We expect to see additional benefits from incorporating their skill estimation method into our algorithm.

\textbf{Simulating Image Difficulty:}
Our proposed simulation does not account for image-level annotation difficulty, and simulated labels are obtained using a realistic confusion matrix applied to the ground truth label. Improving our simulation to consider this is something we would like to explore in future work.

\textbf{Going beyond classification:}
The proposed method can be used to go beyond classification by changing the likelihood modeling for human annotations. In this vein, ~\cite{Branson_2017_CVPR} show results on keypoint and bounding box annotation. Incorporating learning into the loop requires specific attention to detail per task, and we leave this to future work.

%\AK{Complete the discussion of limitations with bullets like we have in conc.tex}

\section{Class Names in Each Subtask}
Tab.~\ref{sup:tab:class_names}, shows the classes that comprise each subset in our ImageNet-sandbox dataset.

\begin{table*}[t!]
\begin{tabularx}{\textwidth}{l|X}
\textbf{Dataset} & \textbf{Class Names} \\
\hline \hline
Dog & Komondor, Mexican Hairless, Vizsla, Hungarian Pointer, Toy Terrier, Papillon, Boxer, Rottweiler, English Foxhound, Chihuahua, Shih-Tzu, Chesapeake Bay Retriever, Saluki, Gazelle Hound, Walker Hound, Walker Foxhound, Borzoi, Russian Wolfhound, Standard Poodle, Kuvasz, American Staffordshire Terrier, Staffordshire Terrier, American Pit Bull Terrier, Pit Bull Terrier, Doberman, Doberman Pinscher, Great Dane \\ \hline
Vertebrate & Meerkat, Mierkat, Hare, Robin, American Robin, Turdus Migratorius, Little Blue Heron, Egretta Caerulea, Tabby, Tabby Cat, Goose, Langur, Wild Boar, Boar, Sus Scrofa, Lorikeet, Garter Snake, Grass Snake, African Hunting Dog, Hyena Dog, Cape Hunting Dog, Lycaon Pictus, Gibbon, Hylobates Lar, Coyote, Prairie Wolf, Brush Wolf, Canis Latrans, Hognose Snake, Puff Adder, Sand Viper, American Coot, Marsh Hen, Mud Hen, Water Hen, Fulica Americana, Green Mamba, Gila Monster, Heloderma Suspectum, Red Fox, Vulpes Vulpes \\ \hline
Insect + Fungus & Gyromitra, Cauliflower, Fiddler Crab, Dung Beetle, Head Cabbage, American Lobster, Northern Lobster, Maine Lobster, Homarus Americanus, Stinkhorn, Carrion Fungus, Leafhopper, Rock Crab, Cancer Irroratus, Garden Spider, Aranea Diademata, Carbonara, Walking Stick, Walkingstick, Stick Insect, Chocolate Sauce, Chocolate Syrup \\ \hline
Commodity & Vacuum, Vacuum Cleaner, Computer Keyboard, Keypad, Bottlecap, Milk Can, Iron, Smoothing Iron, Mortarboard, Bonnet, Poke Bonnet, Sarong, Modem, Tub, Vat, Purse, Cocktail Shaker, Rotisserie, Jean, Blue Jean, Denim, Dutch Oven, Football Helmet\\ \hline
ImageNet20 & Robin, American Robin, Turdus Migratorius, Gila Monster, Heloderma Suspectum, Hognose Snake, Puff Adder, Sand Viper, Garter Snake, Grass Snake, Green Mamba, Garden Spider, Aranea Diademata, Lorikeet, Goose, Rock Crab, Cancer Irroratus, Fiddler Crab, American Lobster, Northern Lobster, Maine Lobster, Homarus Americanus, Little Blue Heron, Egretta Caerulea, American Coot, Marsh Hen, Mud Hen, Water Hen, Fulica Americana, Chihuahua, Shih-Tzu, Papillon, Toy Terrier, Walker Hound, Walker Foxhound, English Foxhound, Borzoi, Russian Wolfhound 
\end{tabularx}
\caption{Class names used in each split.}
\label{sup:tab:class_names}
\end{table*}

\subsection{More Qualitative Results on ImageNet100}
\label{sup:sec:more_qualitative_results}

\begin{figure*}[h]
  \centering
  \parbox{0.15\textwidth}{\centering Fiddler Crab}
  \parbox{0.15\textwidth}{\centering Walker Hound}
  \parbox{0.15\textwidth}{\centering Reel}
  \parbox{0.15\textwidth}{\centering Park Bench}
  \parbox{0.15\textwidth}{\centering Rocking Chair}
  \parbox{0.15\textwidth}{\centering Great Dane}
  \includegraphics[width=0.15\textwidth,height=1.6cm]{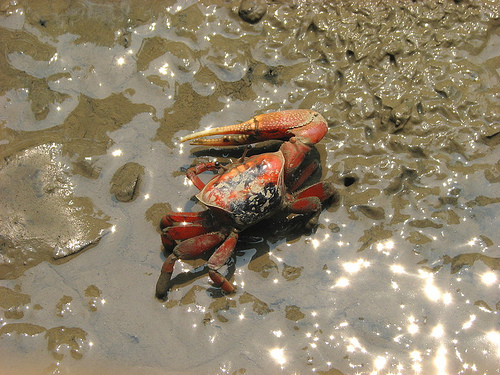}
  \includegraphics[width=0.15\textwidth,height=1.6cm]{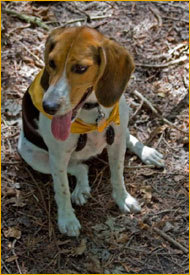}
  \includegraphics[width=0.15\textwidth,height=1.6cm]{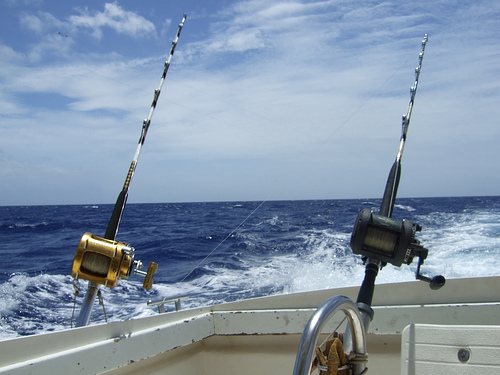}
  \includegraphics[width=0.15\textwidth,height=1.6cm]{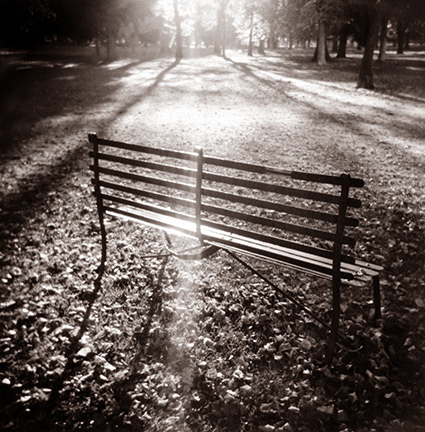}
  \includegraphics[width=0.15\textwidth,height=1.6cm]{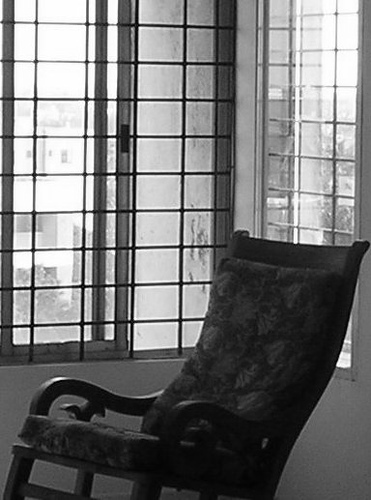}
  \includegraphics[width=0.15\textwidth,height=1.6cm]{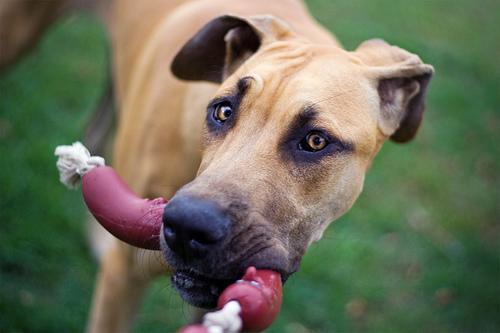}
  \parbox{0.15\textwidth}{\centering American Lobster}
  \parbox{0.15\textwidth}{\centering Cauliflower}
  \parbox{0.15\textwidth}{\centering American Staffordshire Terrier}
  \parbox{0.15\textwidth}{\centering Borzoi}
  \parbox{0.15\textwidth}{\centering Laptop}
  \parbox{0.15\textwidth}{\centering Bassinet}
  \includegraphics[width=0.15\textwidth,height=1.6cm]{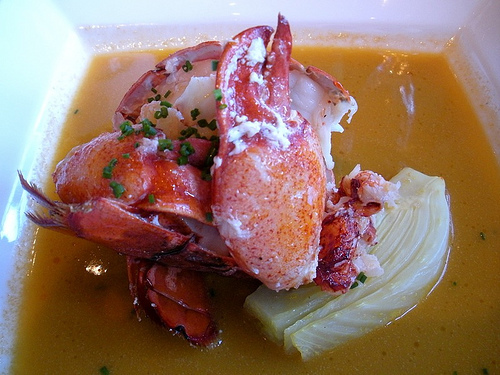}
  \includegraphics[width=0.15\textwidth,height=1.6cm]{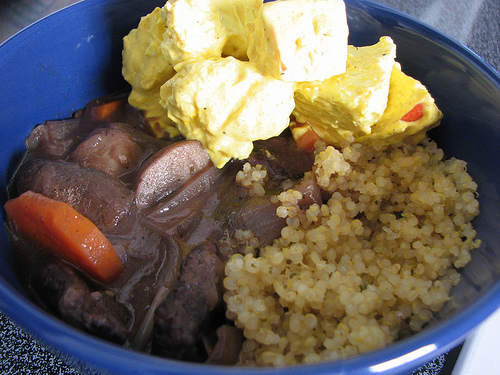}
  \includegraphics[width=0.15\textwidth,height=1.6cm]{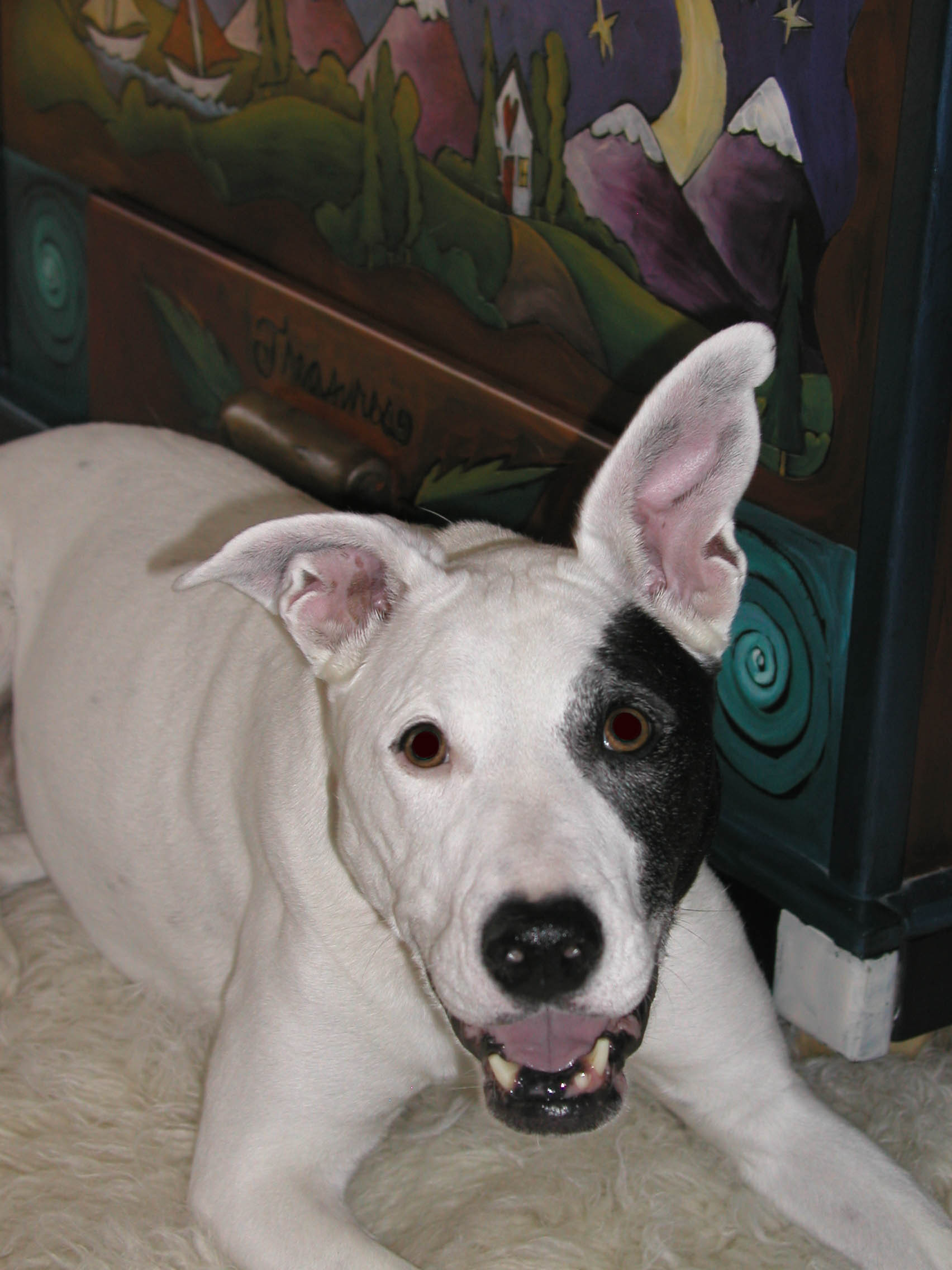}
  \includegraphics[width=0.15\textwidth,height=1.6cm]{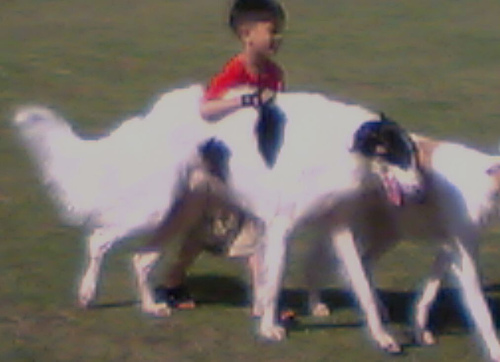}
  \includegraphics[width=0.15\textwidth,height=1.6cm]{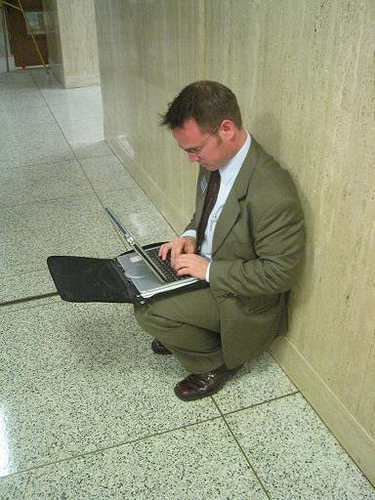}
  \includegraphics[width=0.15\textwidth,height=1.6cm]{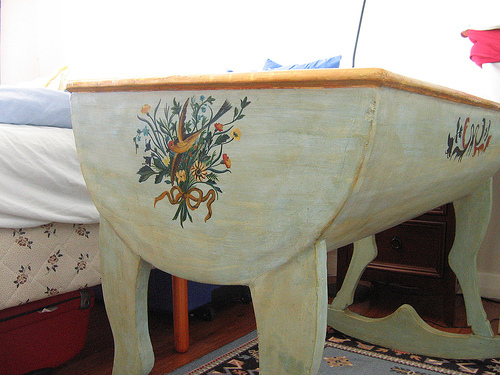}
  \parbox{0.15\textwidth}{\centering Chime}
  \parbox{0.15\textwidth}{\centering American Lobster}
  \parbox{0.15\textwidth}{\centering Reel}
  \parbox{0.15\textwidth}{\centering Chihuahua}
  \parbox{0.15\textwidth}{\centering American Staffordshire Terrier}
  \parbox{0.15\textwidth}{\centering Milk Can}
  \includegraphics[width=0.15\textwidth,height=1.6cm]{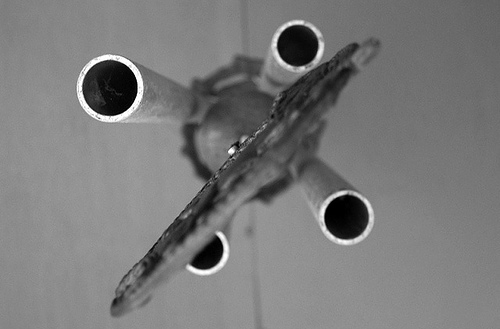}
  \includegraphics[width=0.15\textwidth,height=1.6cm]{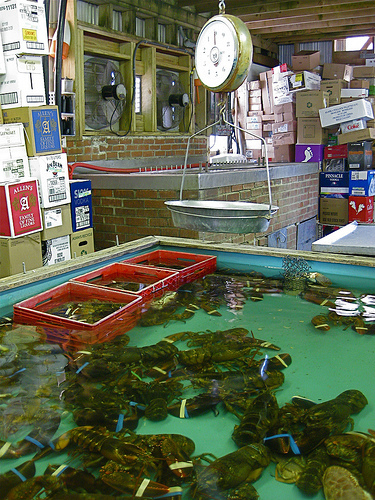}
  \includegraphics[width=0.15\textwidth,height=1.6cm]{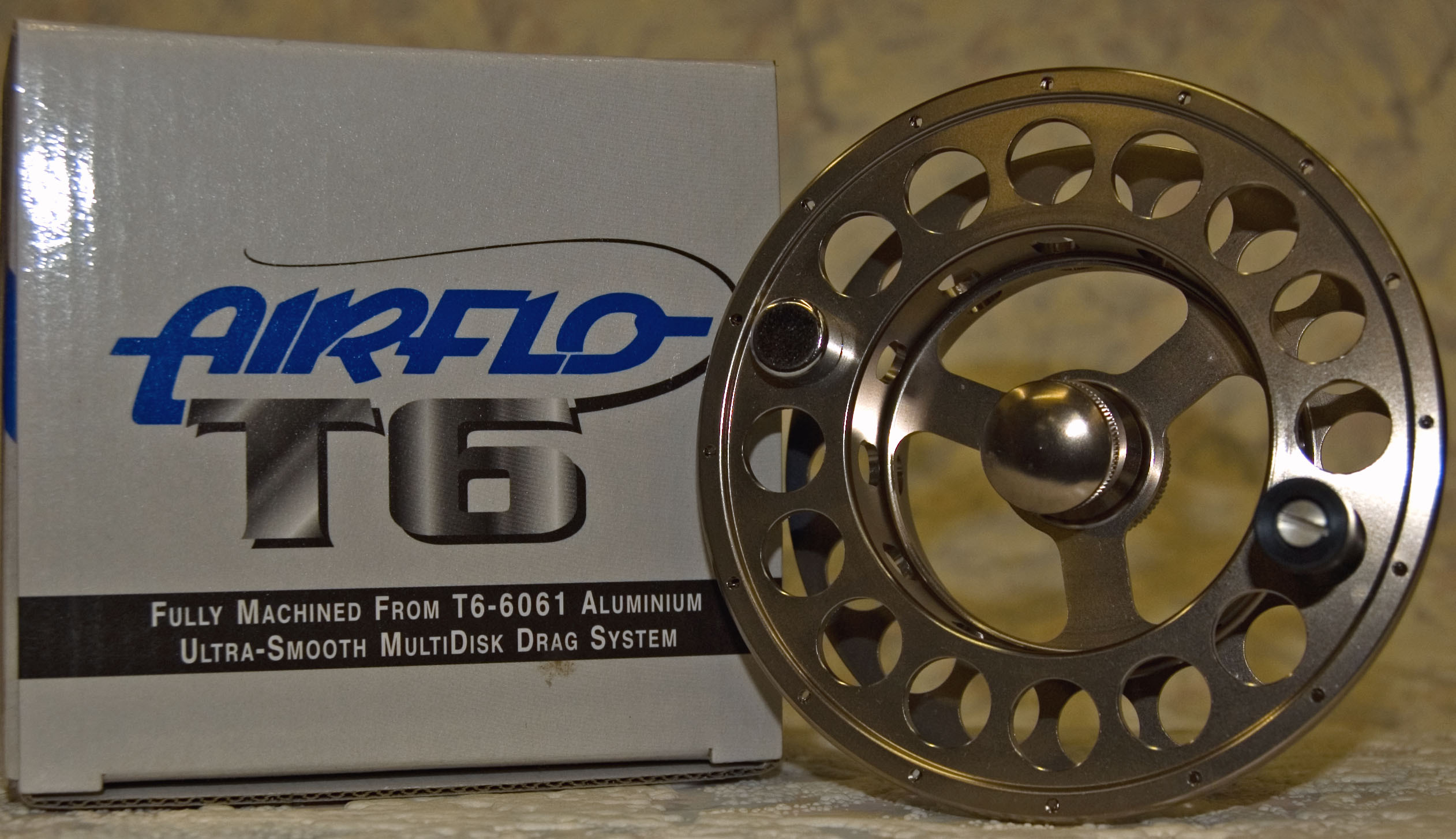}
  \includegraphics[width=0.15\textwidth,height=1.6cm]{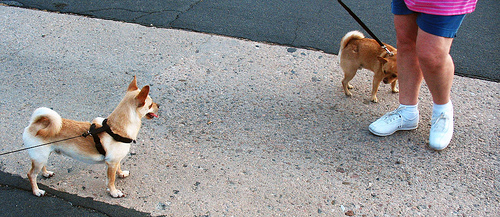}
  \includegraphics[width=0.15\textwidth,height=1.6cm]{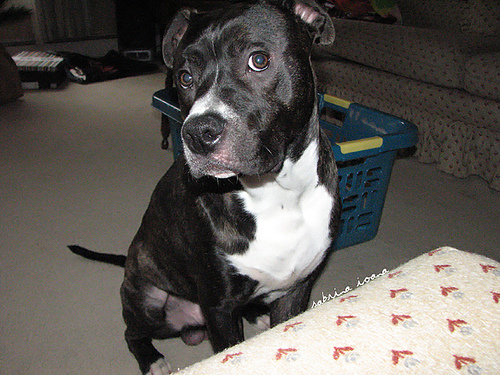}
  \includegraphics[width=0.15\textwidth,height=1.6cm]{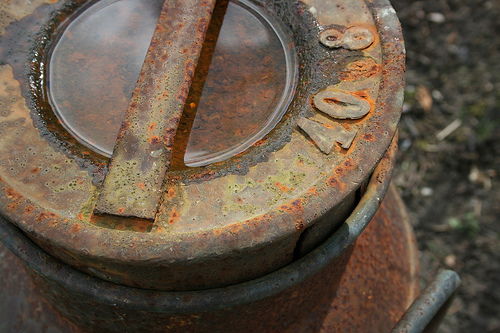}
\caption{Images in the unfinished set on ImageNet100}
\label{sup:fig:qualitative_unfinished}
\end{figure*}

\begin{figure*}[h]
\centering
\parbox{0.15\textwidth}{\centering Kuvasz}
\parbox{0.15\textwidth}{\centering Chocolate Sauce}
\parbox{0.15\textwidth}{\centering Theater Curtain}
\parbox{0.15\textwidth}{\centering Lampshade}
\parbox{0.15\textwidth}{\centering Mortarboard}
\parbox{0.15\textwidth}{\centering Tabby}
\includegraphics[width=0.15\textwidth,height=1.6cm]{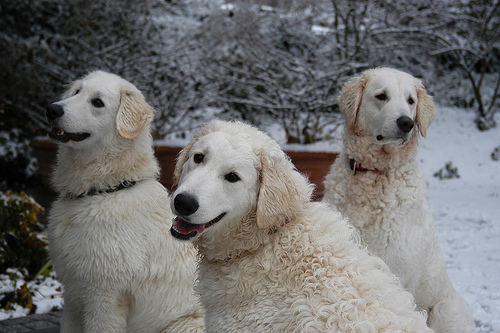}
\includegraphics[width=0.15\textwidth,height=1.6cm]{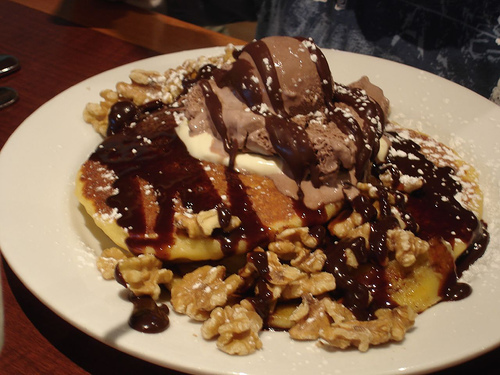}
\includegraphics[width=0.15\textwidth,height=1.6cm]{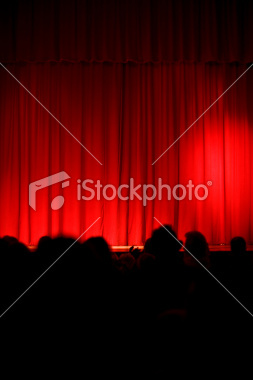}
\includegraphics[width=0.15\textwidth,height=1.6cm]{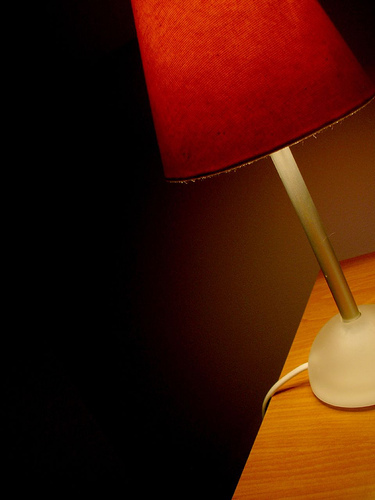}
\includegraphics[width=0.15\textwidth,height=1.6cm]{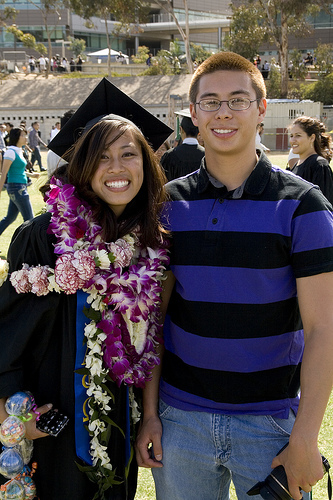}
\includegraphics[width=0.15\textwidth,height=1.6cm]{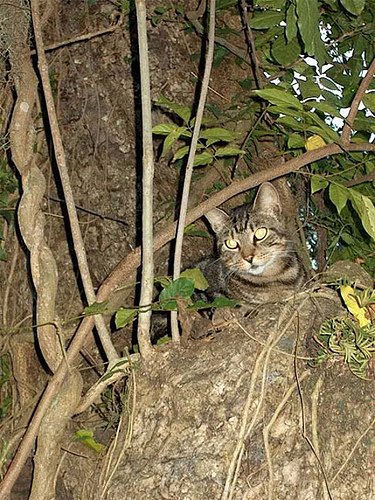}
\parbox{0.15\textwidth}{\centering Mortarboard}
\parbox{0.15\textwidth}{\centering Rock Crab}
\parbox{0.15\textwidth}{\centering English Foxhound}
\parbox{0.15\textwidth}{\centering Garter Snake}
\parbox{0.15\textwidth}{\centering Gasmask}
\parbox{0.15\textwidth}{\centering Kuvasz}
\includegraphics[width=0.15\textwidth,height=1.6cm]{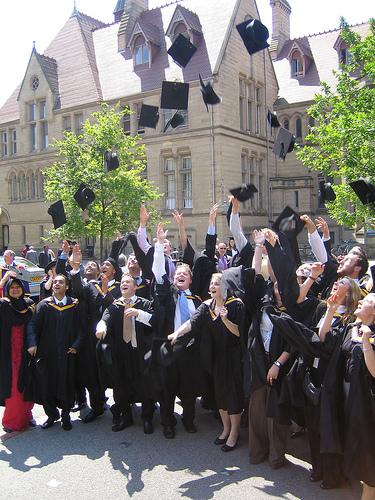}
\includegraphics[width=0.15\textwidth,height=1.6cm]{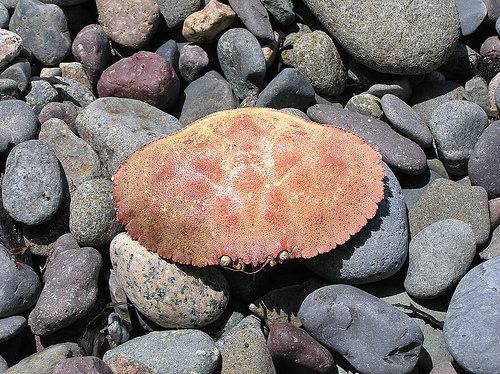}
\includegraphics[width=0.15\textwidth,height=1.6cm]{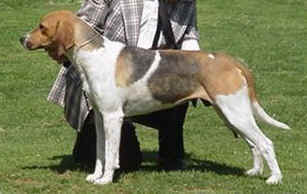}
\includegraphics[width=0.15\textwidth,height=1.6cm]{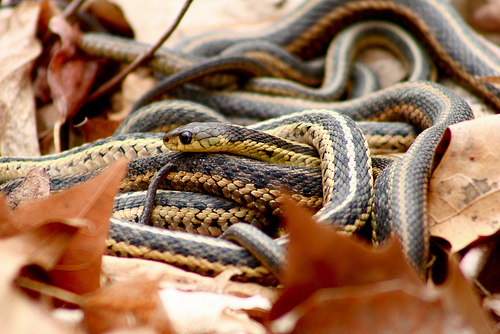}
\includegraphics[width=0.15\textwidth,height=1.6cm]{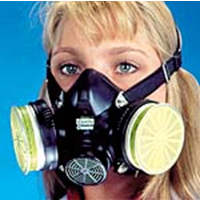}
\includegraphics[width=0.15\textwidth,height=1.6cm]{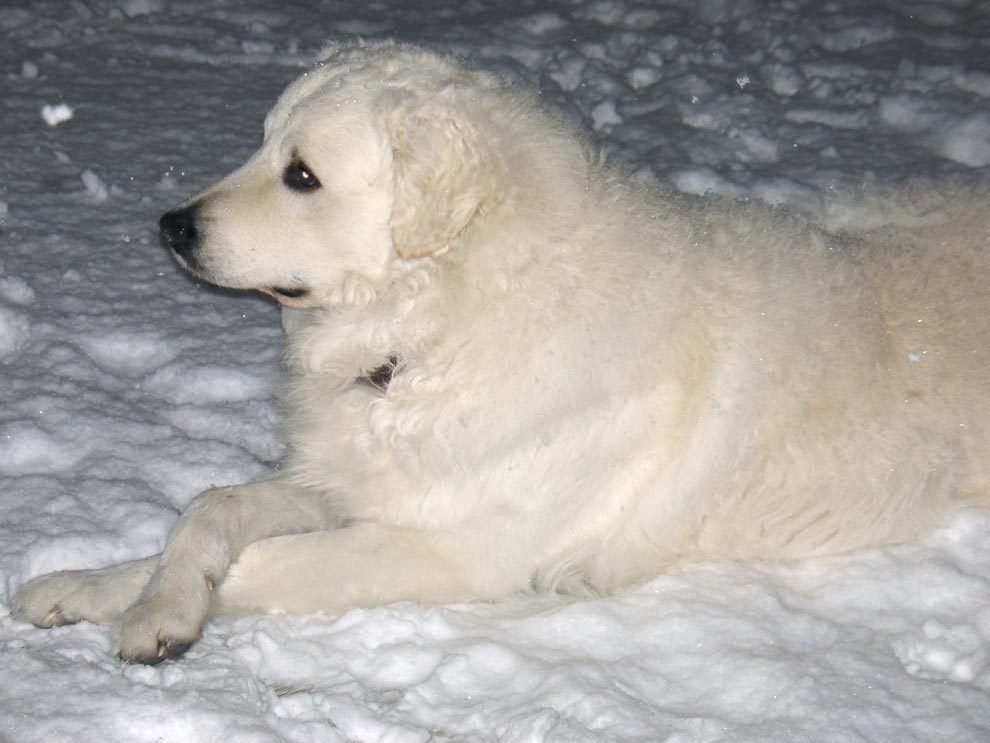}
\parbox{0.15\textwidth}{\centering Head Cabbage}
\parbox{0.15\textwidth}{\centering Computer Keyboard}
\parbox{0.15\textwidth}{\centering Theater Curtain}
\parbox{0.15\textwidth}{\centering Honeycomb}
\parbox{0.15\textwidth}{\centering Carbonara}
\parbox{0.15\textwidth}{\centering Wing}
\includegraphics[width=0.15\textwidth,height=1.6cm]{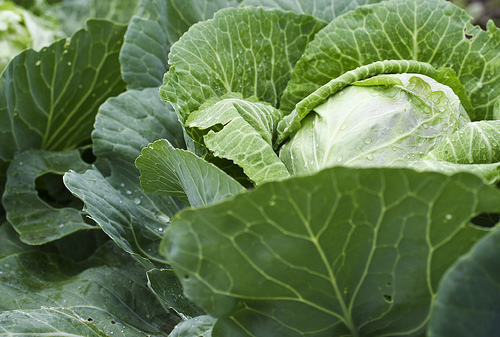}
\includegraphics[width=0.15\textwidth,height=1.6cm]{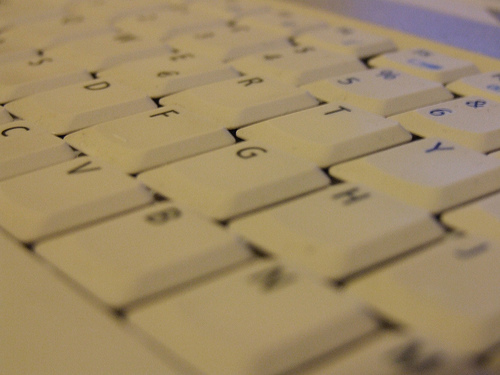}
\includegraphics[width=0.15\textwidth,height=1.6cm]{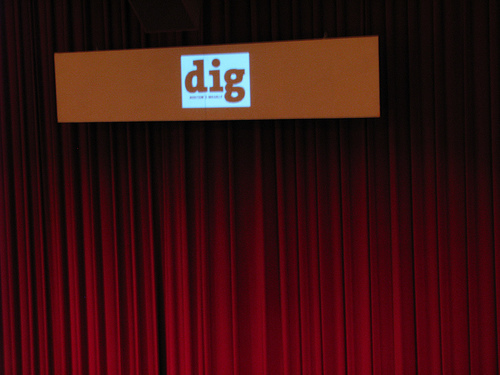}
\includegraphics[width=0.15\textwidth,height=1.6cm]{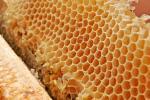}
\includegraphics[width=0.15\textwidth,height=1.6cm]{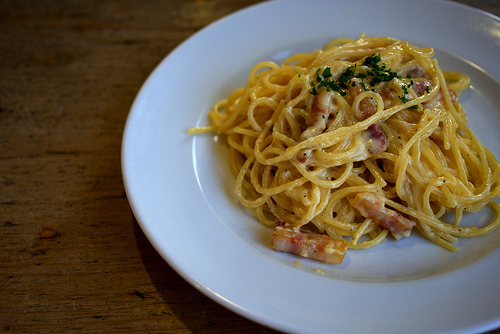}
\includegraphics[width=0.15\textwidth,height=1.6cm]{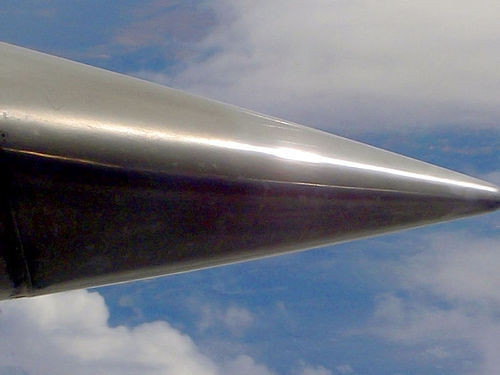}
\caption{Images with no annotations on ImageNet100}
\label{sup:fig:0_annotation}
\end{figure*}

\begin{figure*}[h]
\centering
\parbox{0.15\textwidth}{\centering Boathouse}
\parbox{0.15\textwidth}{\centering Head Cabbage}
\parbox{0.15\textwidth}{\centering Safety Pin}
\parbox{0.15\textwidth}{\centering Iron}
\parbox{0.15\textwidth}{\centering Bannister}
\parbox{0.15\textwidth}{\centering Tile Roof}
\includegraphics[width=0.15\textwidth,height=1.6cm]{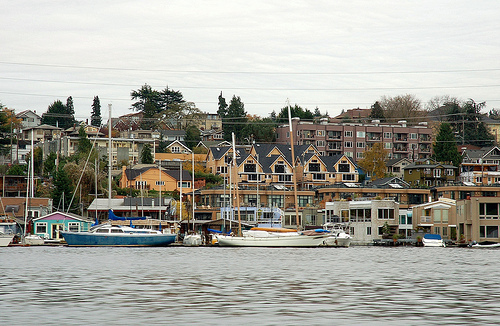}
\includegraphics[width=0.15\textwidth,height=1.6cm]{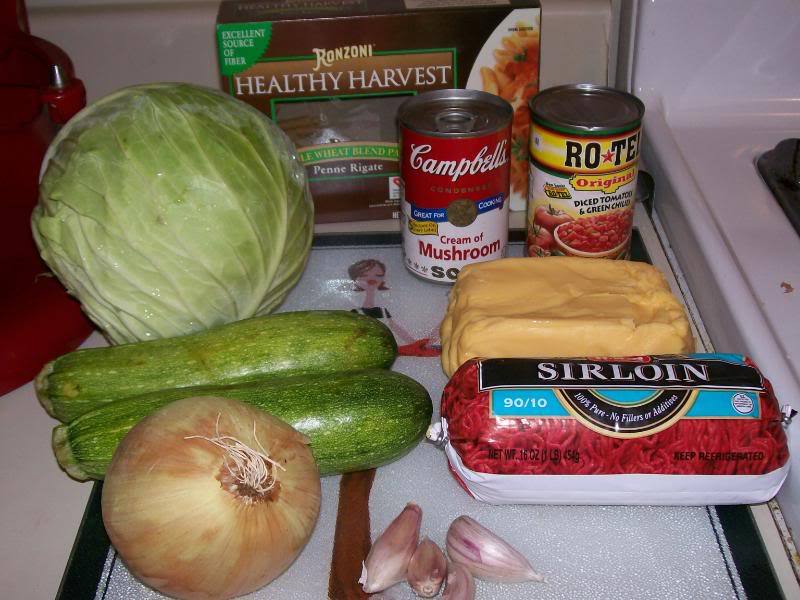}
\includegraphics[width=0.15\textwidth,height=1.6cm]{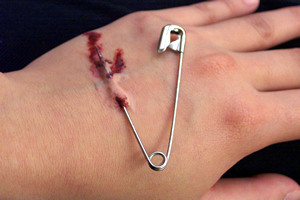}
\includegraphics[width=0.15\textwidth,height=1.6cm]{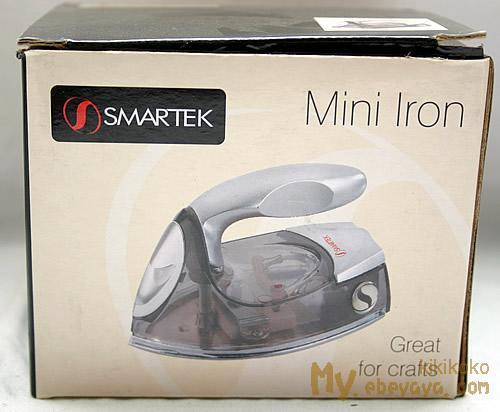}
\includegraphics[width=0.15\textwidth,height=1.6cm]{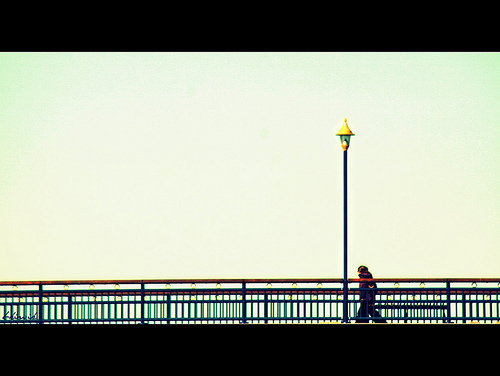}
\includegraphics[width=0.15\textwidth,height=1.6cm]{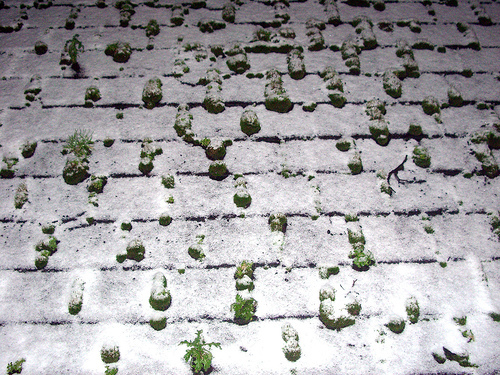}
\parbox{0.15\textwidth}{\centering Cinema}
\parbox{0.15\textwidth}{\centering Modem}
\parbox{0.15\textwidth}{\centering Football Helmet}
\parbox{0.15\textwidth}{\centering American Staffordshire Terrier}
\parbox{0.15\textwidth}{\centering Papillon}
\parbox{0.15\textwidth}{\centering Cinema}
\includegraphics[width=0.15\textwidth,height=1.6cm]{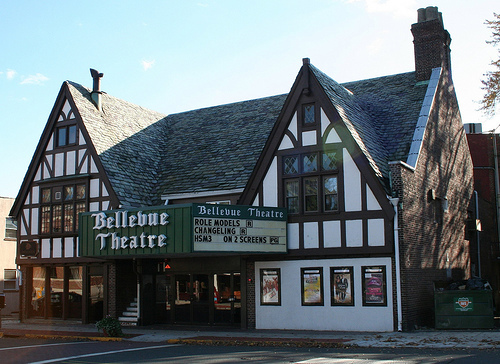}
\includegraphics[width=0.15\textwidth,height=1.6cm]{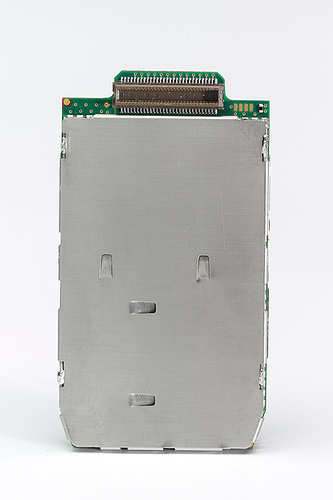}
\includegraphics[width=0.15\textwidth,height=1.6cm]{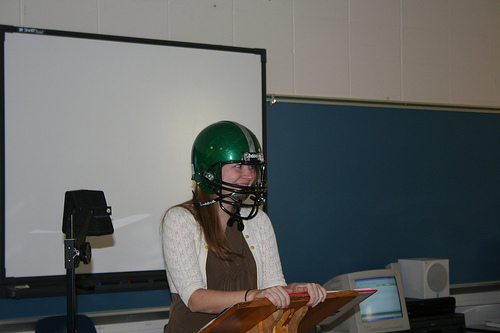}
\includegraphics[width=0.15\textwidth,height=1.6cm]{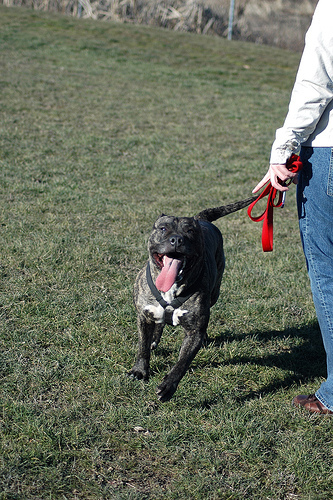}
\includegraphics[width=0.15\textwidth,height=1.6cm]{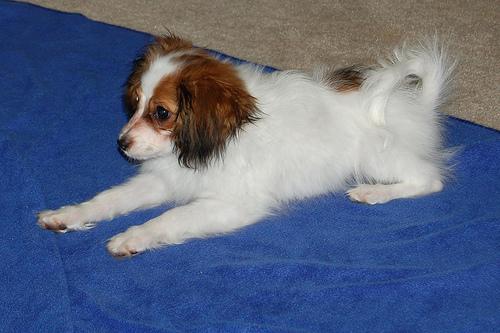}
\includegraphics[width=0.15\textwidth,height=1.6cm]{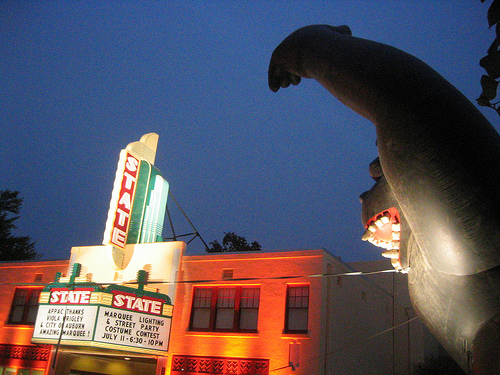}
\parbox{0.15\textwidth}{\centering Gibbon}
\parbox{0.15\textwidth}{\centering Sarong}
\parbox{0.15\textwidth}{\centering Doberman}
\parbox{0.15\textwidth}{\centering Pickup}
\parbox{0.15\textwidth}{\centering Bonnet}
\parbox{0.15\textwidth}{\centering Gibbon}
\includegraphics[width=0.15\textwidth,height=1.6cm]{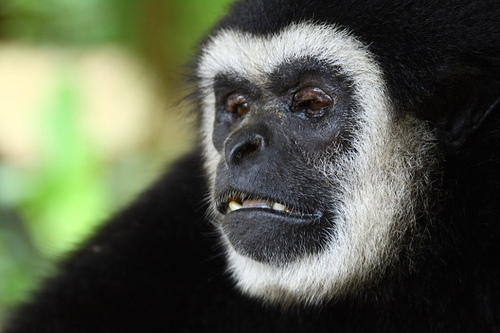}
\includegraphics[width=0.15\textwidth,height=1.6cm]{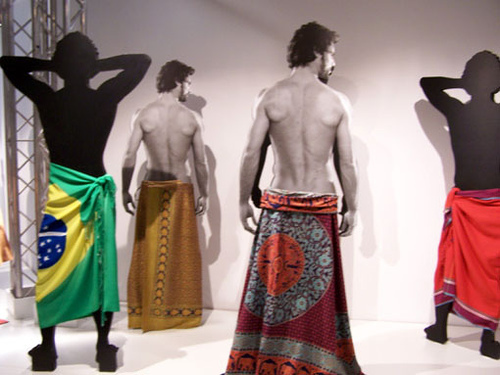}
\includegraphics[width=0.15\textwidth,height=1.6cm]{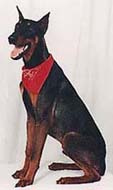}
\includegraphics[width=0.15\textwidth,height=1.6cm]{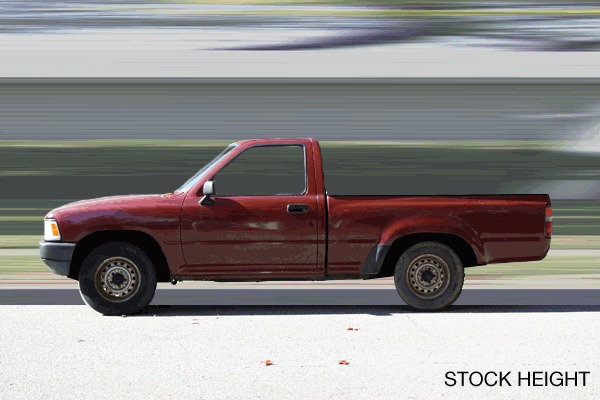}
\includegraphics[width=0.15\textwidth,height=1.6cm]{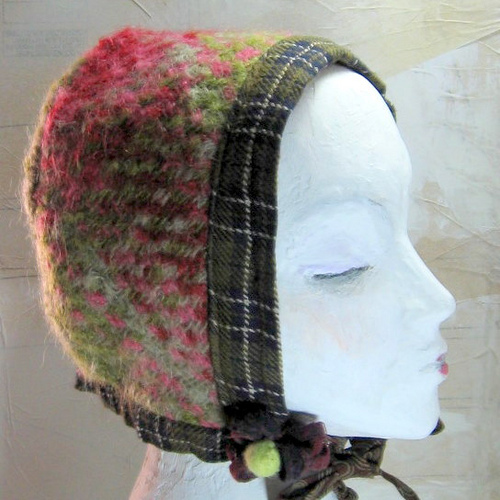}
\includegraphics[width=0.15\textwidth,height=1.6cm]{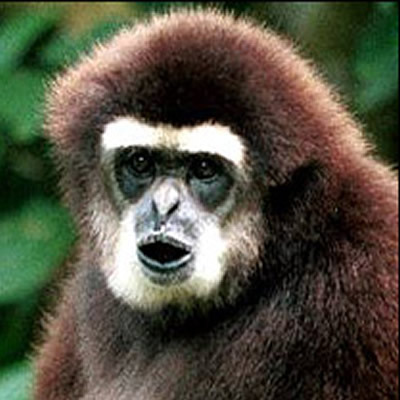}
\caption{Images with 1 annotation on ImageNet100}
\label{sup:fig:1_annotation}
\end{figure*}

\begin{figure*}[h]
\centering
\parbox{0.15\textwidth}{\centering Cinema}
\parbox{0.15\textwidth}{\centering Chesapeake Bay Retriever}
\parbox{0.15\textwidth}{\centering American Staffordshire Terrier}
\parbox{0.15\textwidth}{\centering Ambulance}
\parbox{0.15\textwidth}{\centering Great Dane}
\parbox{0.15\textwidth}{\centering Pirate}
\includegraphics[width=0.15\textwidth,height=1.6cm]{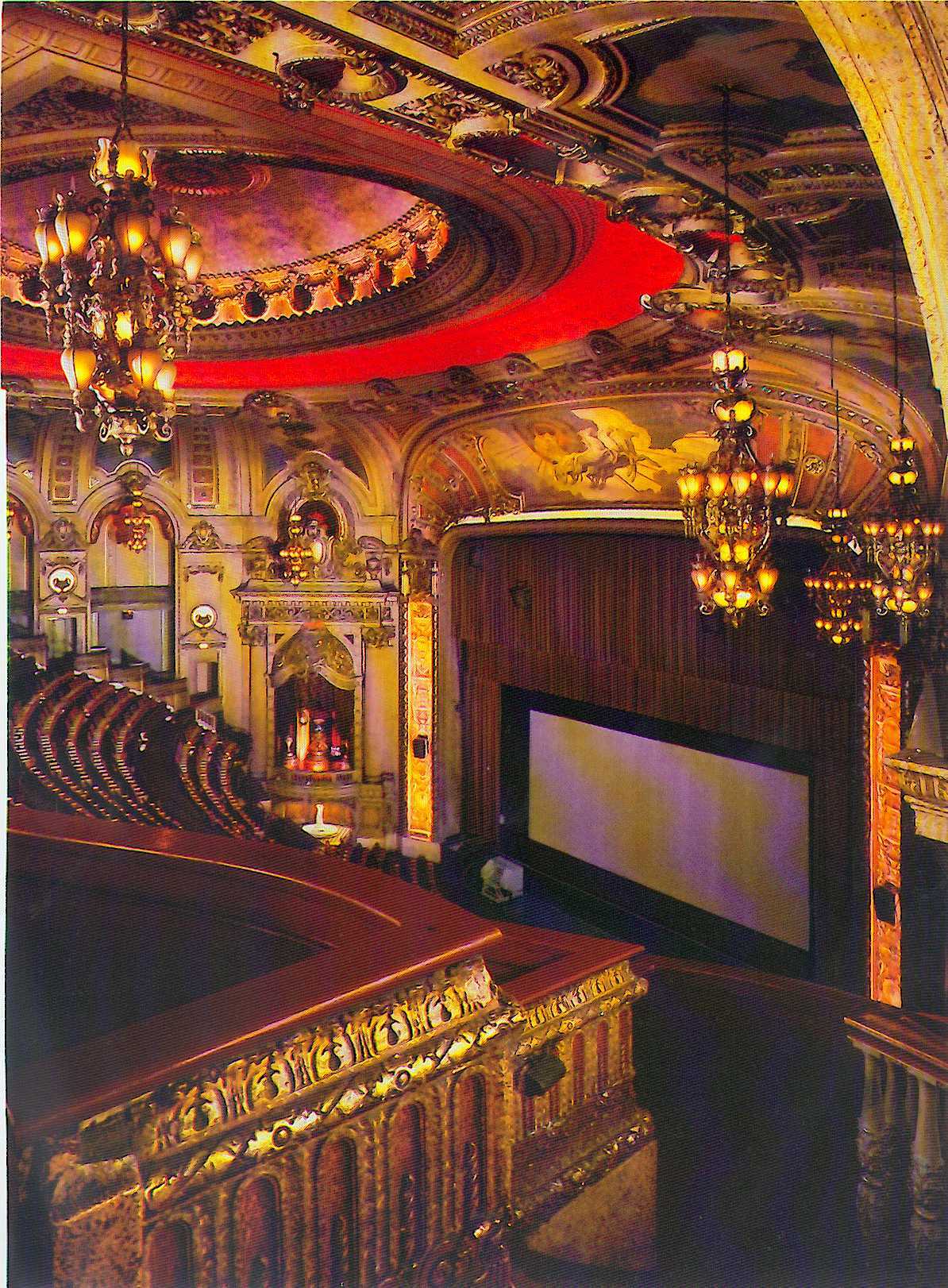}
\includegraphics[width=0.15\textwidth,height=1.6cm]{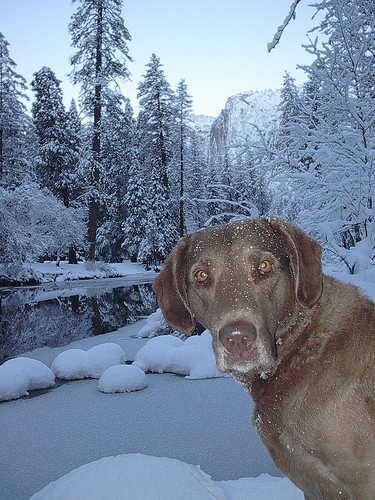}
\includegraphics[width=0.15\textwidth,height=1.6cm]{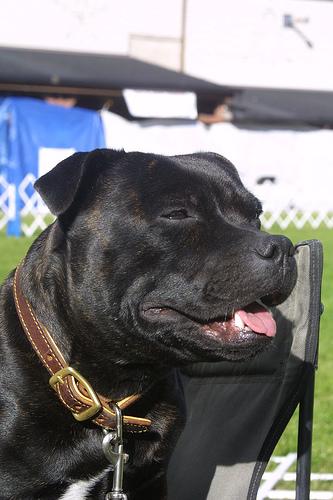}
\includegraphics[width=0.15\textwidth,height=1.6cm]{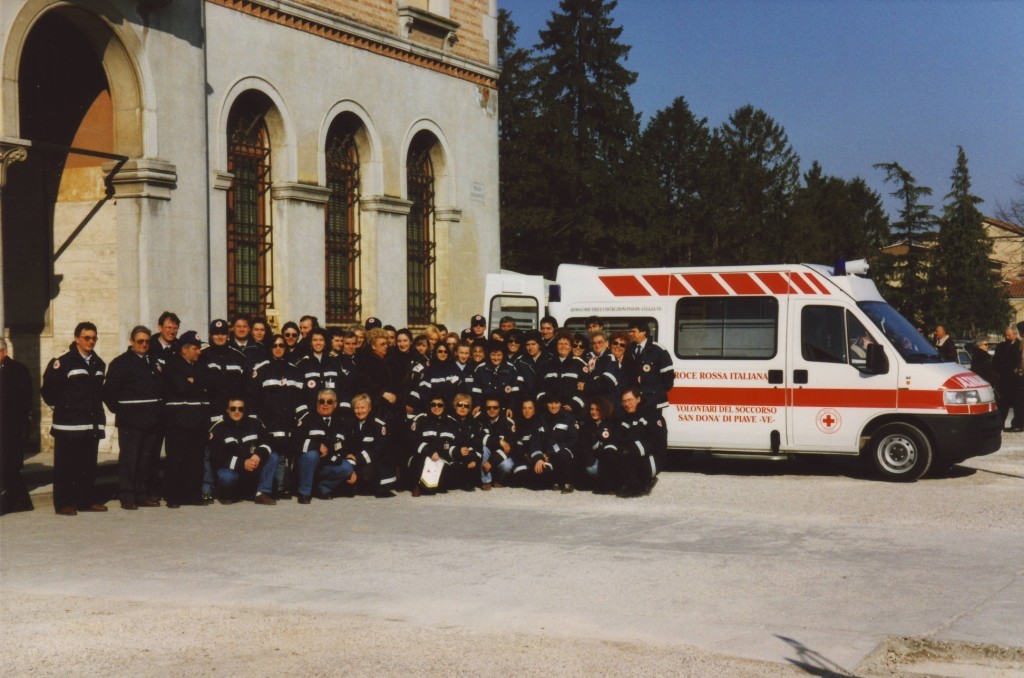}
\includegraphics[width=0.15\textwidth,height=1.6cm]{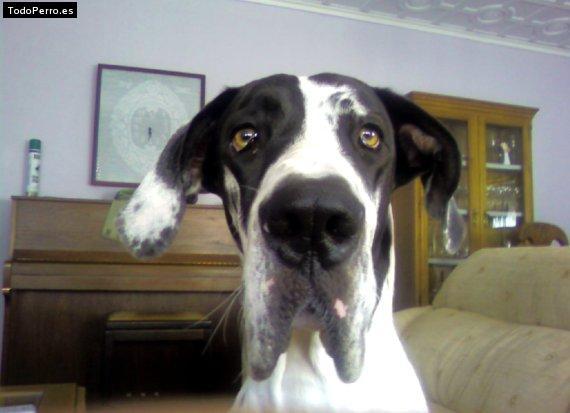}
\includegraphics[width=0.15\textwidth,height=1.6cm]{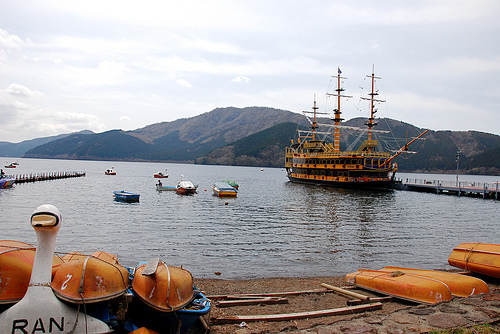}
\parbox{0.15\textwidth}{\centering Boathouse}
\parbox{0.15\textwidth}{\centering Bannister}
\parbox{0.15\textwidth}{\centering Harmonica}
\parbox{0.15\textwidth}{\centering Pineapple}
\parbox{0.15\textwidth}{\centering Stretcher}
\parbox{0.15\textwidth}{\centering Rock Crab}
\includegraphics[width=0.15\textwidth,height=1.6cm]{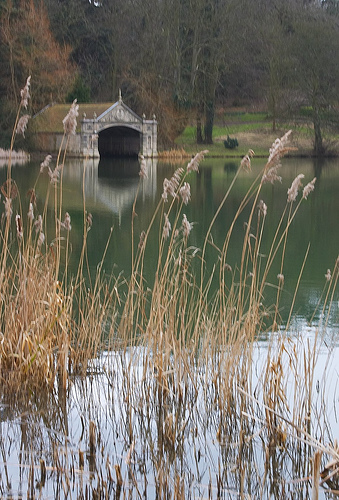}
\includegraphics[width=0.15\textwidth,height=1.6cm]{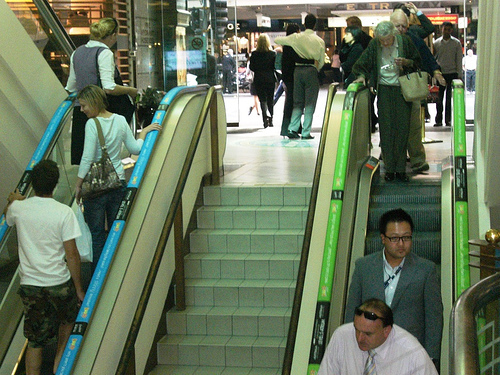}
\includegraphics[width=0.15\textwidth,height=1.6cm]{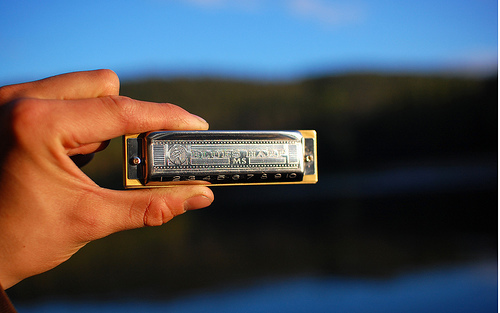}
\includegraphics[width=0.15\textwidth,height=1.6cm]{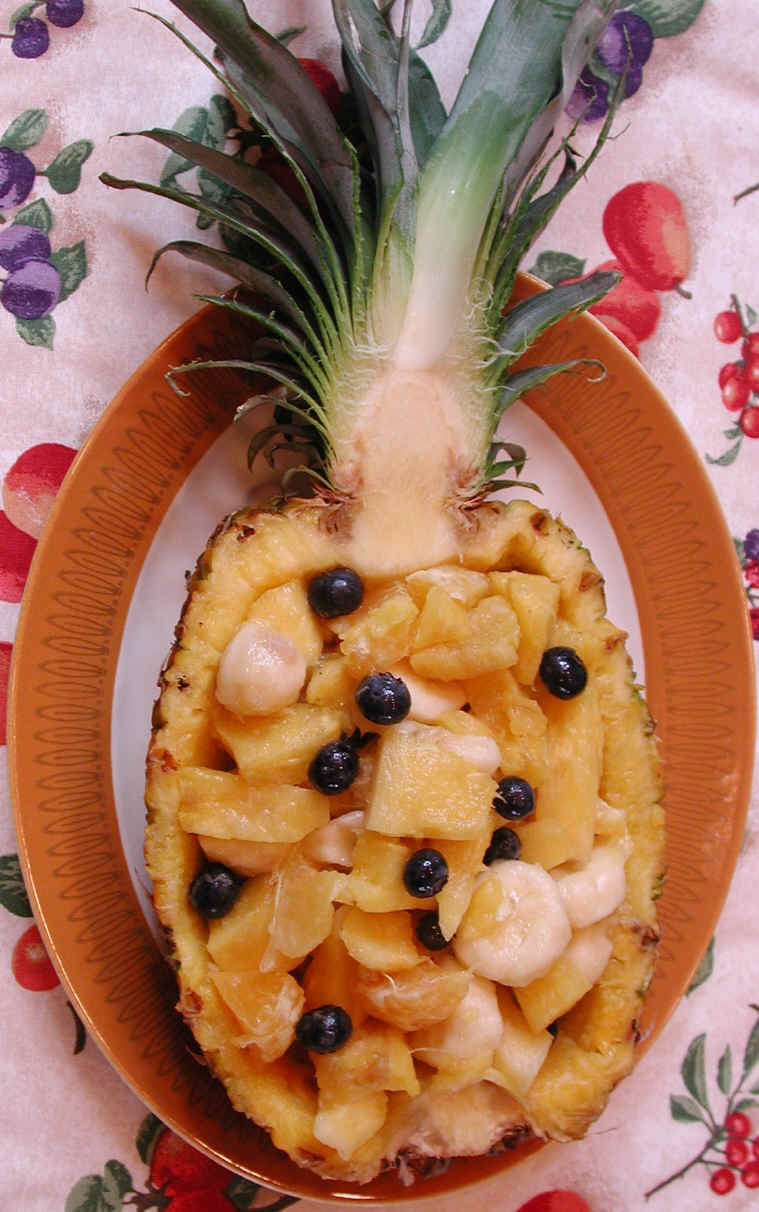}
\includegraphics[width=0.15\textwidth,height=1.6cm]{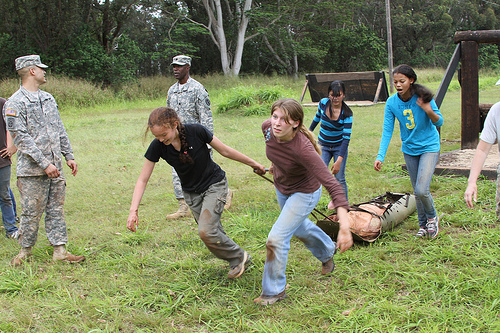}
\includegraphics[width=0.15\textwidth,height=1.6cm]{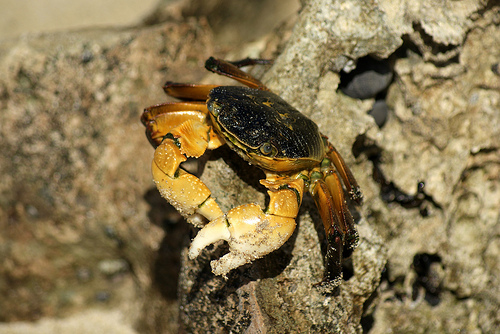}
\parbox{0.15\textwidth}{\centering African Hunting Dog}
\parbox{0.15\textwidth}{\centering Boxer}
\parbox{0.15\textwidth}{\centering Sarong}
\parbox{0.15\textwidth}{\centering Rocking Chair}
\parbox{0.15\textwidth}{\centering Mixing Bowl}
\parbox{0.15\textwidth}{\centering Window Screen}
\includegraphics[width=0.15\textwidth,height=1.6cm]{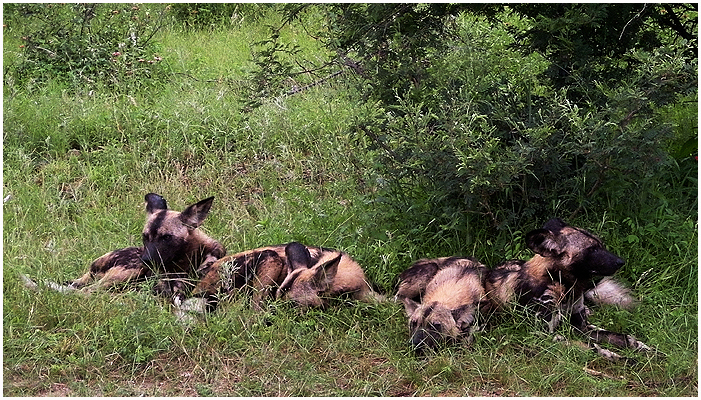}
\includegraphics[width=0.15\textwidth,height=1.6cm]{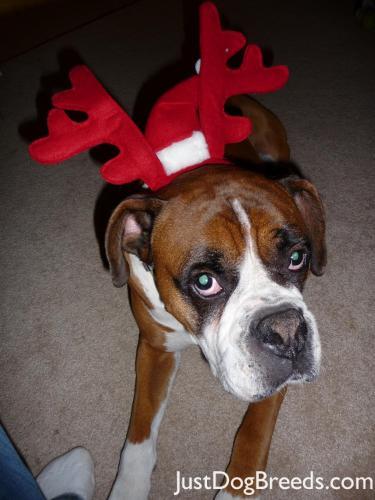}
\includegraphics[width=0.15\textwidth,height=1.6cm]{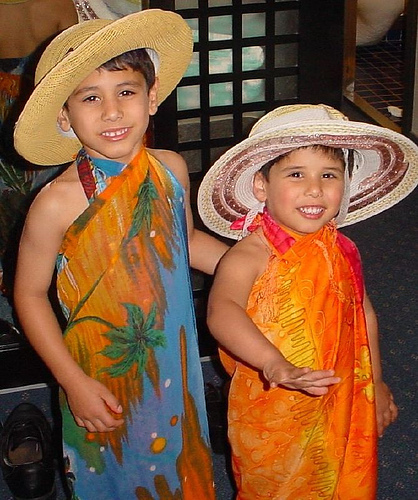}
\includegraphics[width=0.15\textwidth,height=1.6cm]{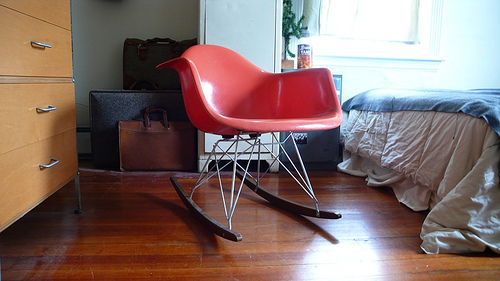}
\includegraphics[width=0.15\textwidth,height=1.6cm]{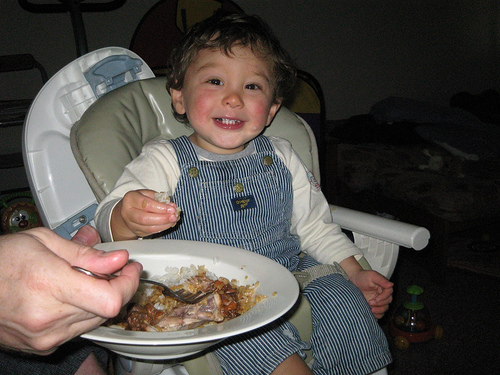}
\includegraphics[width=0.15\textwidth,height=1.6cm]{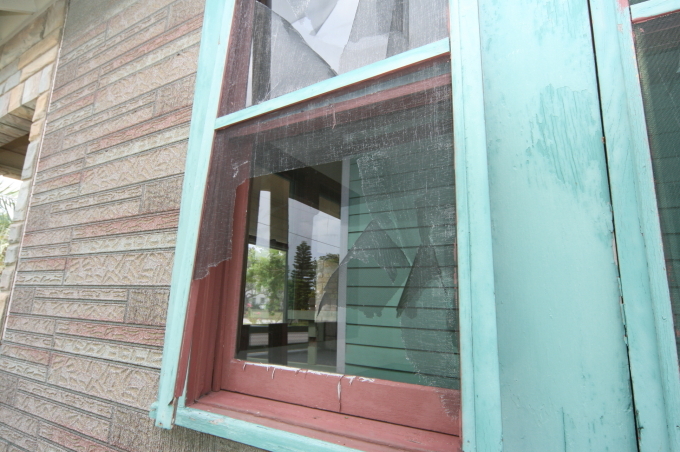}
\caption{Images with 2 annotations on ImageNet100}
\label{sup:fig:2_annotation}
\end{figure*}

\begin{figure*}[h]
\centering
\parbox{0.15\textwidth}{\centering Jean}
\parbox{0.15\textwidth}{\centering Komondor}
\parbox{0.15\textwidth}{\centering Harmonica}
\parbox{0.15\textwidth}{\centering Bottlecap}
\parbox{0.15\textwidth}{\centering American Staffordshire Terrier}
\parbox{0.15\textwidth}{\centering English Foxhound}
\includegraphics[width=0.15\textwidth,height=1.6cm]{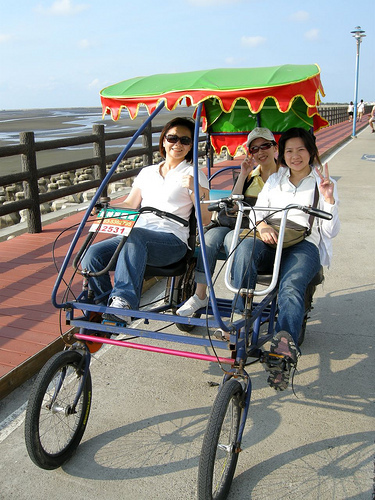}
\includegraphics[width=0.15\textwidth,height=1.6cm]{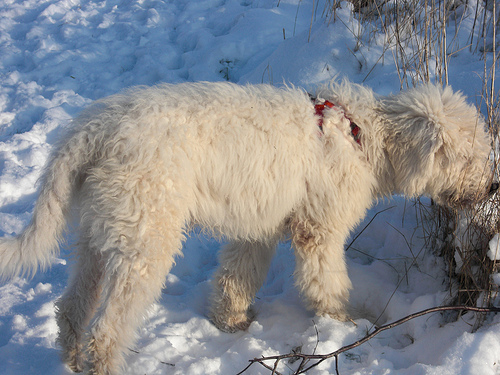}
\includegraphics[width=0.15\textwidth,height=1.6cm]{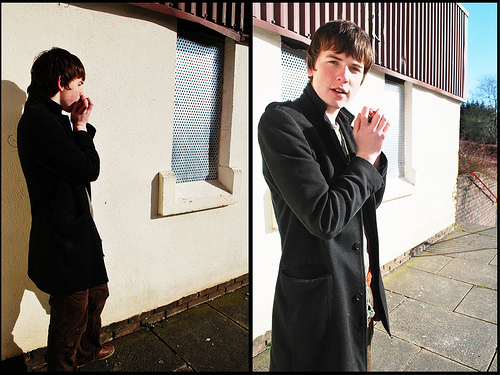}
\includegraphics[width=0.15\textwidth,height=1.6cm]{assets/imagenet100_qualitative/n02877765/n02877765_2225.JPEG}
\includegraphics[width=0.15\textwidth,height=1.6cm]{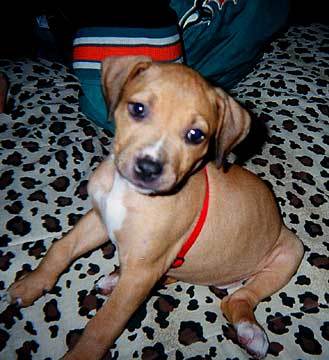}
\includegraphics[width=0.15\textwidth,height=1.6cm]{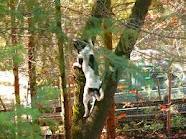}
\parbox{0.15\textwidth}{\centering Vizsla}
\parbox{0.15\textwidth}{\centering Football Helmet}
\parbox{0.15\textwidth}{\centering Walker Hound}
\parbox{0.15\textwidth}{\centering American Coot}
\parbox{0.15\textwidth}{\centering American Staffordshire Terrier}
\parbox{0.15\textwidth}{\centering American Lobster}
\includegraphics[width=0.15\textwidth,height=1.6cm]{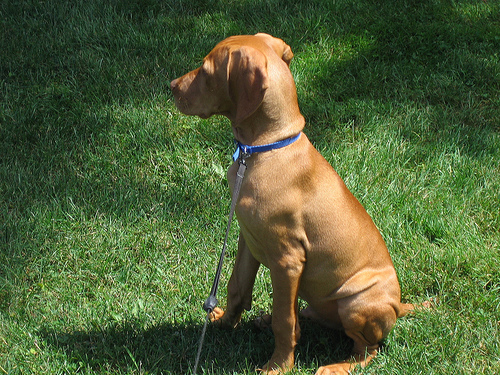}
\includegraphics[width=0.15\textwidth,height=1.6cm]{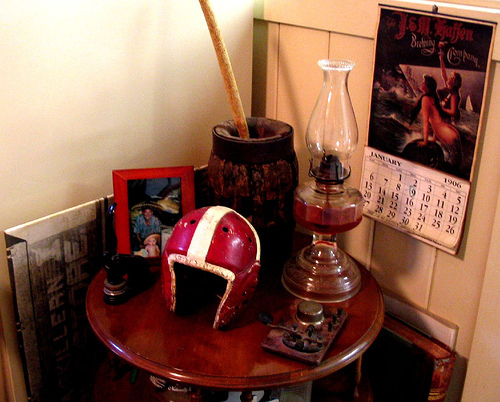}
\includegraphics[width=0.15\textwidth,height=1.6cm]{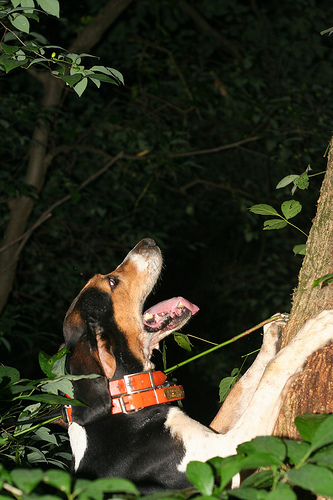}
\includegraphics[width=0.15\textwidth,height=1.6cm]{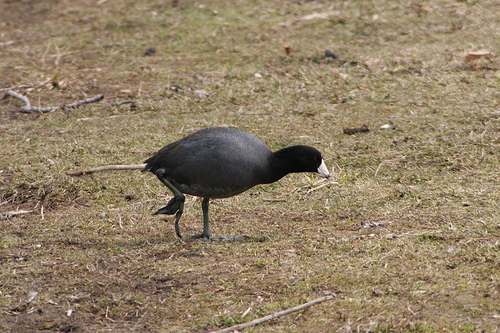}
\includegraphics[width=0.15\textwidth,height=1.6cm]{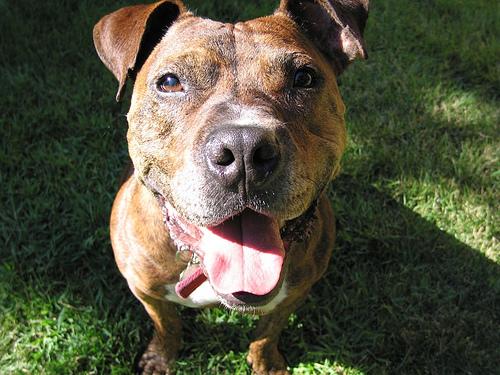}
\includegraphics[width=0.15\textwidth,height=1.6cm]{assets/imagenet100_qualitative/n01983481/n01983481_20024.JPEG}
\parbox{0.15\textwidth}{\centering Harmonica}
\parbox{0.15\textwidth}{\centering Saluki}
\parbox{0.15\textwidth}{\centering Pickup}
\parbox{0.15\textwidth}{\centering Walker Hound}
\parbox{0.15\textwidth}{\centering Moped}
\parbox{0.15\textwidth}{\centering Rock Crab}
\includegraphics[width=0.15\textwidth,height=1.6cm]{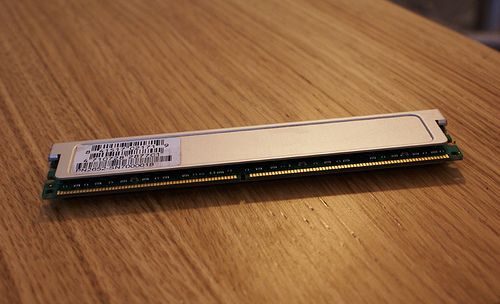}
\includegraphics[width=0.15\textwidth,height=1.6cm]{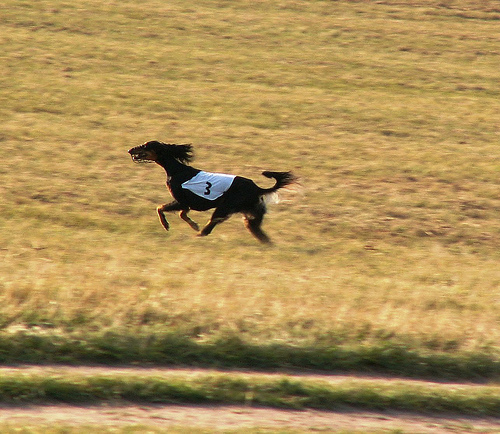}
\includegraphics[width=0.15\textwidth,height=1.6cm]{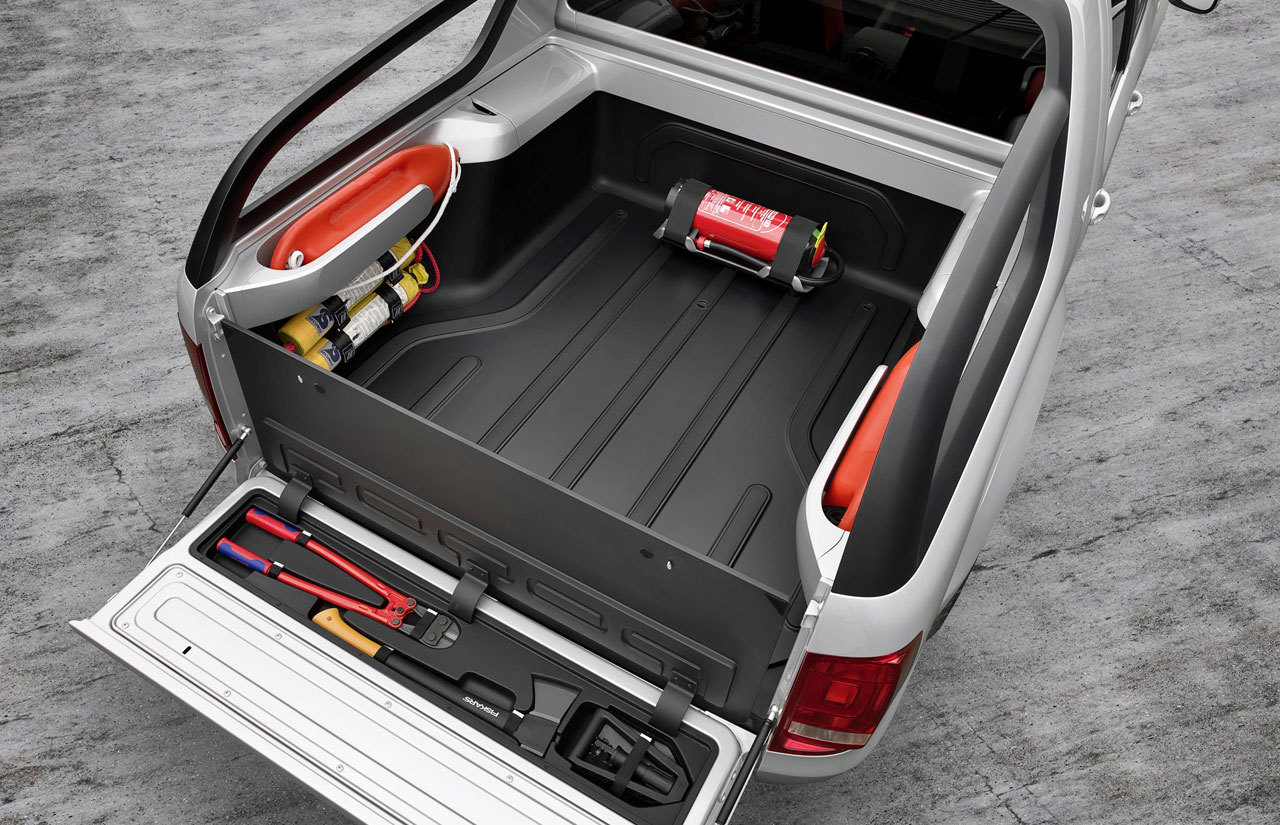}
\includegraphics[width=0.15\textwidth,height=1.6cm]{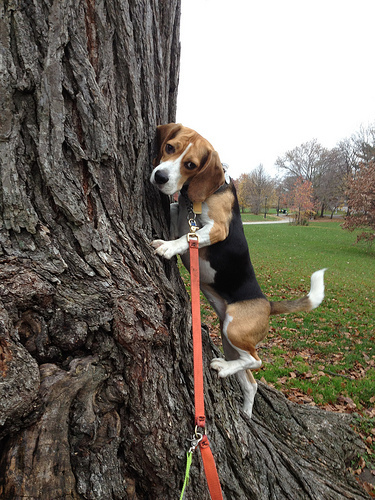}
\includegraphics[width=0.15\textwidth,height=1.6cm]{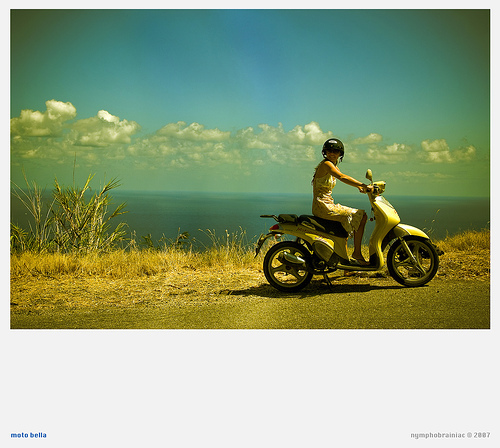}
\includegraphics[width=0.15\textwidth,height=1.6cm]{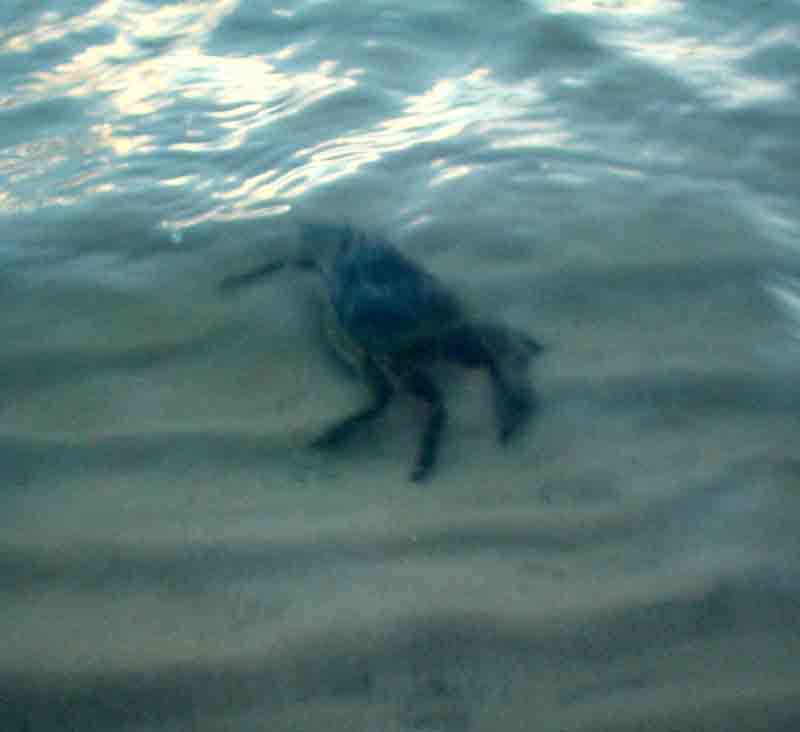}
\caption{Images with 3 annotations on ImageNet100}
\label{sup:fig:3_annotation}
\end{figure*}

In Fig.~\ref{sup:fig:qualitative_unfinished}, we show randomly sampled images from the unfinished set. 
Most of them are images with a fine-grained label, \eg Fiddler Crab, Great Dane, Borzoi, etc, and some of them are shot from unusual angles, \eg Chime and Basinet. We also show additional random examples with zero, one, two and three annotations in our ImageNet100 experiment in Fig.~\ref{sup:fig:0_annotation}, Fig.~\ref{sup:fig:1_annotation}, Fig.~\ref{sup:fig:2_annotation} and Fig.~\ref{sup:fig:3_annotation} respectively.

\end{document}